\documentclass[10pt,twocolumn,letterpaper]{article}

\usepackage[pagenumbers]{cvpr} 

\newif\ifdrafting
\draftingfalse 
\ifdrafting
    \newcommand{\kelvin}[1]{{\color{cyan}[Kelvin: #1]}}
    \newcommand{\yuan}[1]{{\color{ForestGreen}[Li-Yuan: #1]}}
    \newcommand{\ds}[1]{\textcolor{red}{[ds: #1]}}

\else
    \newcommand{\yuan}[1]{}
    \newcommand{\kelvin}[1]{}
    \newcommand{\ds}[1]{}

\fi
\usepackage{multirow}
\usepackage{graphicx}  
\usepackage[hypcap=false]{caption}  
\usepackage{subcaption}
\usepackage{mwe}
\usepackage{colortbl}
\usepackage{booktabs}

\newcommand{\ourmethod}{HoliSDiP}
\definecolor{cvprblue}{rgb}{0.21,0.49,0.74}
\usepackage[pagebackref,breaklinks,colorlinks,allcolors=cvprblue]{hyperref}

\makeatletter
\let\@oldmaketitle\@maketitle
\renewcommand{\@maketitle}{\@oldmaketitle
\begin{center} 
    \vspace{-5mm}
    \hspace{-2mm}
    \begin{minipage}{\textwidth} 
        \begin{minipage}[t]{0.5\textwidth}
            \centering
            \begin{minipage}[t]{0.32\textwidth}
                \vtop{
                    \centering
                    \includegraphics[width=1\linewidth]{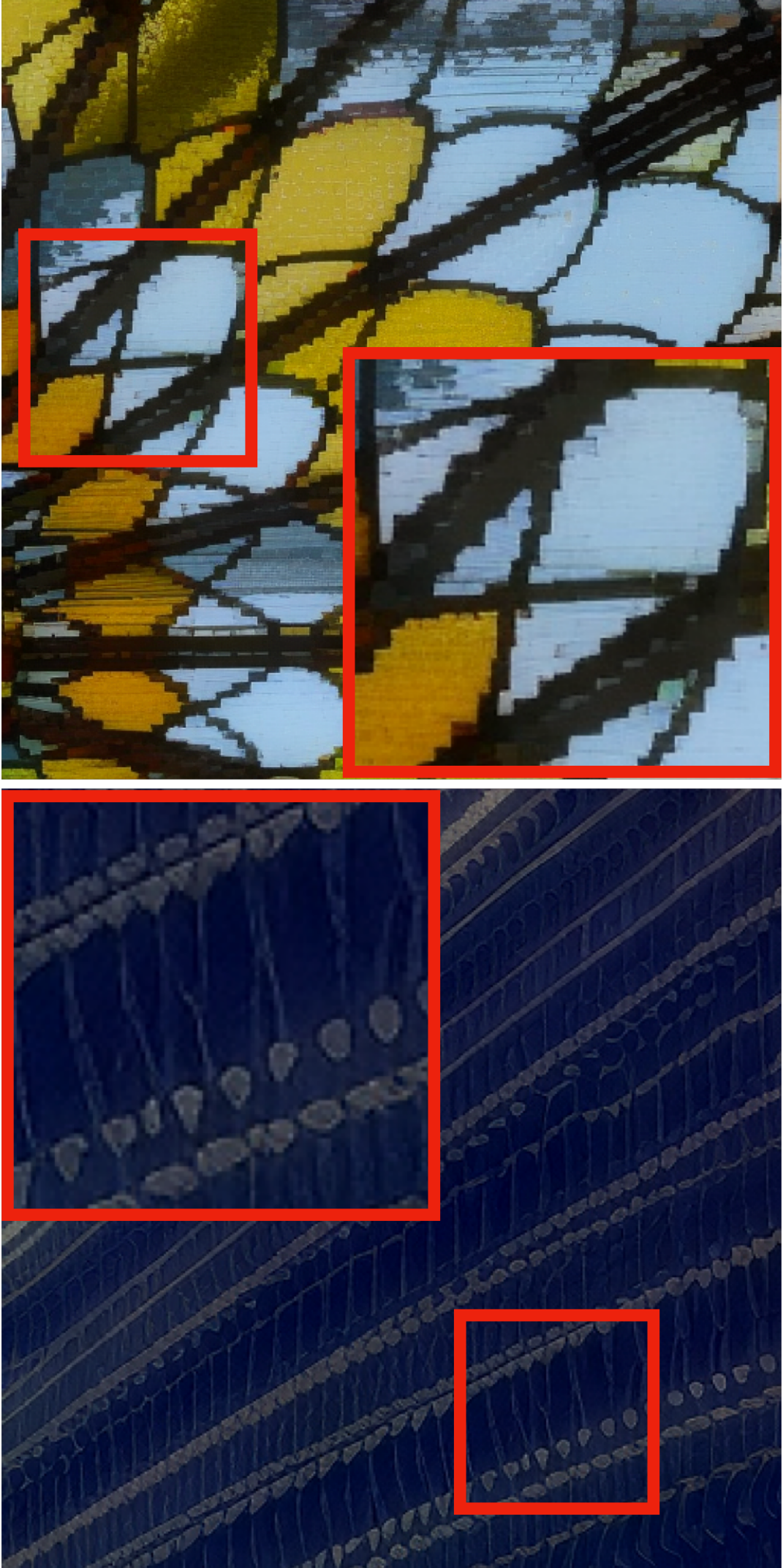}
                    \par\scriptsize DiffBIR~\cite{diffbir}
                }
            \end{minipage}
            \begin{minipage}[t]{0.32\textwidth}
                \vtop{
                    \centering
                    \includegraphics[width=1\linewidth]{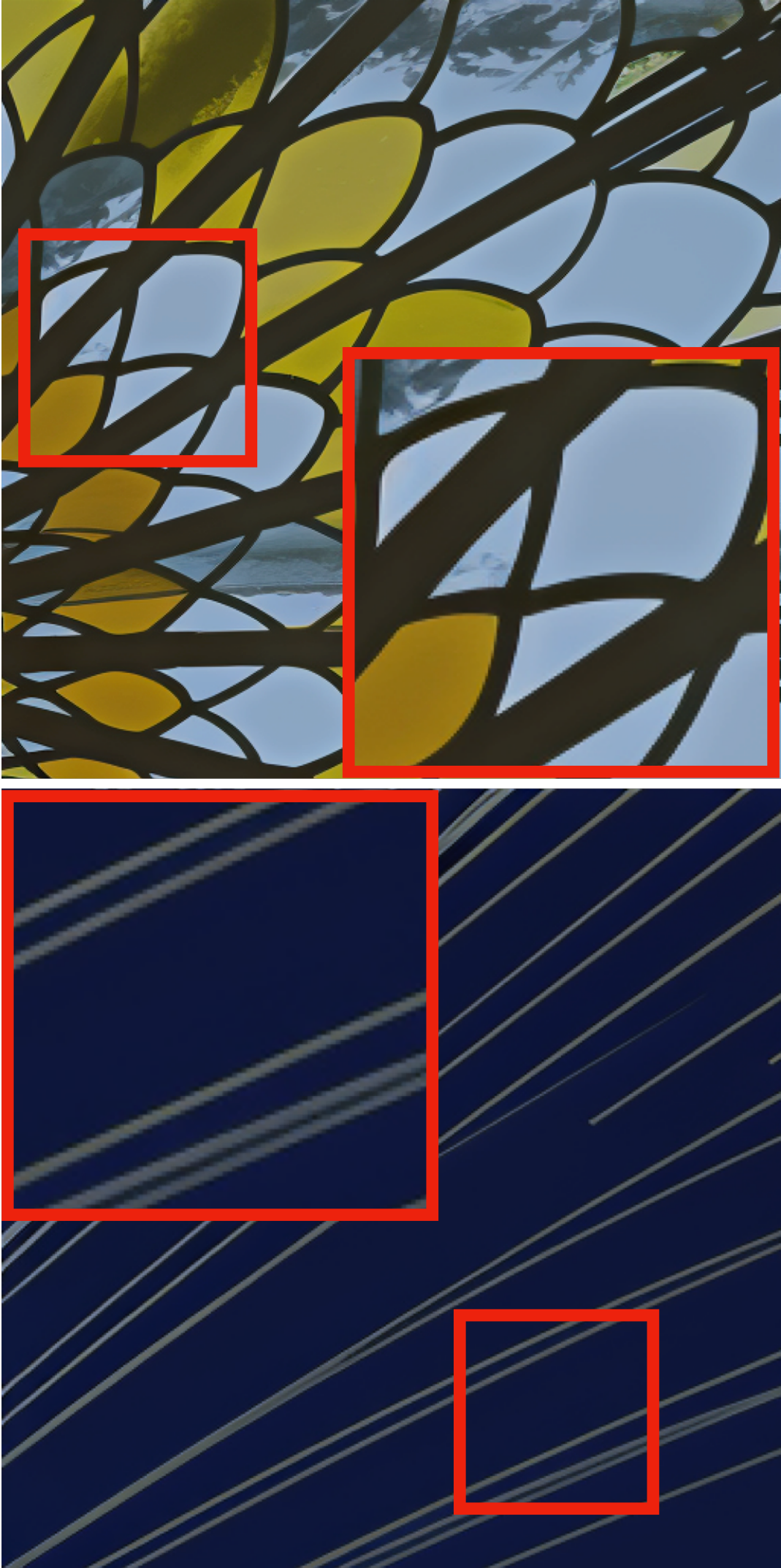}
                    \par\scriptsize \ourmethod{} (Ours)
                }
            \end{minipage}
            \begin{minipage}[t]{0.32\textwidth}
                \vtop{
                    \centering
                    \includegraphics[width=1\linewidth]{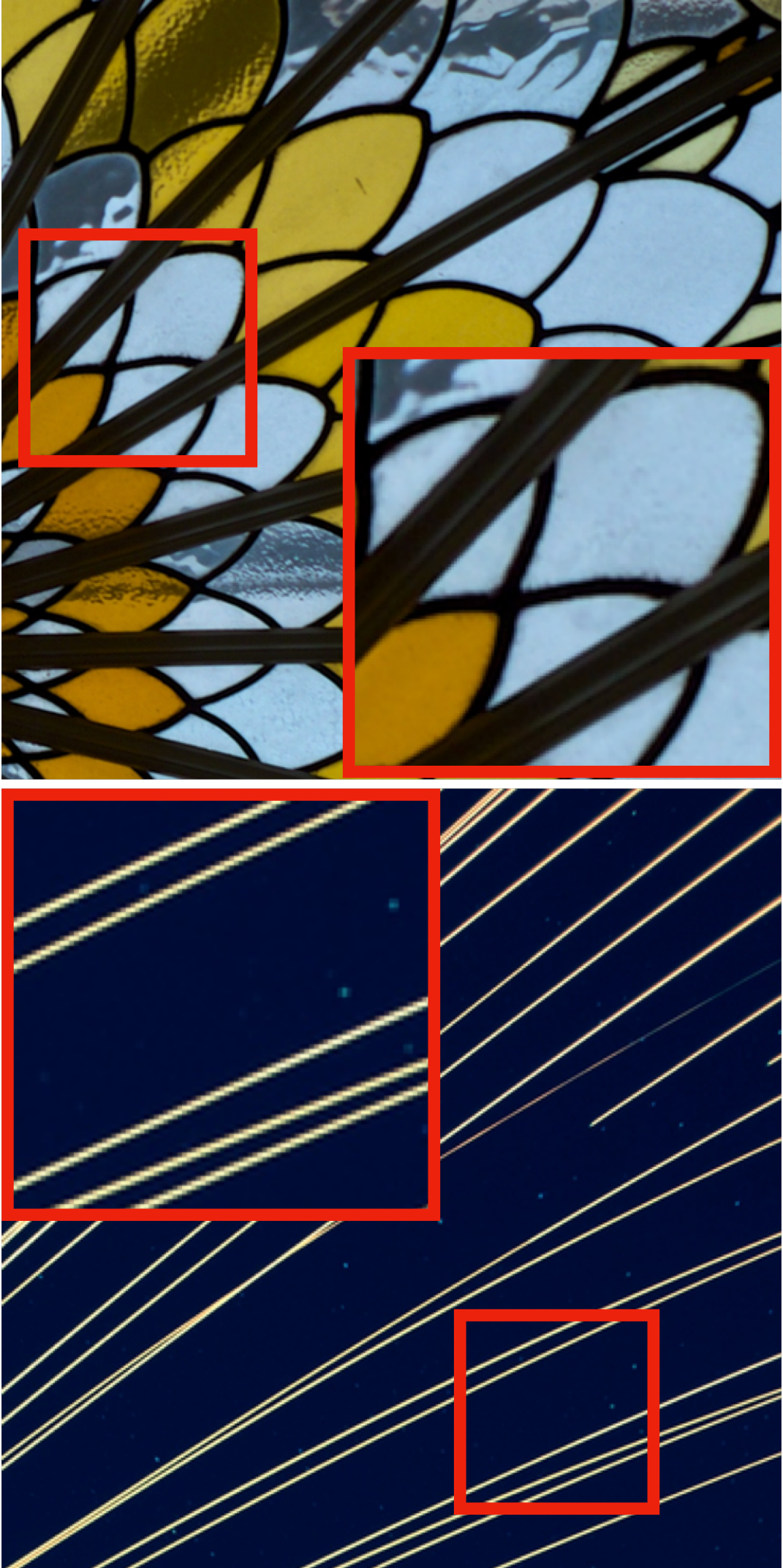}
                    \par\scriptsize Ground Truth
                }
            \end{minipage}
        \end{minipage}
        \hspace{-0mm}
        \begin{minipage}[t]{0.5\textwidth}
            \centering
            \begin{minipage}[t]{0.32\textwidth}
                \vtop{
                    \centering
                    \includegraphics[width=\linewidth]{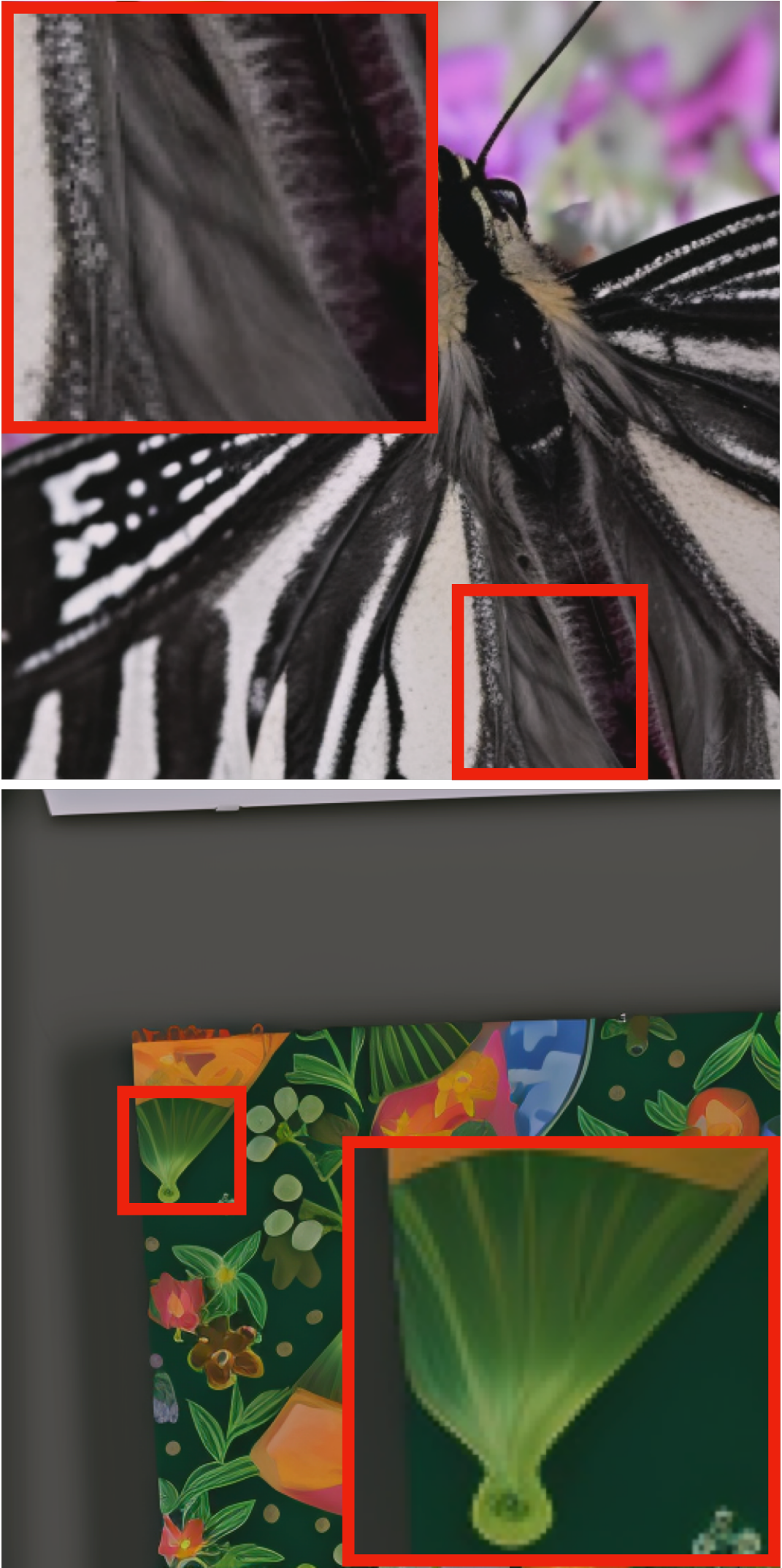}
                    \par\scriptsize SeeSR~\cite{seesr}
                }
            \end{minipage}
            \begin{minipage}[t]{0.32\textwidth}
                \vtop{
                    \centering
                    \includegraphics[width=\linewidth]{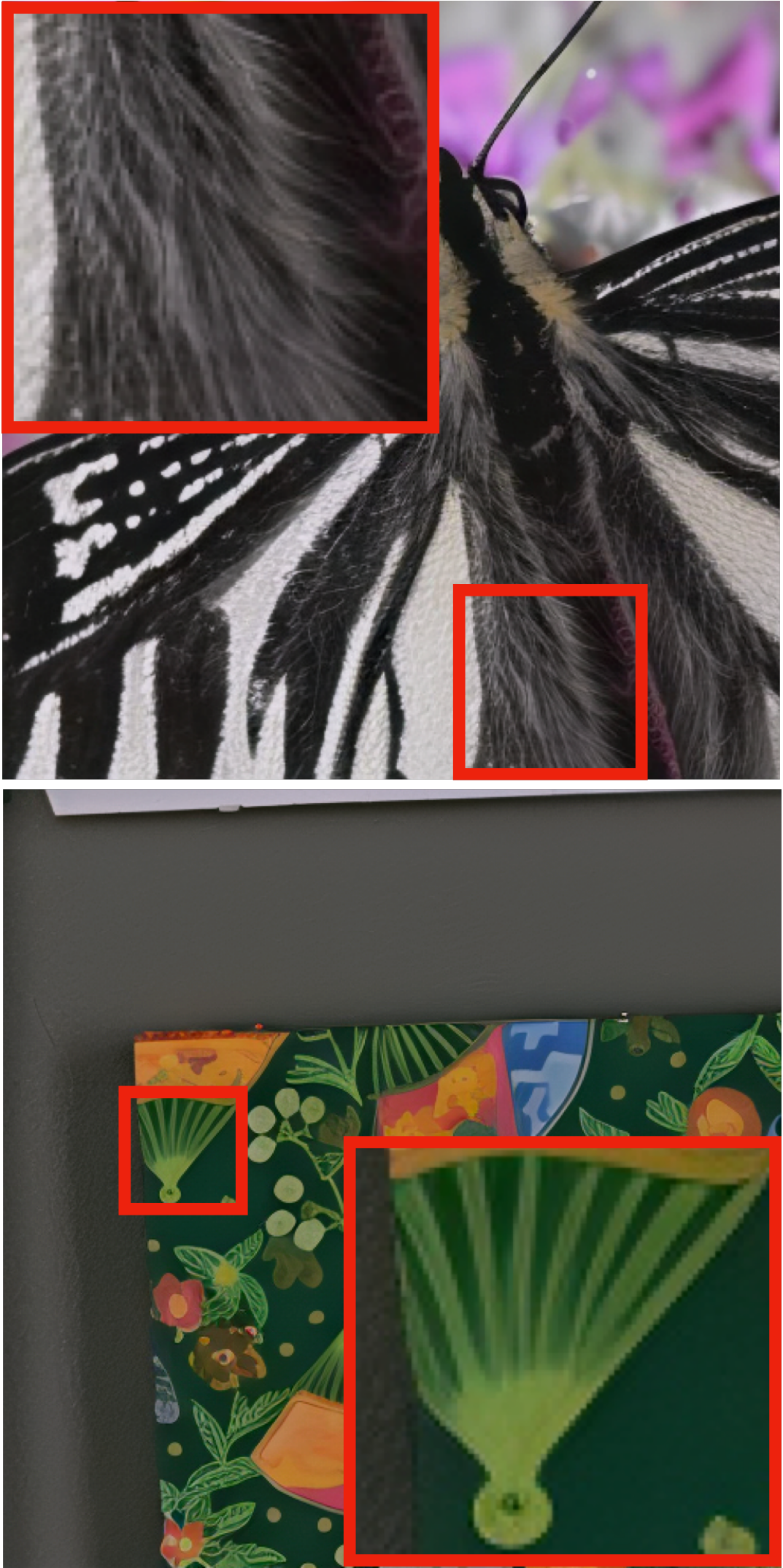}
                    \par\scriptsize \ourmethod{} (Ours)
                }
            \end{minipage}
            \begin{minipage}[t]{0.32\textwidth}
                \vtop{
                    \centering
                    \includegraphics[width=\linewidth]
                    {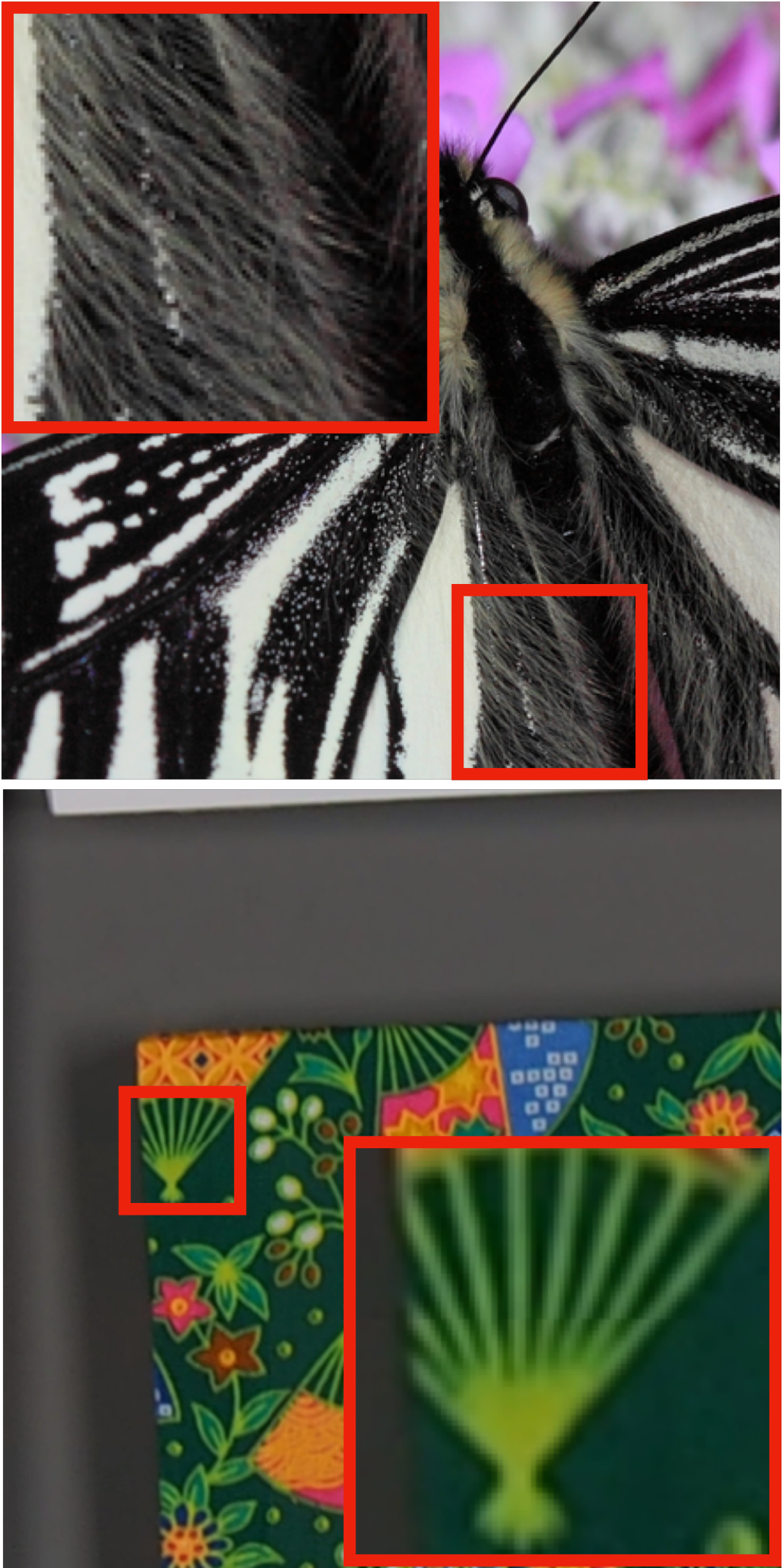}
                    \par\scriptsize Ground Truth
                }
            \end{minipage}
        \end{minipage}
    \end{minipage}
    \vspace{-3pt}
    \captionof{figure}{The proposed \ourmethod{} performs well against the state-of-the-art frameworks~\cite{seesr, diffbir} by offering precise and multi-scale semantics, guiding the text-to-image diffusion model to synthesize high-quality images with fine details.}
    \label{fig:teaser}
\end{center}%
}
\makeatother

\def\paperID{1822} 
\def\confName{CVPR}
\def\confYear{2025}

\title{\ourmethod{}: Image Super-Resolution via Holistic Semantics and Diffusion Prior}

\author{
Li-Yuan Tsao$^{1}$ \quad\quad Hao-Wei Chen$^{2}$ \quad\quad Hao-Wei Chung$^{3}$ \quad\quad Deqing Sun$^{4}$\\
Chun-Yi Lee$^{2}$ \quad\quad Kelvin C.K. Chan$^{4}$ \quad\quad Ming-Hsuan Yang$^{4,5}$\\ \\
$^{1}$National Tsing Hua University \quad $^{2}$National Taiwan University \quad $^{3}$Carnegie Mellon University\\ $^{4}$Google DeepMind \quad $^{5}$University of California, Merced
}

\begin{document}
\maketitle

\begin{abstract}

Text-to-image diffusion models have emerged as powerful priors for real-world image super-resolution (Real-ISR). However, existing methods may produce unintended results due to noisy text prompts and their lack of spatial information. In this paper, we present \ourmethod{}, a framework that leverages semantic segmentation to provide both precise textual and spatial guidance for diffusion-based Real-ISR. Our method employs semantic labels as concise text prompts while introducing dense semantic guidance through segmentation masks and our proposed Segmentation-CLIP Map. Extensive experiments demonstrate that \ourmethod{} achieves significant improvement in image quality across various Real-ISR scenarios through reduced prompt noise and enhanced spatial control. Our project page: \url{https://liyuantsao.github.io/HoliSDiP}
   \vspace{\baselineskip}%
\end{abstract}

\vspace*{-3em}
\section{Introduction}
\label{sec::introduction}

Real-world Image Super-Resolution (Real-ISR) aims to reconstruct high-resolution (HR) images from their low-resolution (LR) counterparts, 
which are often corrupted by a wide range of unknown factors, such as sensor noise, lens blur, compression artifacts, and various types of distortions. Recovering high-quality, HR images from these degraded LR inputs is an inherently ill-posed problem, as there are infinitely many possible HR solutions that could have generated the same LR observation. This is where the role of image priors~\cite{dip, srsr, gpp} becomes crucial.
\ds{cite some important prior models for SR.}

\begin{figure}[t]
    \centering    
    \begin{subfigure}[b]{\linewidth}
        \centering
        \includegraphics[width=\linewidth]{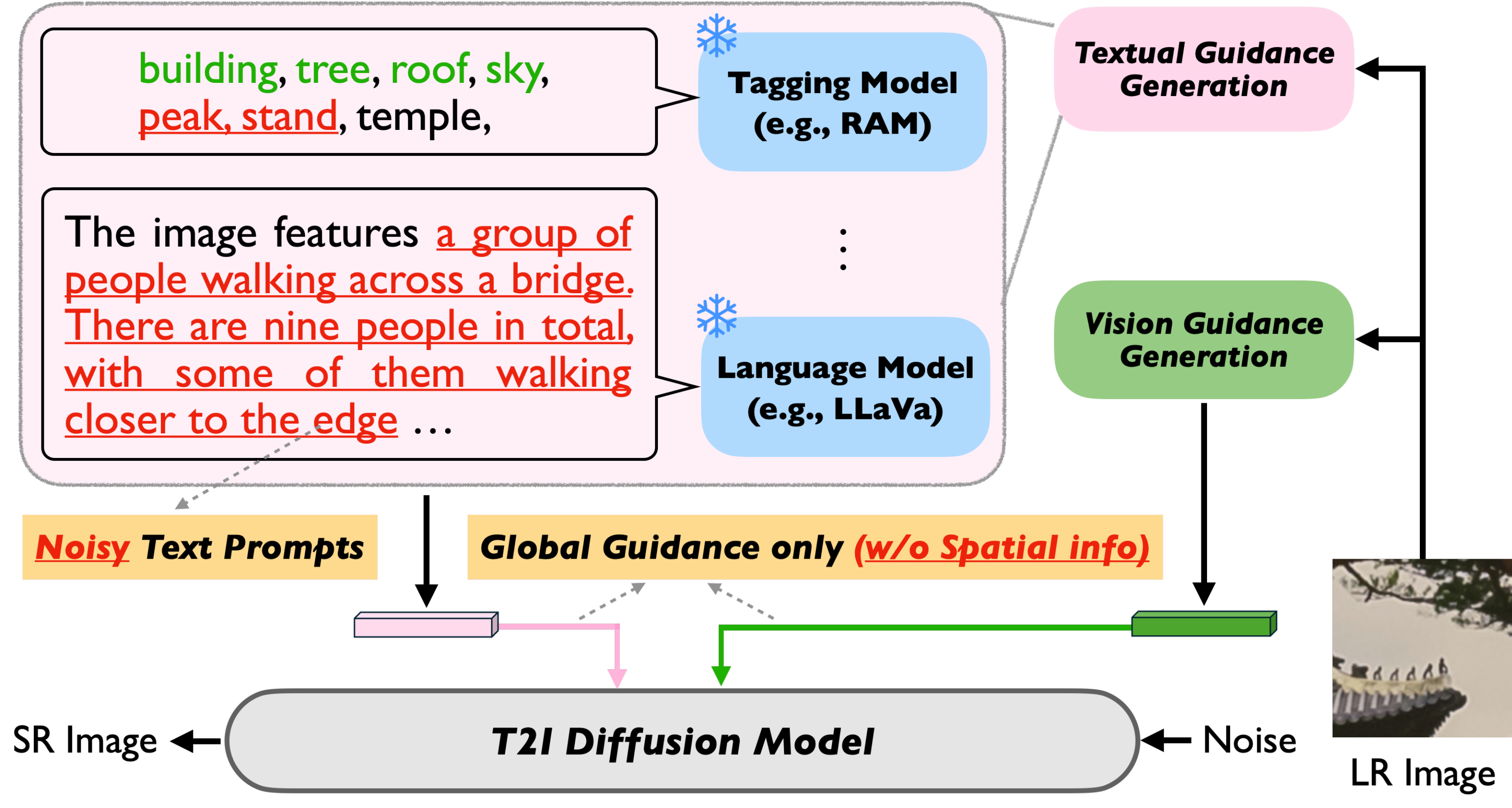}
    \end{subfigure}
    
    \centerline{\footnotesize (a) Contemporary Real-ISR methods}
    \vspace{1mm}
    \begin{subfigure}[b]{\linewidth}
        \centering
        \includegraphics[width=\linewidth]{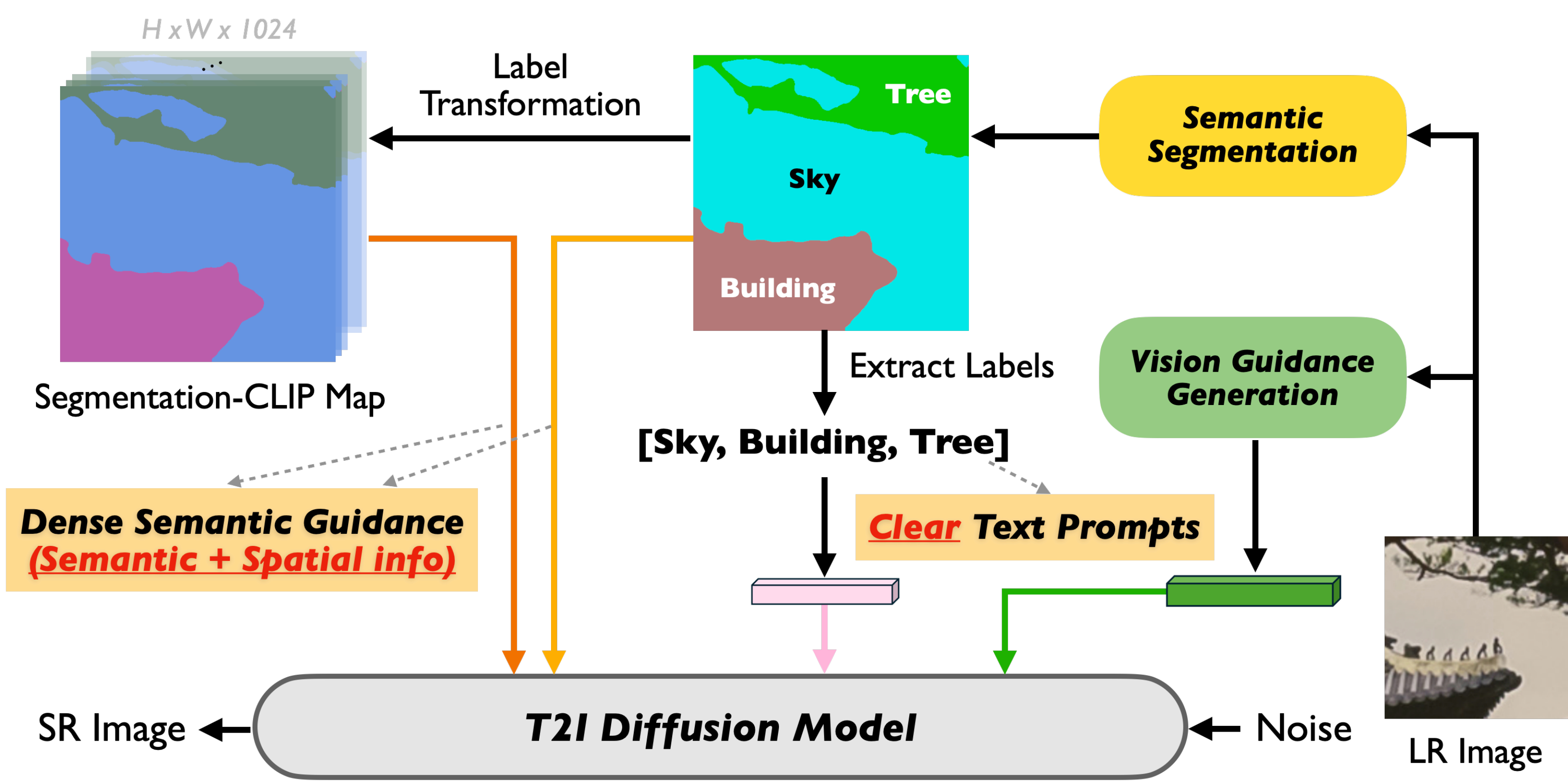}
    \end{subfigure}
    \centerline{\footnotesize (b) The proposed {\ourmethod}}
    \caption{Comparison between the proposed {\ourmethod} and existing Real-ISR methods. (a) Current studies leverage text prompts that include redundant descriptions and lack localized priors. (b) Our {\ourmethod} leverages semantic segmentation to offer clear text prompts and dense semantic guidance.}
    \label{fig:concept}
\end{figure}

A recent promising direction is to leverage pre-trained text-to-image (T2I) diffusion models~\cite{stablediffusion} as image priors for super-resolution. These models have been trained on vast, diverse datasets of high-quality images, enabling them to learn rich, semantically-aware representations of visual content. However, existing methods often face challenges in effectively utilizing these T2I diffusion priors. Current approaches typically rely on automatic tagging methods~\cite{ram} or large language models (LLMs)~\cite{llava} to generate text prompts. These generated prompts may be redundant (containing unnecessary or repetitive information), erroneous (misidentifying objects or attributes), or lack precise spatial alignment with the objects' shapes, scales, and locations in the image, as illustrated in Fig.~\ref{fig:concept} (a).

To address these challenges, we propose a \textit{Holistic Semantic and Diffusion Prior} (\ourmethod{}), a novel framework that leverages semantic segmentation to provide precise and holistic guidance for T2I diffusion models. Our key insight is that semantic segmentation simultaneously offers both textual labels and spatial masks, enabling effective cross-modal guidance from a unified representation. This dual nature of semantic segmentation makes it particularly well-suited for controlling T2I diffusion models in the context of image super-resolution.

\ourmethod{} consists of two key components, as shown in Fig.~\ref{fig:concept} (b). First, we introduce \textit{Semantic Label-Based Prompting} (SLBP), which mitigates the noise and redundancy commonly found in automatically generated text prompts. SLBP extracts semantic labels directly from the segmentation map and utilizes them as text prompts, leveraging segmentation's inherent ability to identify and summarize the primary objects within an image. SLBP provides the T2I diffusion model with concise, accurate descriptions that capture a global understanding of the LR input.

Second, we present \textit{Dense Semantic Guidance} (DSG) to enable fine-grained control over local details. DSG complements SLBP's coarse-grained descriptions by incorporating pixel-wise class priors through two key representations: (1) the segmentation mask directly extracted from the segmentation results, and (2) our proposed \textit{Segmentation-CLIP Map} (SCMap), which enriches the spatial information by converting each pixel's semantic label into CLIP embeddings. These dense semantic signals are integrated with the T2I model's intermediate features through our proposed \textit{Guidance Fusion Module} (GFM), enabling precise feature refinement at multiple scales.

Through the combination of SLBP and DSG, {\ourmethod} effectively addresses the limitations of existing methods by reducing prompt redundancy while providing spatially-aligned, multi-scale semantic guidance for the T2I diffusion model. This holistic approach enables more accurate and controllable image super-resolution results.

Extensive experiments show that {\ourmethod} achieves competitive performance compared to state-of-the-art Real-ISR methods across various metrics and scenarios. As shown in Fig.~\ref{fig:teaser}, {\ourmethod} generates more accurate super-resolution results with reduced artifacts, sharper image structure, and improved overall visual quality, while utilizing more concise and precise text descriptions. We make the following contributions in this work: 

\begin{itemize}
    \item We propose {\ourmethod}, a novel framework that leverages semantic segmentation to provide precise and comprehensive guidance for T2I diffusion model-based Real-ISR. Our approach addresses key limitations in existing methods by offering both textual and spatial guidance through a unified representation.
    
    \item We introduce \textit{Semantic Label-Based Prompting (SLBP)}, a technique that extracts and utilizes semantic labels as text prompts. This approach generates more concise and accurate object descriptions compared to traditional automatic tagging methods, leading to improved quality.
    
    \item We develop \textit{Dense Semantic Guidance (DSG)}, combining segmentation masks with our proposed Segmentation-CLIP Map to provide fine-grained, pixel-wise semantic priors. This dense guidance mechanism effectively complements the global context provided by SLBP.
    
    \item We design a \textit{Guidance Fusion Module (GFM)} that effectively integrates DSG signals into the T2I diffusion process, enabling precise refinement of local details through spatially-aligned semantic guidance at multiple scales.
\end{itemize}

\section{Related Work}
\label{sec::related_work}
\vspace{-3pt}
\subsection{Image Super-Resolution}
\vspace{-3pt}
\label{sec:related_isr}
Image Super-Resolution (ISR) methods can be broadly categorized into two groups: Non-blind ISR and Blind ISR, with the latter also referred to as Real-world ISR (Real-ISR)\kelvin{Blind and real-world are not equal?}. Non-blind ISR methods~\cite{srcnn, srgan, edsr, rdn} primarily focus on designing flexible and effective architectures, which are typically trained and evaluated on LR-HR image pairs with simple and predefined degradations (\ie, bicubic downsampling). For instance, continuous ISR methods~\cite{metasr, liif, lte, linf, ciaosr, clit} synthesize images at arbitrary upsampling scales. Additionally, generative ISR approaches use models such as generative adversarial networks (GANs)~\cite{esrgan, sftgan, ranksrgan, glean} or normalizing flows~\cite{srflow, hcflow, bfsr} to enhance image quality. 

Despite their success in non-blind settings%
, these methods often struggle to generalize in real-world scenarios, where the LR images exhibit complex and unknown degradations.
To address these challenges, various types of Real-ISR approaches~\cite{realesrgan, bsrgan, femasr, zssr, ikc, cincgan} are developed, exploring strategies such as degradation modeling~\cite{ikc, fisr, kernelgan}, domain translation~\cite{cincgan, dsgan}, and self training~\cite{zssr, mzsr}. To further enhance the effectiveness in real-world scenarios, Real-ESRGAN~\cite{realesrgan} and BSRGAN~\cite{bsrgan} introduced intricate degradation pipelines
\ds{are these two degradation modeling-based?}%
\yuan{degradation modeling-based method predicts the degradation kernel of a specific image, and uses this kernel as an auxiliary condition for reference. While ESRGAN/BSRGAN focus on creating different combination of degradation for data augmentation.}%
\ds{what do these pipelines do? generate all kinds of degradations?}%
\yuan{Yes, like data augmentation. Apply blur/add noise/jpeg/resize with randomly sampled parameters}%
for synthesizing LR-HR training pairs, and leverage GANs to generate high-quality images, achieving promising Real-ISR performance. 
\ds{Does our method use the same pipeline to simulate LR-HR pairs?}
\yuan{Yes. (in 4.1 Dataset)}
\ds{We don't need to discuss these two methods in detail if we simply used them. We can tell the impelemntation details in 4.1 or supp.}%
\yuan{11/14 fixed}

\subsection{Diffusion Models for Image Super-Resolution}
\vspace{-3pt}
\label{sec::related_diffsr}
Diffusion models~\cite{dm, stablediffusion} have gained increasing attention in ISR tasks due to their strong generative capabilities. SR3~\cite{sr3} and SRDiff~\cite{srdiff} develop diffusion-based ISR frameworks in non-blind ISR settings. On the other hand, ResShift~\cite{resshift} and SinSR~\cite{sinsr} apply diffusion-based ISR to Real-ISR scenarios, further improving the efficiency by reducing iterative refinement steps.

Recently, pre-trained text-to-image (T2I) diffusion models have gained popularity for Real-ISR because of the powerful diffusion priors, enabling the synthesis of high-quality images. StableSR~\cite{stablesr} integrates learnable SFT layers into the Stable Diffusion~\cite{stablediffusion} backbone, effectively imposing constraints from the LR image and produces high-quality SR images.  Similarly, various other frameworks have been developed to fully exploit diffusion priors for enhancing Real-ISR performance. One prominent research direction focuses on improving vision space guidance, making it easier for the diffusion model to utilize the LR input for accurate reconstruction.
For instance, DiffBIR~\cite{diffbir} employs a two-stage pipeline where a restoration module first reduces degradations in LR images, providing cleaner input conditions for subsequent processing. Additionally, ControlNet~\cite{controlnet} is also widely adopted in T2I diffusion model-based Real-ISR frameworks~\cite{ccsr, pasd, xpsr, seesr, supir}, enabling effective integration of LR information.

Another line of research aims to exploit the potential of T2I diffusion models by leveraging intricate text prompts to guide the generation process. PASD~\cite{pasd} employs architectures such as ResNet~\cite{resnet}, YOLO~\cite{yolo}, and BLIP~\cite{blip} to extract high-level descriptions from LR images, which are then used as text prompts. SeeSR~\cite{seesr} leverages image tags generated by RAM~\cite{ram}
to enhance local details of objects, while SUPIR~\cite{supir} uses LLaVa~\cite{llava} to supply detailed descriptions. XPSR~\cite{xpsr} further refines the instructions provided by LLaVa by disentangling high-level and low-level semantic prompts. 
However, the text descriptions may be erroneous or not well-aligned with the spatial content. 
Here we introduce a holistic semantic guidance for T2I diffusion model by utilizing semantic segmentation, which provides both concise text descriptions and dense semantic cues.

\section{Method}
\label{sec::methodology}
In this section, we elaborate on the motivation and design of \ourmethod{}, which aims to reduce noise in text prompts for accurate guidance, while offering fine-grained semantics to enhance detail refinement. We first provide an overview of \ourmethod{} in Section~\ref{sec::overview}, followed by detailed explanations of Semantic Label-Based Prompting in Section~\ref{sec::slbp}, Dense Semantic Guidance in Section~\ref{sec::dense}, and Guidance Fusion Module in Section~\ref{sec::gfm}.

\subsection{Framework Overview}
\label{sec::overview}
\begin{figure*}[t]
  \centering
  \includegraphics[width=0.89\textwidth]{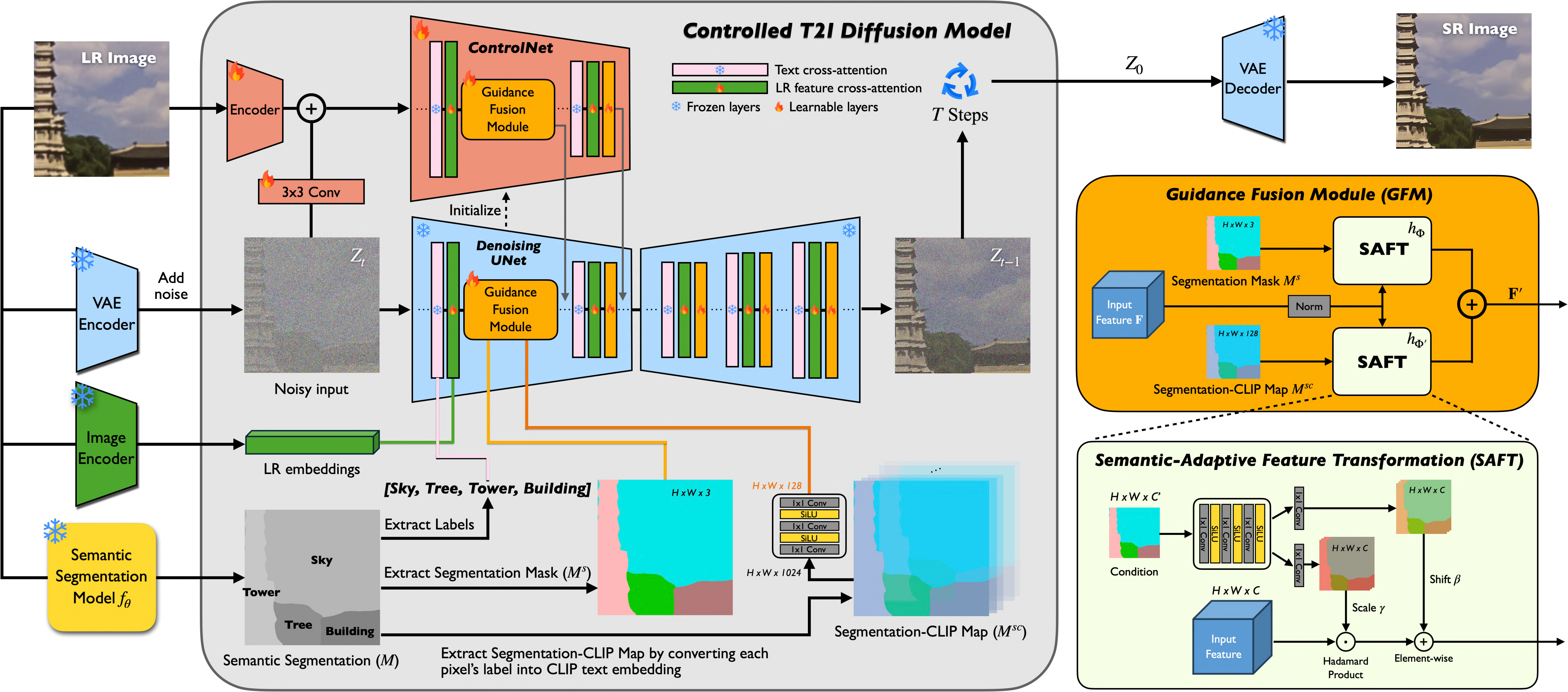}
  \caption{
    Overview of {\ourmethod}. The segmentation model first processes the LR image to generate segmentation results, which is used for extracting semantic labels, segmentation mask, and Segmentation-Clip Map (SCMap). The semantic labels are employed as text prompts to inject textual guidance through cross-attention layers, while the segmentation mask and SCMap are integrated by our Guidance Fusion Module to facilitate semantic-adaptive feature transformation. Additionally, ControlNet and LR cross-attention layers are utilized to strengthen guidance from the LR image. These conditions are incorporated into the denoising UNet, which iteratively refines the noisy input to produce the final SR image.
  }
  \label{fig:framework}
  \vspace{-5pt}
\end{figure*}
Fig.~\ref{fig:framework} presents an overview of {\ourmethod}, which includes three key components: Semantic Label-Based Prompting (SLBP), Dense Semantic Guidance (DSG), and the Guidance Fusion Module (GFM). SLBP collects the labels in segmentation map to offer high-level text descriptions, while DSG utilizes segmentation mask and the proposed Segmentation-CLIP Map (SCMap) 
to supply dense semantic priors. The GFM leverages DSG signals to perform semantic-adaptive feature transformation, refining intermediate features. Collectively, these components embed precise and fine-grained semantic cues into the iterative refinement process of the pre-trained T2I diffusion model, enhancing its capability in synthesizing high-quality images.

\subsection{Semantic Label-Based Prompting}
\label{sec::slbp}
To effectively activate the generative capabilities of T2I diffusion models through textual guidance, it is crucial to provide the model with concise and object-relevant hints. Inspired by the natural ability of semantic segmentation to summarize the primary objects in an image, we propose Semantic Label-Based Prompting (SLBP), which employs the labels in segmentation map as text prompts, supplying precise descriptions for the T2I diffusion model.

Specifically, SLBP begins by extracting semantic labels from the segmentation results of the LR image. As depicted in Fig.~\ref{fig:framework}, for instance, the segmentation map provides labels such as ``sky'', ``tree'', ``tower'', and ``building'', which align well with the image content. These labels are subsequently encoded as text embeddings using a CLIP~\cite{clip} text encoder and integrated into the generation process via cross-attention layers. By leveraging essential object information from the segmentation map, SLBP offers clear textual guidance to the pre-trained T2I diffusion model, preventing the generation of undesired artifacts. A detailed analysis on the effects of SLBP is presented in Section~\ref{sec::ablation_slbp}.

\subsection{Dense Semantic Guidance}
\label{sec::dense}
To provide holistic semantic guidance for pre-trained T2I diffusion models, we propose Dense Semantic Guidance (DSG), which supplies fine-grained semantic cues that complement the high-level reasoning introduced by SLBP. Specifically, DSG utilizes a segmentation mask $M^s \in \mathbb{R}^{H \times W \times 3}$ to deliver pixel-level semantic priors. In practice, the segmentation mask $M^s$ is represented in RGB format, where pixels belonging to the same category are labeled with a consistent color. 
To further enrich DSG, we introduce a novel representation, the \textit{Segmentation-CLIP Map} (SCMap), which preserves the spatial structure of the segmentation mask while incorporating the rich semantic information from CLIP embeddings. Given a semantic segmentation model $f_\theta$, a CLIP~\cite{clip} text encoder $g_\phi$, and the LR image $x \in \mathbb{R}^{H \times W \times 3}$, the SCMap $M^{sc} \in \mathbb{R}^{H \times W \times 1024}$ is derived by converting each pixel's semantic label into CLIP text embeddings, expressed as:
\begin{equation}
    M^{sc}_{i,j} = g_\phi(t(M_{i,j})), \; \forall i, j,
\label{eq:scmap}
\end{equation}%
where $M = f_\theta(x) \in \mathbb{R}^{H \times W}$ denotes the semantic segmentation result, which contains the semantic label index of each pixel, and $(i,j)$ represents pixel coordinates. The function $t(\cdot)$ converts a semantic label index into the corresponding class name in plain text. These dense guidance maps serve different but complementary roles: $M^s$ emphasizes structural details of the objects present in the scene, while $M^{sc}$ provides richer semantic embeddings that help interpret the complex image content. A detailed explanation of their properties is provided in Section~\ref{sec::ablation_gfm}. 

Additionally, to optimize the efficiency when utilizing SCMap, we introduce an additional encoder to compress the channel dimension of SCMap from 1024 to 128, providing a more compact representation for subsequent utilization. 
Overall, DSG integrates the segmentation mask with the proposed SCMap, providing localized, fine-grained semantic priors. This approach enhances the ability of T2I diffusion models to synthesize high-quality images with fine texture and semantically accurate content, as shown in Fig.~\ref{fig:ablation_dsg}.
\kelvin{Refer to the examples showing that.}%
\yuan{11/14 fixed}

\subsection{Guidance Fusion Module}
\label{sec::gfm}
With the segmentation mask $M^s$ and the SCMap $M^{sc}$ as dense semantic guidance, it is crucial to have an architecture that effectively leverages their semantic priors to guide the synthesis of object details. To achieve this, we propose the Guidance Fusion Module (GFM), which utilizes $M^s$ and $M^{sc}$ as conditioning inputs to refine an intermediate feature $\mathbf{F}$ from the diffusion model, as illustrated in Fig.~\ref{fig:framework}. Specifically, GFM consists of two Semantic-Adaptive Feature Transformation (SAFT) blocks, adapted from the SPADE~\cite{spade} block, which predict pixel-wise scale and shift parameters to apply affine transformations on the input feature. 
The two SAFT blocks $h_\Phi$ and $h_{\Phi'}$ are conditioned on $M^s$ and $M^{sc}$, respectively, enabling distinct transformations tailored to each type of guidance.

First, $M^s$ and $M^{sc}$ are processed through encoding layers $G_\Psi$ and $G_{\Psi'}$ in separate SAFT blocks to produce scale and shift parameter maps $\gamma, \beta \in \mathbb{R}^{H \times W \times 3}$, defined as:
\vspace{-5pt}
\begin{equation}
    \gamma^s, \beta^s = G_\Psi(M^s),  \quad  \gamma^{sc}, \beta^{sc} = G_{\Psi'}(M^{sc}),
\label{eq:gfm1}
\vspace{-5pt}
\end{equation}%
where $G_\Psi$ and $G_{\Psi'}$ consist of a sequence of 1$\times$1 convolutions, ensuring that $\gamma$ and $\beta$ maintain the spatial structure of $M^s$ and $M^{sc}$. This property allows SAFT blocks to apply distinct transformations to different semantic regions, enabling object-specific feature enhancement.
Then, $\gamma$ and $\beta$ are applied to the input feature $\mathbf{F}$ via Hadamard product for scaling and element-wise addition for shifting, yielding the transformed features $\mathbf{F}'_c$ and $\mathbf{F}'_{sc}$, given by:
\begin{align}
    \mathbf{F}'_c &= h_\Phi(\mathbf{F}, M^s) = \gamma^s \odot \mathbf{F} + \beta^s, \\
    \mathbf{F}'_{sc} &= h_{\Phi'}(\mathbf{F}, M^{sc}) = \gamma^{sc} \odot \mathbf{F} + \beta^{sc}.
\label{eq:gfm2}
\vspace{-5pt}
\end{align}%
Finally, the transformed representations $\mathbf{F}'_c$ and $\mathbf{F}'_{sc}$ are merged to produce the final output feature $\mathbf{F}'$, which feeds into the subsequent diffusion process, expressed as:
\vspace{-5pt}
\begin{equation}
    \mathbf{F}' = \mathbf{F}'_c + \mathbf{F}'_{sc}.
\label{eq:gfm3}
\vspace{-5pt}
\end{equation}%
This fusion operation ensures that the structural details from segmentation masks and contextual information from SCMap are well integrated, providing the diffusion model with holistic semantics for enhanced synthesis quality.
%
%

\section{Experiments}
\label{sec::experiments}
\subsection{Experimental Setup}
\label{sec::setup}
\paragraph{Implementation Details.}
\label{sec::implementation}
We employ the SD-2-base\footnote{https://huggingface.co/stabilityai/stable-diffusion-2-base} 
\ds{Why not SD3? }%
\yuan{I align this setting with our baseline SeeSR to ablate our component. Some people release the SD-turbo/SDXL version but for demo/tech report only}%
model as our base T2I model, along with the pre-trained DAPE~\cite{seesr} image encoder to generate LR embeddings for the attention layers. For semantic segmentation model, we employ Mask2Former~\cite{mask2former} with Swin-L~\cite{swint} backbone, pre-trained on the ADE20K~\cite{ade20k} dataset, which includes 150 object and stuff categories. These modules are interchangeable with contemporary models, and their weights are kept frozen during training. 

Our model is trained for 100,000 iterations using AdamW~\cite{adamw} optimizer, with the batch size and the learning rate set to 32 and 5$\times$$10^{-5}$, respectively. The training process is conducted on four NVIDIA RTX A6000 Ada GPUs for 30 hours, with conventional noise prediction objective~\cite{ddpm} in diffusion models. For inference, we use DDPM~\cite{ddpm} sampling with 50 timesteps.

\vspace{-6pt}
\paragraph{Datasets.}
\label{sec::datasets}
We follow the training settings of SeeSR~\cite{seesr}, using the LSDIR~\cite{lsdir} dataset and the first 10,000 images from FFHQ~\cite{stylagan}, with the degradation pipeline proposed by Real-ESRGAN~\cite{realesrgan} to generate LR-HR training pairs.

We assess our method on both synthetic and real-world datasets, including DIV2K~\cite{div2k}, RealSR~\cite{realsr}, DrealSR~\cite{drealsr}, and DPED-iPhone~\cite{dped-iphone}. We employ the testing data provided by StableSR\footnote{https://huggingface.co/datasets/Iceclear/StableSR-TestSets}~\cite{stablesr} for evaluation.
\kelvin{You follow the settings or you use their data.}%
\yuan{11/14 fixed}%
Specifically, they randomly crop images in DIV2K validation set into 3,000 patches, then apply the degradation pipeline of Real-ESRGAN to synthesize LR-HR pairs. For RealSR, DrealSR and DPED-iPhone, center-cropping is adopted. Note that the resolution of LR and HR images is set to 128 $\times$ 128 and 512 $\times$ 512, respectively. 

\vspace{-6pt}
\paragraph{Evaluation Metrics.}
\label{sec::metrics}
We employ widely used metrics to evaluate our method. To assess image fidelity, we utilize PSNR and SSIM~\cite{ssim}, with their values calculated in the YCbCr color space. 
\ds{Explain why PSNR/SSIM may not be good metrics or should be given less weights.}%
\yuan{Provide in quantitative comparison}%
For perceptual quality, we adopt both reference-based and non-reference metrics, including LPIPS~\cite{lpips},  MUSIQ~\cite{musiq}, MANIQA~\cite{maniqa}, and CLIPIQA~\cite{clipiqa}. Note that the DPED-iPhone dataset relies solely on non-reference metrics for evaluation due to the absence of ground truth images. For our evaluations, we utilize the public IQA-PyTorch toolbox\footnote{https://github.com/chaofengc/IQA-PyTorch} and follow their default settings to produce the final scores. Note that following the common practice of border-shaving~\cite{edsr}, we set the value of \textit{crop\_border} to 4 when measuring PSNR and SSIM. The evaluation code is available at: \url{https://github.com/liyuantsao/SR-IQA}.

\subsection{Comparison with State-of-the-art Methods}
\paragraph{Settings.}
We compare our approach against three categories of Real-ISR methods: GAN-based methods, including Real-ESRGAN~\cite{realesrgan}, BSRGAN~\cite{bsrgan}, and SwinIR-GAN~\cite{swinir}; diffusion-based methods, such as ResShift~\cite{resshift} and SinSR~\cite{sinsr}; and diffusion prior-based methods, including StableSR~\cite{stablesr}, DiffBIR~\cite{diffbir}, SeeSR~\cite{seesr}, and OSEDiff~\cite{osediff}. We use the publicly released codes and pre-trained models for each method. For a fair comparison, \textbf{all the models are evaluated without manually added prompts} (\eg, ``clean'', ``sharp'') during inference to ensure that we compare the inherent capabilities of each model.

\vspace{-5pt}
\paragraph{Quantitative Comparison.}
\begin{table*}[t]
\renewcommand{\arraystretch}{1.25}
\newcommand{\mytoprule}{\toprule[1.5pt]}
\newcommand{\mybottomrule}{\bottomrule[1.5pt]}
\newcommand{\first}[1]{\textcolor{red}{\underline{#1}}}
\newcommand{\second}[1]{\textcolor{blue}{#1}}
\caption{Quantitative comparison of Real-ISR methods on both synthetic~\cite{div2k} and real-world~\cite{realsr, drealsr, dped-iphone} datasets. The best and second-best results are highlighted in \first{red}, and \second{blue}, respectively. The proposed \ourmethod{} achieves leading performance across multiple non-reference metrics, demonstrating favorable visual quality against contemporary frameworks.}
\centering
\rowcolors{2}{gray!15}{white}
\resizebox{2.0\columnwidth}{!}{
    \begin{tabular}{llcccccccccc}
    \toprule
    Dataset & Metric & Real-ESRGAN & BSRGAN & SwinIR-GAN & ResShift & SinSR & StableSR & DiffBIR & SeeSR & OSEDiff & \textbf{{\ourmethod}} \\
    %
     \midrule
     \cellcolor{white} & PSNR$\uparrow$ & 24.1546 & \first{24.7653} & \second{24.7252} & 24.6561 & 24.5333 & 23.1801 & 23.9906 & 24.5315 & 23.6114 & 23.5303 \\
     \cellcolor{white} & SSIM$\uparrow$ & 0.7366 & \second{0.7404} & \first{0.7489} & 0.7326 & 0.7067 & 0.6804 & 0.6364 & 0.7271 & 0.7079 & 0.6883 \\
     \cellcolor{white} & LPIPS$\downarrow$ & 0.2710 & \second{0.2656} & \first{0.2539} & 0.3133 & 0.3214 & 0.3002 & 0.3618 & 0.2824 & 0.2921 & 0.3165 \\
     \multirow{-2}{*}{\cellcolor{white}\textit{RealSR}} & MUSIQ$\uparrow$ & 60.3695 & 63.2870 & 58.6931 & 59.7996 & 60.5363 & 65.8833 & 63.9985 & 64.0156 & \second{69.0987} & \first{69.7217} \\
     \cellcolor{white} & MANIQA$\uparrow$ & 0.3743 & 0.3792 & 0.3466 & 0.3893 & 0.3962 & 0.4281 & \second{0.4966} & 0.4557 & 0.4740 & \first{0.5741} \\
     \cellcolor{white} & CLIPIQA$\uparrow$ & 0.4494 & 0.5116 & 0.4363 & 0.5377 & 0.6094 & 0.6234 & 0.6409 & 0.5619 & \second{0.6680} & \first{0.7109} \\
     %
     \midrule
     \cellcolor{white} & PSNR$\uparrow$ & 26.3053 & \second{26.4506} & 26.2644 & 26.0362 & 25.8743 & 25.9333 & 24.9134 & \first{26.8408} & 25.9387 & 25.4919 \\
     \cellcolor{white} & SSIM$\uparrow$ & \second{0.7769} & 0.7743 & 0.7759 & 0.7598 & 0.7144 & 0.7229 & 0.6123 & \first{0.7770} & 0.7552 & 0.7126 \\
     \cellcolor{white} & LPIPS$\downarrow$ & \second{0.2819} & 0.2858 & \first{0.2743} & 0.3503 & 0.3722 & 0.3284 & 0.4591 & 0.2982 & 0.2967 & 0.3682 \\
     \multirow{-2}{*}{\cellcolor{white}\textit{DrealSR}} & MUSIQ$\uparrow$ & 54.2680 & 57.1647 & 52.7359 & 52.4267 & 55.0289 & 58.5118 & 60.9892 & 57.0266 & \second{64.6679} & \first{65.9168} \\
     \cellcolor{white} & MANIQA$\uparrow$ & 0.3449 & 0.3436 & 0.3301 & 0.3496 & 0.3833 & 0.3886 & \second{0.4969} & 0.4142 & 0.4696 & \first{0.5479} \\
     \cellcolor{white} & CLIPIQA$\uparrow$ & 0.4525 & 0.5092 & 0.4443 & 0.5478 & 0.6343 & 0.6356 & 0.6512 & 0.5867 & \second{0.6964} & \first{0.7138} \\
    %
    \midrule
    \cellcolor{white} & PSNR$\uparrow$ & 22.6408 & \first{22.8354} & 22.2512 & \second{22.7583} & 22.5402 & 21.6477 & 21.8460 & 22.7328 & 22.0930 & 21.8953 \\
    \cellcolor{white} & SSIM$\uparrow$ & \first{0.5987} & 0.5907 & 0.5894 & 0.5889 & 0.5672 & 0.5344 & 0.5027 & \second{0.5971} & 0.5738 & 0.5551 \\
    \cellcolor{white} & LPIPS$\downarrow$ & \second{0.3112} & 0.3351 & 0.3160 & 0.3404 & 0.3245 & 0.3113 & 0.3786 & 0.3348 & \first{0.2942} & 0.3211 \\
    \multirow{-2}{*}{\cellcolor{white}\textit{DIV2K-Val}} & MUSIQ$\uparrow$ & 61.0583 & 61.1959 & 60.2172 & 60.8121 & 62.8263 & 65.9177 & 66.4712 & 61.1407 & \second{67.9671} & \first{68.8999} \\
    \cellcolor{white} & MANIQA$\uparrow$ & 0.3819 & 0.3561 & 0.3655 & 0.3992 & 0.4251 & 0.4210 & \second{0.5243} & 0.4105 & 0.4436 & \first{0.5427} \\
    \cellcolor{white} & CLIPIQA$\uparrow$ & 0.5277 & 0.5246 & 0.5338 & 0.5919 & 0.6490 & 0.6771 & \second{0.6846} & 0.5779 & 0.6681 & \first{0.7112} \\
    %
    \midrule
    & MUSIQ$\uparrow$ & 42.4330 & 45.8945 & 43.2885 & 44.0291 & 45.9643 & 50.4749 & 53.2033 & 49.7585 & \second{56.3688} & \first{57.3402} \\
    \cellcolor{white} \textit{DPED-iPhone} & MANIQA$\uparrow$ & 0.3047 & 0.3148 & 0.3017 & 0.3165 & 0.3521 & 0.3433 & \second{0.4419} & 0.3769 & 0.4051 & \first{0.4881} \\
    & CLIPIQA$\uparrow$ & 0.3380 & 0.4022 & 0.3373 & 0.4701 & 0.5488 & 0.4799 & 0.5280 & 0.4848 & \second{0.5934} & \first{0.6191} \\
    \bottomrule
    \end{tabular}
}
\label{tab:main}
\end{table*}
\label{sec::quantative}
Table~\ref{tab:main} presents a quantitative comparison between the proposed {\ourmethod} and contemporary Real-ISR approaches. {\ourmethod} consistently achieves leading results across the three non-reference perceptual quality metrics: MUSIQ, MANIQA, and CLIPIQA. Specifically, on the RealSR dataset, {\ourmethod} achieves a 15.6\% increase in MANIQA and a 10.9\% improvement in CLIPIQA over the second-best results. Similarly, on the DrealSR dataset, {\ourmethod} gain a 10.3\% enhancement in MANIQA, with similar performance trends across all these four evaluation sets. These results highlight the capability of {\ourmethod} in generating perceptually realistic images. 

\begin{figure*}[ht]
    \centering
    
    \vspace{1mm}
    \begin{subfigure}[b]{0.12\linewidth}
        \vtop{\centering
        \includegraphics[width=\linewidth]{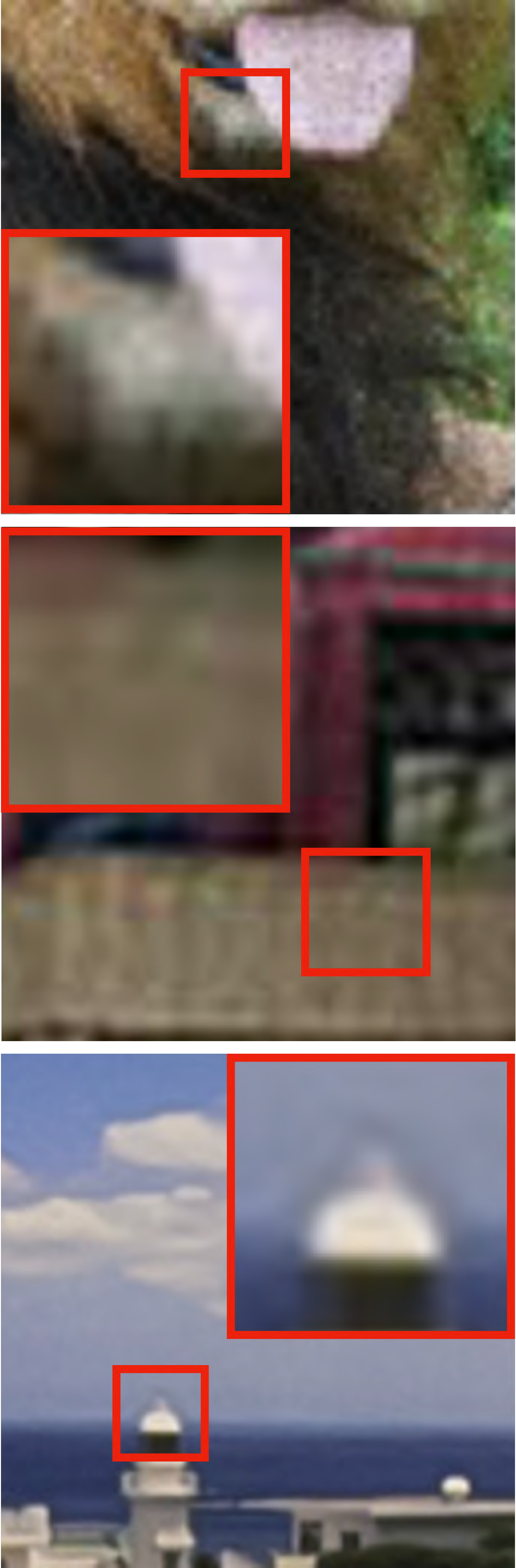}
        \caption*{\centering \scriptsize LR Image}}
    \end{subfigure}
    \begin{subfigure}[b]{0.12\linewidth}
        \vtop{\centering
        \includegraphics[width=\linewidth]{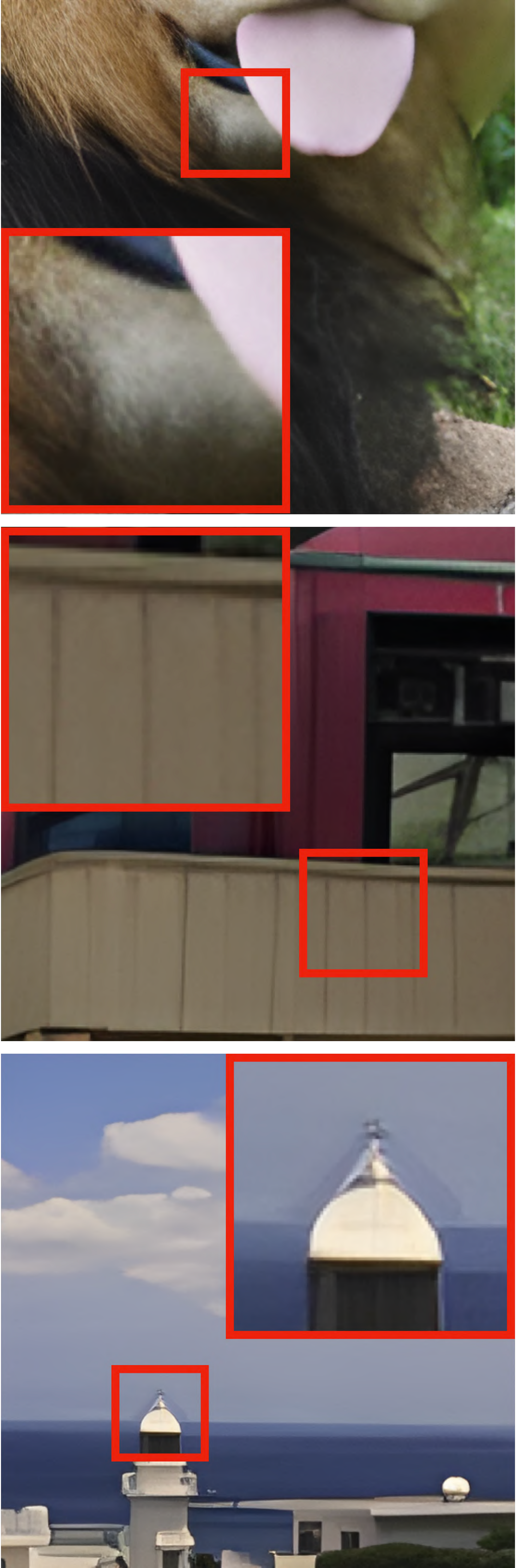}
        \caption*{\centering \scriptsize Real-ESRGAN}}
    \end{subfigure}
    \begin{subfigure}[b]{0.12\linewidth}
        \vtop{\centering
        \includegraphics[width=\linewidth]{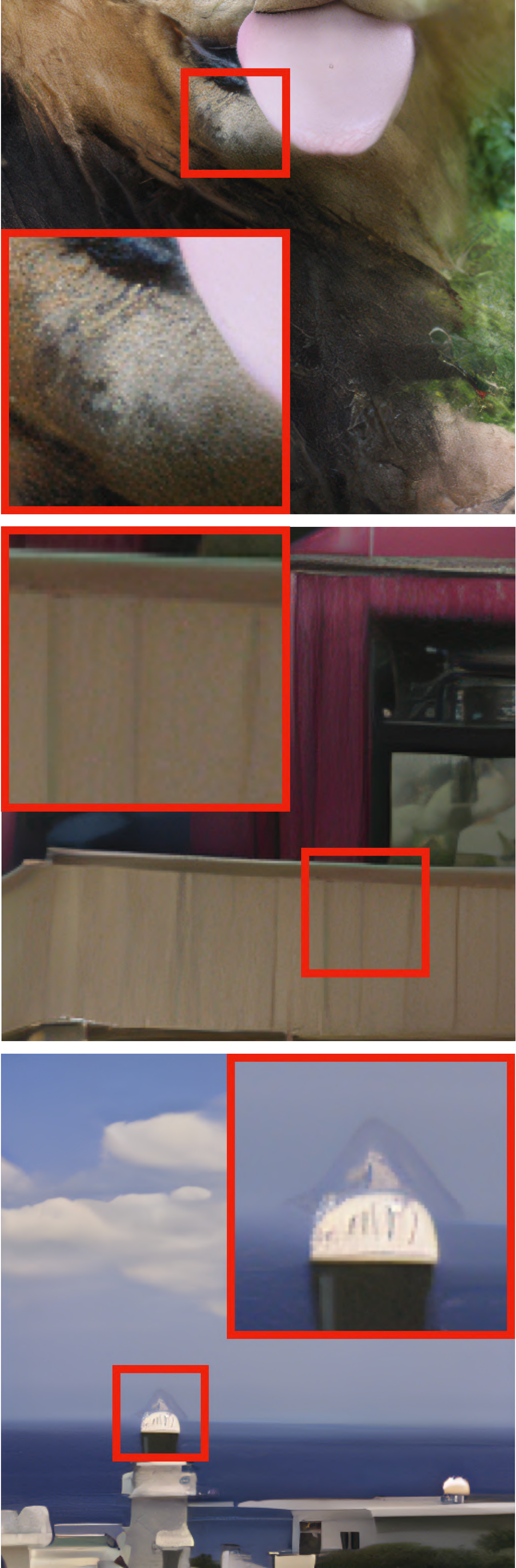}
        \caption*{\centering \scriptsize ResShift}}
    \end{subfigure}
    \begin{subfigure}[b]{0.12\linewidth}
        \vtop{\centering
        \includegraphics[width=\linewidth]{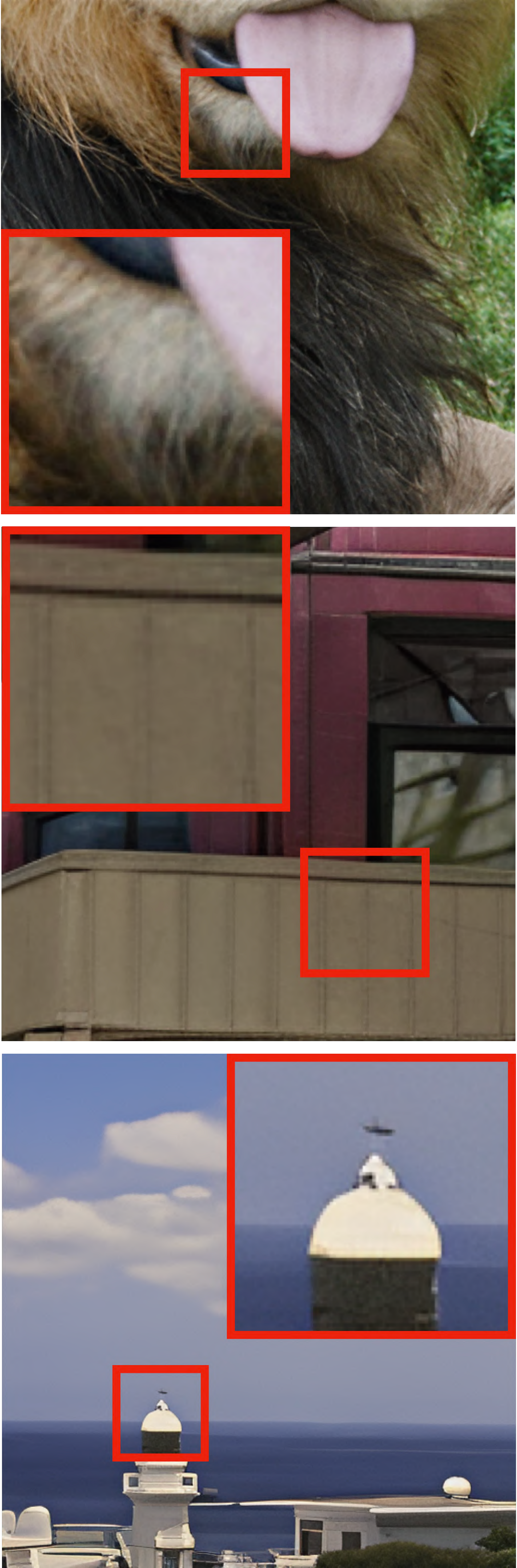}
        \caption*{\centering \scriptsize StableSR}}
    \end{subfigure}
    \begin{subfigure}[b]{0.12\linewidth}
        \vtop{\centering
        \includegraphics[width=\linewidth]{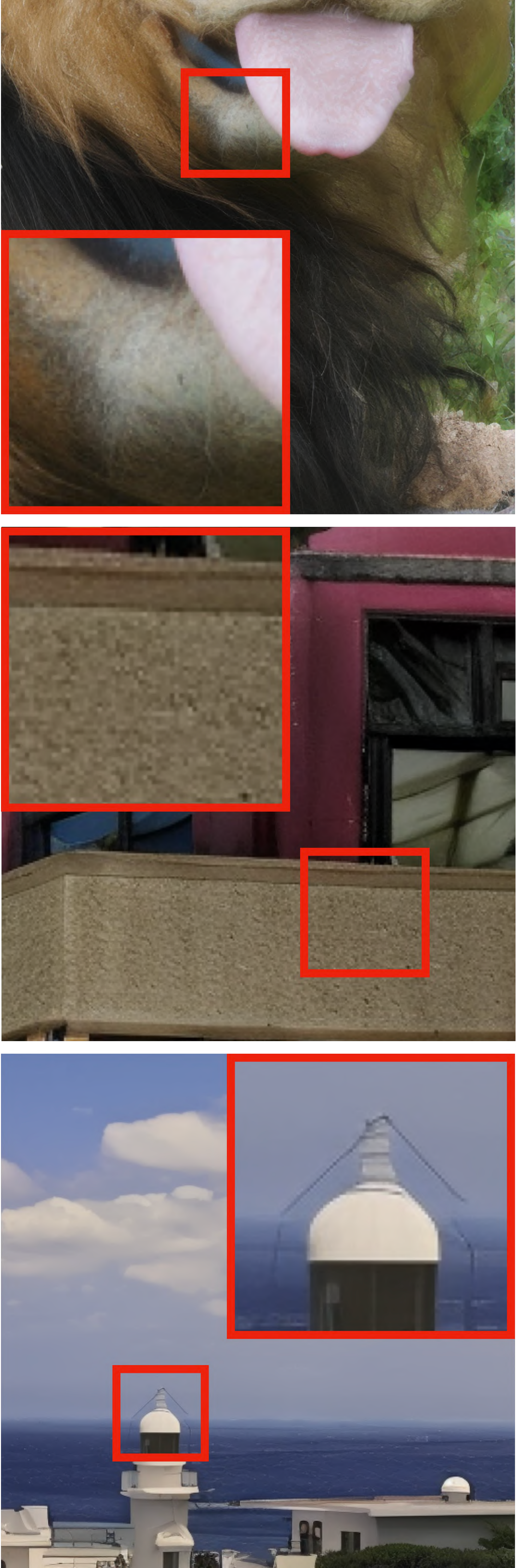}
        \caption*{\centering \scriptsize DiffBIR}}
    \end{subfigure}
    \begin{subfigure}[b]{0.12\linewidth}
        \vtop{\centering
        \includegraphics[width=\linewidth]{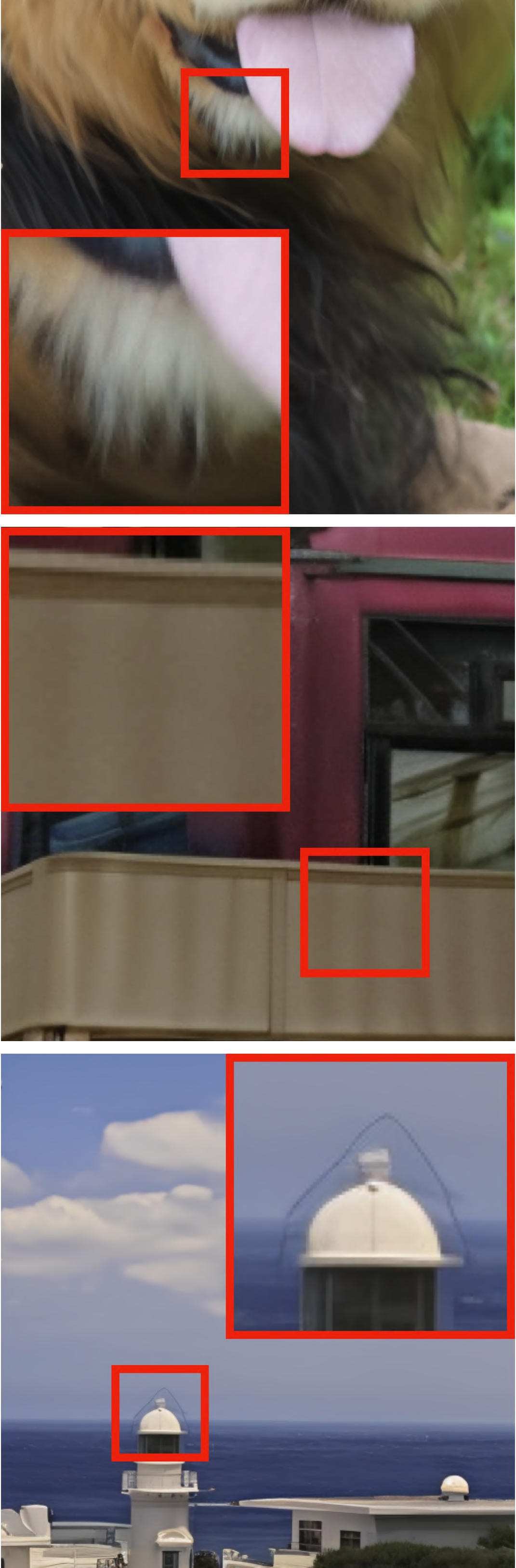}
        \caption*{\centering \scriptsize SeeSR}}
    \end{subfigure}
    \begin{subfigure}[b]{0.12\linewidth}
        \vtop{\centering
        \includegraphics[width=\linewidth]{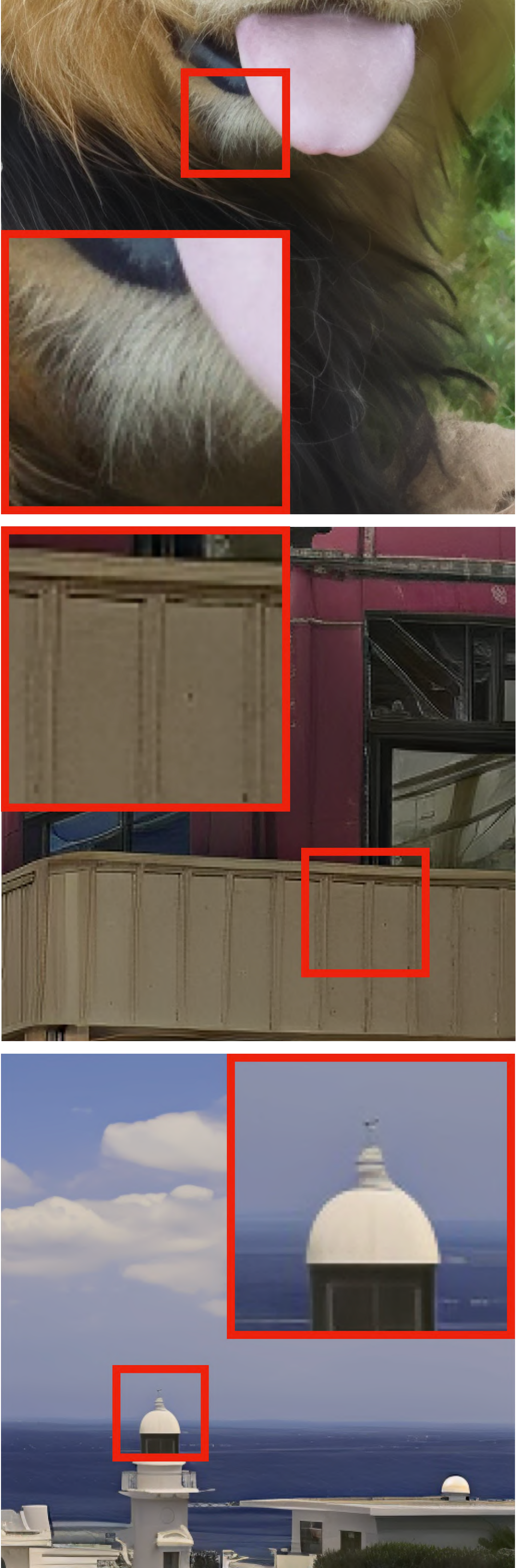}
        \caption*{\centering \scriptsize \textbf{{\ourmethod}}}}
    \end{subfigure}
    \begin{subfigure}[b]{0.12\linewidth}
        \vtop{\centering
        \includegraphics[width=\linewidth]{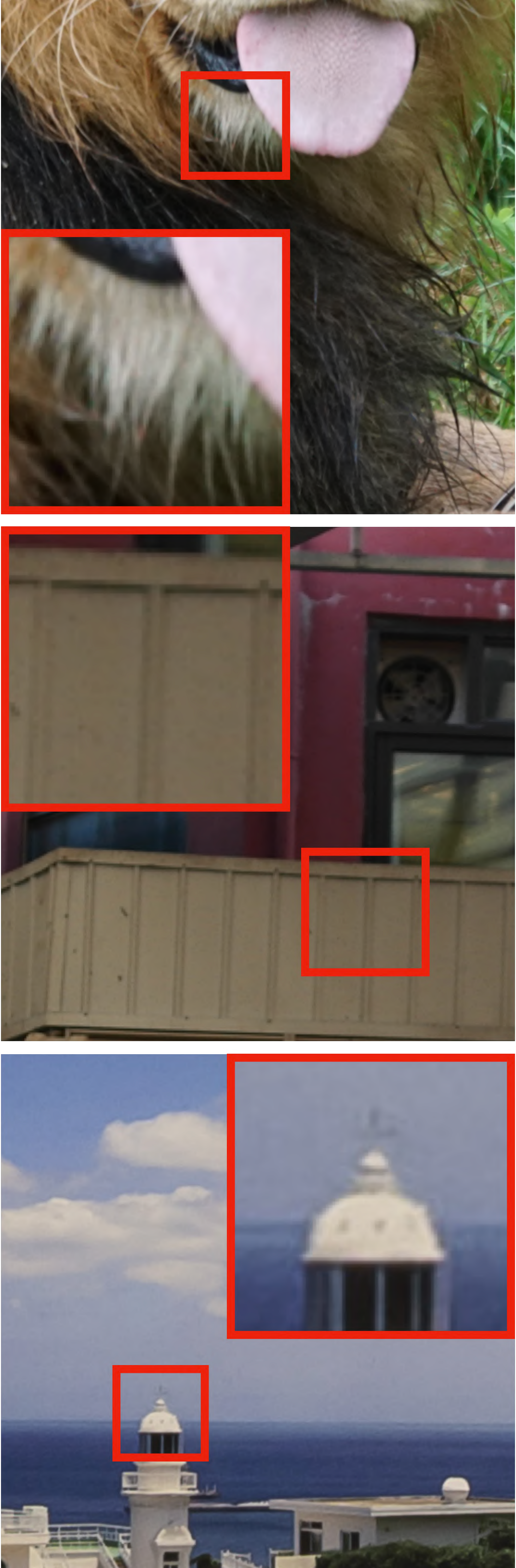}
        \caption*{\centering \scriptsize Ground Truth}}
    \end{subfigure}
    
    \caption{Qualitative comparison between the proposed {\ourmethod} and contemporary Real-ISR methods. \ourmethod{} presents sharper details without introducing noticeable visual artifacts across various Real-ISR scenarios.}
    \label{fig:qualitative}
\end{figure*}

For reference-based metrics: PSNR, SSIM and LPIPS, {\ourmethod} demonstrates competitive performance among diffusion prior-based methods, indicating that this framework produces visually plausible results while maintaining a reasonable level of fidelity. Notably, as shown in Table~\ref{tab:main}, GAN-based methods excel in reference-based metrics, while diffusion prior-based methods perform well on non-reference metrics. This difference arises from the perception-distortion trade-off~\cite{tradeoff}, where reference-based metrics tend to favor smooth texture. While diffusion prior-based models show fewer advantages in reference-based metrics, they often produce sharper and visually appealing content, as the outputs of our \ourmethod{} depicted in Fig.~\ref{fig:qualitative}. 
\ds{Explain why GAN-based methods are much better on PSNR/SSIM.}%
\yuan{11/15 fixed}%

\vspace{-8pt}
\paragraph{Qualitative Comparison.}
\label{sec::qualitative}
Fig.~\ref{fig:qualitative} presents the qualitative comparison between the proposed {\ourmethod} and contemporary Real-ISR frameworks. Across various challenging scenarios, {\ourmethod} performs well in recovering fine textures, structures, and high-frequency patterns. In the first row, for instance, {\ourmethod} faithfully restores intricate fur details, while other methods produce blurred or distorted textures. 
In the second row, {\ourmethod} preserves clear and sharp building structures, while other approaches introduce artifacts or do not capture the fine patterns accurately. In the final row, {\ourmethod} produces a clear boundary between the lighthouse and its background, effectively preventing artifacts that could arise from the shadow effects present in the LR image. These qualitative results underscore the effectiveness of {\ourmethod} in producing perceptually realistic and polished restoration results. More qualitative results are available in the supplementary material.

\subsection{Ablation Studies}
In this section, we present ablation studies with in-depth analyses for the design of {\ourmethod}. For evaluation, we use the RealSR~\cite{realsr} testing set and employ both fidelity-oriented and perceptual-oriented metrics, including PSNR, LPIPS~\cite{lpips}, MANIQA~\cite{maniqa}, and CLIPIQA~\cite{clipiqa}, for comprehensive assessments. Notably,  we utilize both the reference-based metric LPIPS, and non-reference metrics MANIQA and CLIPIQA for evaluating visual quality, validating if the proposed framework generates realistic images without introducing excessive hallucinations.

\begin{figure}[t]
    \centering
    \begin{subfigure}[b]{0.32\linewidth}
        \vtop{\centering
        \includegraphics[width=\linewidth]{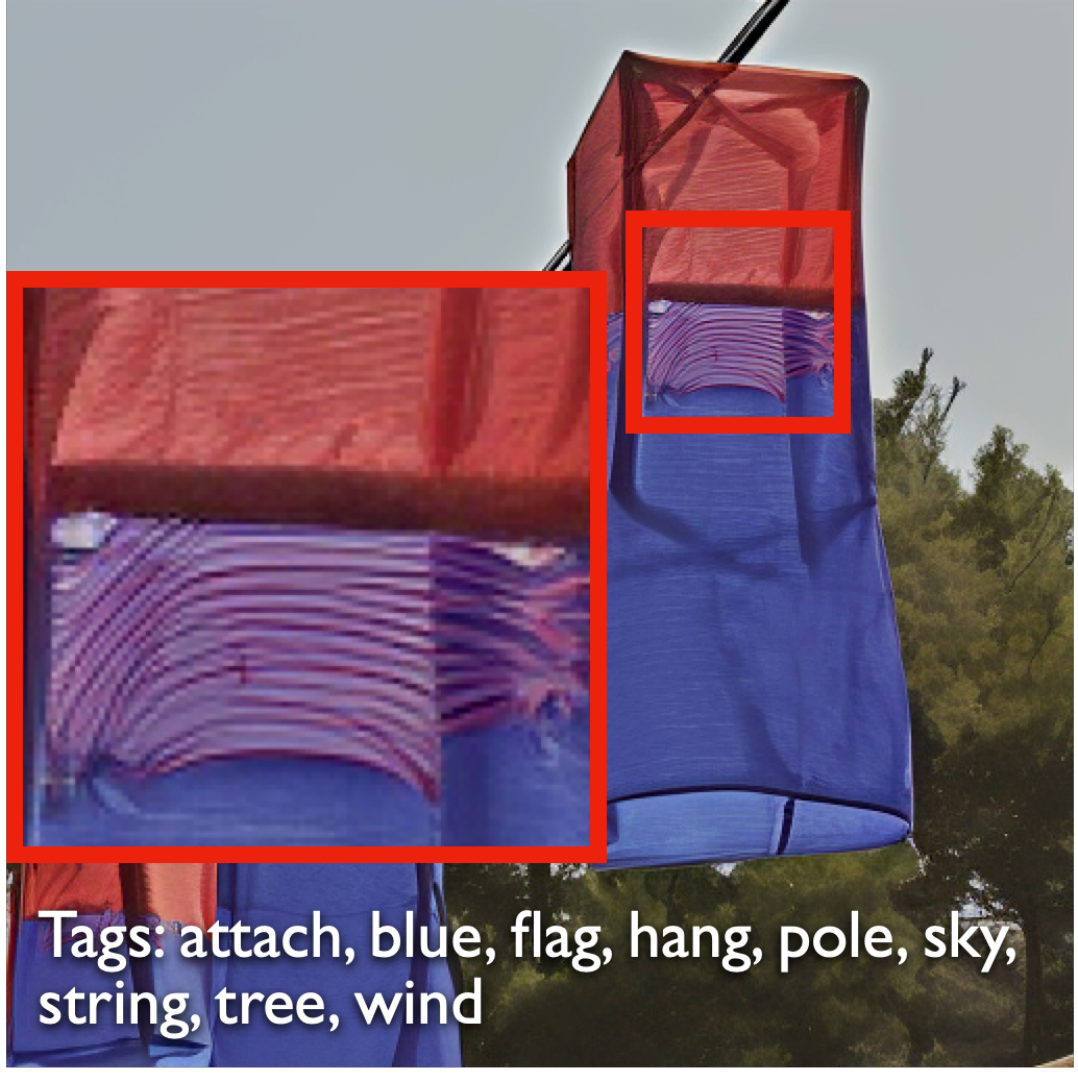}
        \caption*{\centering \scriptsize {\ourmethod} (+T)}}
    \end{subfigure}
    \begin{subfigure}[b]{0.32\linewidth}
        \vtop{\centering
        \includegraphics[width=\linewidth]{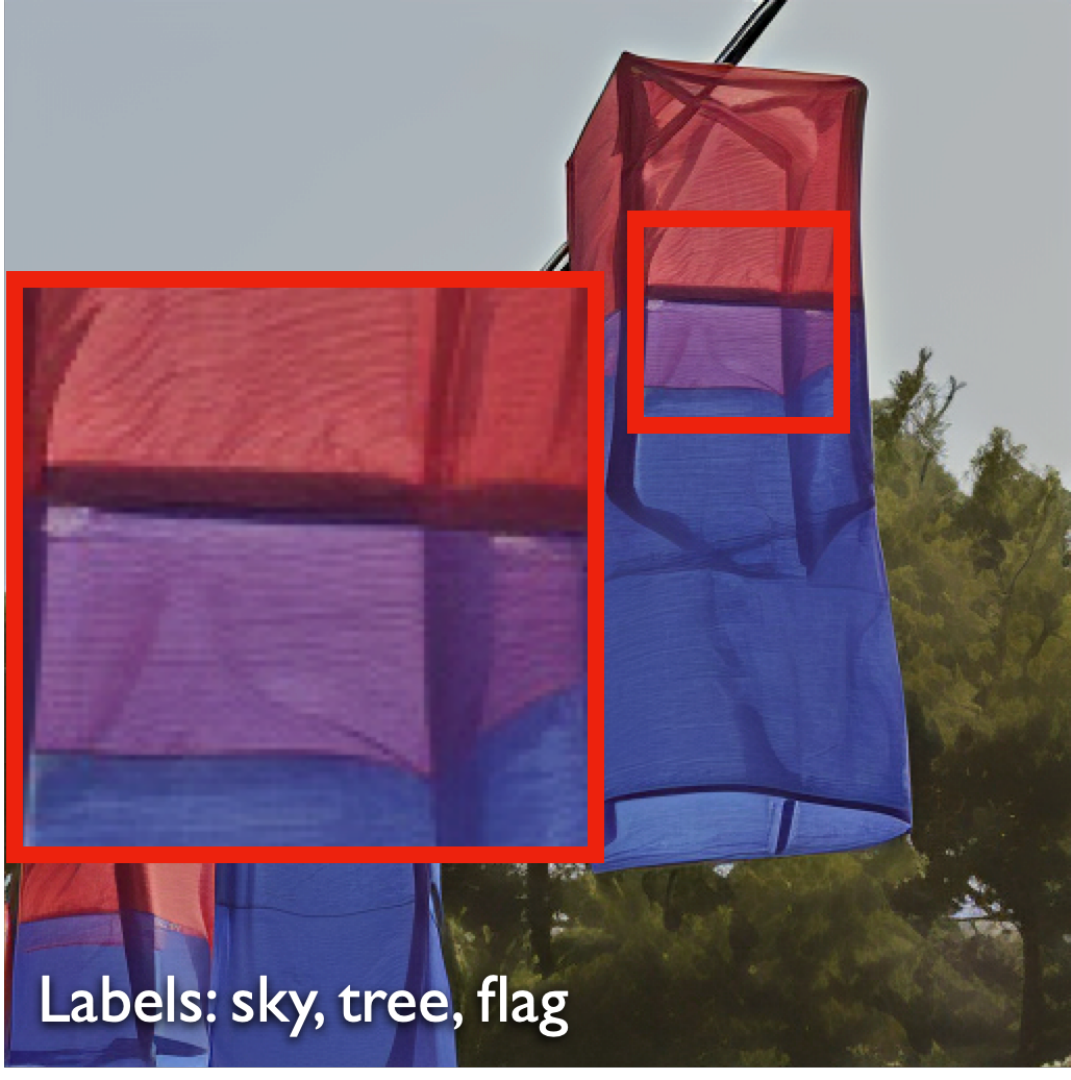}
        \caption*{\centering \scriptsize \textcolor{red}{{\ourmethod} (+L)}}}
    \end{subfigure}
    \begin{subfigure}[b]{0.32\linewidth}
        \vtop{\centering
        \includegraphics[width=\linewidth]{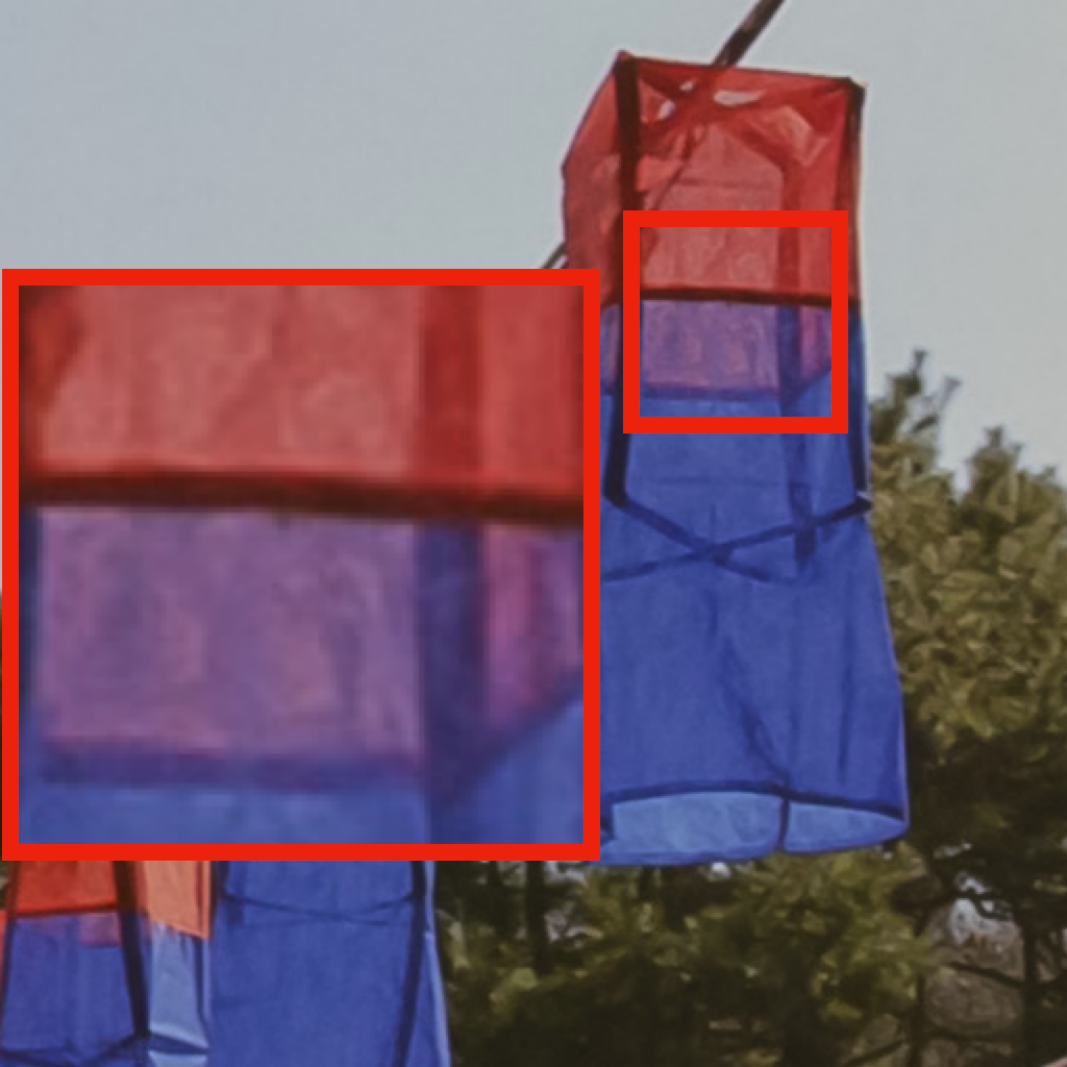}
        \caption*{\centering \scriptsize Ground Truth}}
    \end{subfigure}
    \caption{Qualitative comparison between using image tags~\cite{seesr} as prompts (+T) and employing the proposed Semantic Label-Based Prompting (+L) in \ourmethod{}. The proposed prompting scheme provides more concise description, reducing visual artifacts.}
    \label{fig:ablation_slbp}
\end{figure}

\vspace{-6pt}
\paragraph{Semantic Label-Based Prompting.}
\label{sec::ablation_slbp}
We evaluate the effectiveness of Semantic Label-Based Prompting (SLBP) against using image tags~\cite{seesr} as text prompts to guide the proposed {\ourmethod}. Note that we do not include experiments utilizing descriptions from LLMs (\eg, LLaVa) due to the computational infeasibility of querying LLMs during training.
\kelvin{How about those using LLM?}%
\yuan{11/15 fixed}%
As depicted in Fig.~\ref{fig:ablation_slbp}, SLBP generates text prompts including ``sky'', ``tree'', and ``flag'', which closely align with the image content. In contrast, the descriptions provided by image tags are noisy, containing terms irrelevant to the objects, such as ``wind'' or ``blue'', resulting in unwanted artifacts in the output.
Table~\ref{tab:ablation_tag} presents the quantitative assessment on the efficacy of SLBP, which demonstrates consistent improvements on both fidelity and perceptual quality compared to employing image tags as text prompts. Specifically, SLBP achieves a PSNR gain from 22.9629 to 23.5303, emphasizing its impact on enhancing fidelity. By concentrating on essential elements and avoiding irrelevant details, SLBP reduces artifacts and improves the overall quality of the super-resolved images.
\begin{table}[t]
\renewcommand{\arraystretch}{1.25}
\newcommand{\first}[1]{\textcolor{red}{\underline{#1}}}
\newcommand{\second}[1]{\textcolor{blue}{#1}}
\centering
\caption{Quantitative comparison between using image tags~\cite{seesr} as prompts (+T) and employing the proposed Semantic Label-Based Prompting (+L) in \ourmethod{}. The best results are highlighted in \first{red}. The proposed Semantic Label-Based Prompting enables {\ourmethod} to achieve improved fidelity and visual quality.}
\resizebox{1\columnwidth}{!}{%
    \begin{tabular}{lcccc}
    \toprule
    Method & PSNR$\uparrow$ & LPIPS$\downarrow$ & 
    MANIQA$\uparrow$ & CLIPIQA$\uparrow$ \\
    \midrule 
    {\ourmethod} (+T) & 22.9629 & 0.3209 & 0.5717 & 0.7007 \\
    \textbf{{\ourmethod} (+L)} & \first{23.5303} & \first{0.3165} & \first{0.5741} & \first{0.7109} \\
    \bottomrule
    \end{tabular}
}
\label{tab:ablation_tag}
\end{table}

\vspace{-6pt}
\paragraph{Dense Semantic Guidance.}
\label{sec::ablation_gfm}
\begin{table}[t]
\renewcommand{\arraystretch}{1.25}
\newcommand{\mytoprule}{\toprule[1pt]}
\newcommand{\mybottomrule}{\bottomrule[1pt]}
\newcommand{\first}[1]{\textcolor{red}{\underline{#1}}}
\newcommand{\second}[1]{\textcolor{blue}{#1}}
\centering
\footnotesize
\caption{Ablation study of Guidance Fusion Module. ``-s'' and ``-c'' denote the removal of segmentation mask and SCMap conditions, respectively. The best results are highlighted in \first{red}, with the second-best in \second{blue}. By utilizing both representations, \ourmethod{} achieves the best visual quality.} 
\resizebox{1\columnwidth}{!}{%
    \begin{tabular}{lcccc}
    \toprule
    Method & PSNR$\uparrow$ & LPIPS$\downarrow$ & 
    MANIQA$\uparrow$ & CLIPIQA$\uparrow$ \\
    \midrule 
    {\ourmethod} (-s) & 23.4428 & 0.3194 & \second{0.5684} & \second{0.7081} \\
    {\ourmethod} (-c) & \first{23.7343} & \first{0.3043} & 0.5229 & 0.6811 \\
    \midrule
    \textbf{{\ourmethod}} & \second{23.5303} & \second{0.3165} & \first{0.5741} & \first{0.7109} \\
    \bottomrule
    \end{tabular}
}
\label{tab:ablation_gfm}
\end{table}
We evaluate the impact of each component in the proposed Dense Semantic Guidance (DSG) by selectively removing the segmentation mask and SCMap conditions in the Guidance Fusion Module (GFM). As illustrated in Table~\ref{tab:ablation_gfm}, the segmentation mask plays a vital role in achieving high fidelity, as evidenced by the leading PSNR score of 23.7343 when it is solely applied as condition. This demonstrates that segmentation mask provides essential structural information for maintaining image fidelity. On the other hand, using SCMap as guidance presents better perceptual quality, reflected in higher MANIQA and CLIPIQA values than applying segmentation mask. This finding indicates that SCMap offers rich contextual embeddings, allowing for in-depth interpretation of complex scene semantics. These results highlight the complementary roles of the segmentation mask and SCMap: \textit{The segmentation mask maintains structural integrity, while SCMap enhances semantic richness}. 
\begin{figure}[t]
    \centering
    \begin{subfigure}[b]{0.32\linewidth}
        \vtop{\centering
        \includegraphics[width=\linewidth]{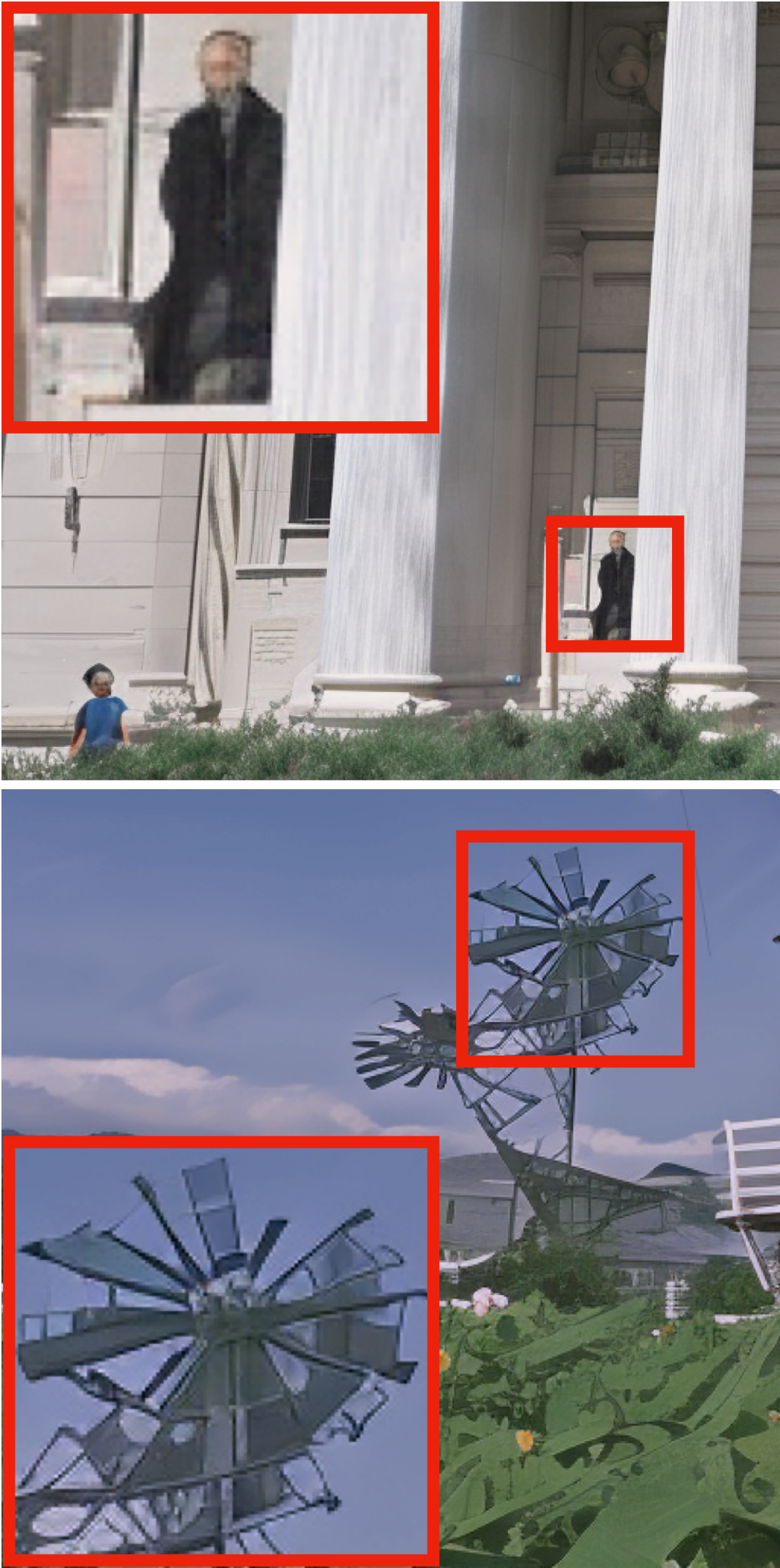}
        \caption*{\centering \scriptsize w/o DSG}}
    \end{subfigure}
    \begin{subfigure}[b]{0.32\linewidth}
        \vtop{\centering
        \includegraphics[width=\linewidth]{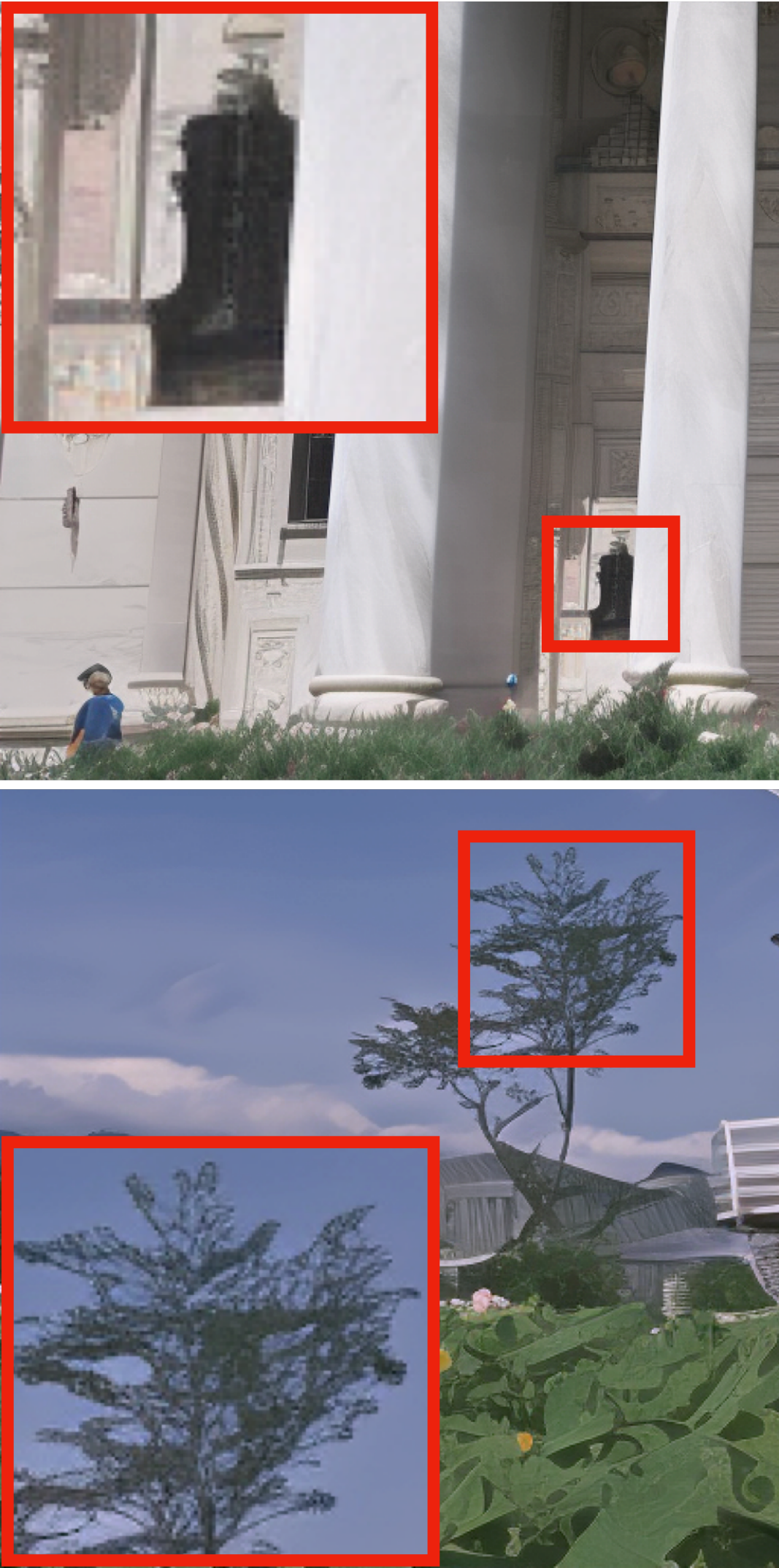}
        \caption*{\centering \scriptsize \textcolor{red}{w/ DSG}}}
    \end{subfigure}
    \begin{subfigure}[b]{0.32\linewidth}
        \vtop{\centering
        \includegraphics[width=\linewidth]{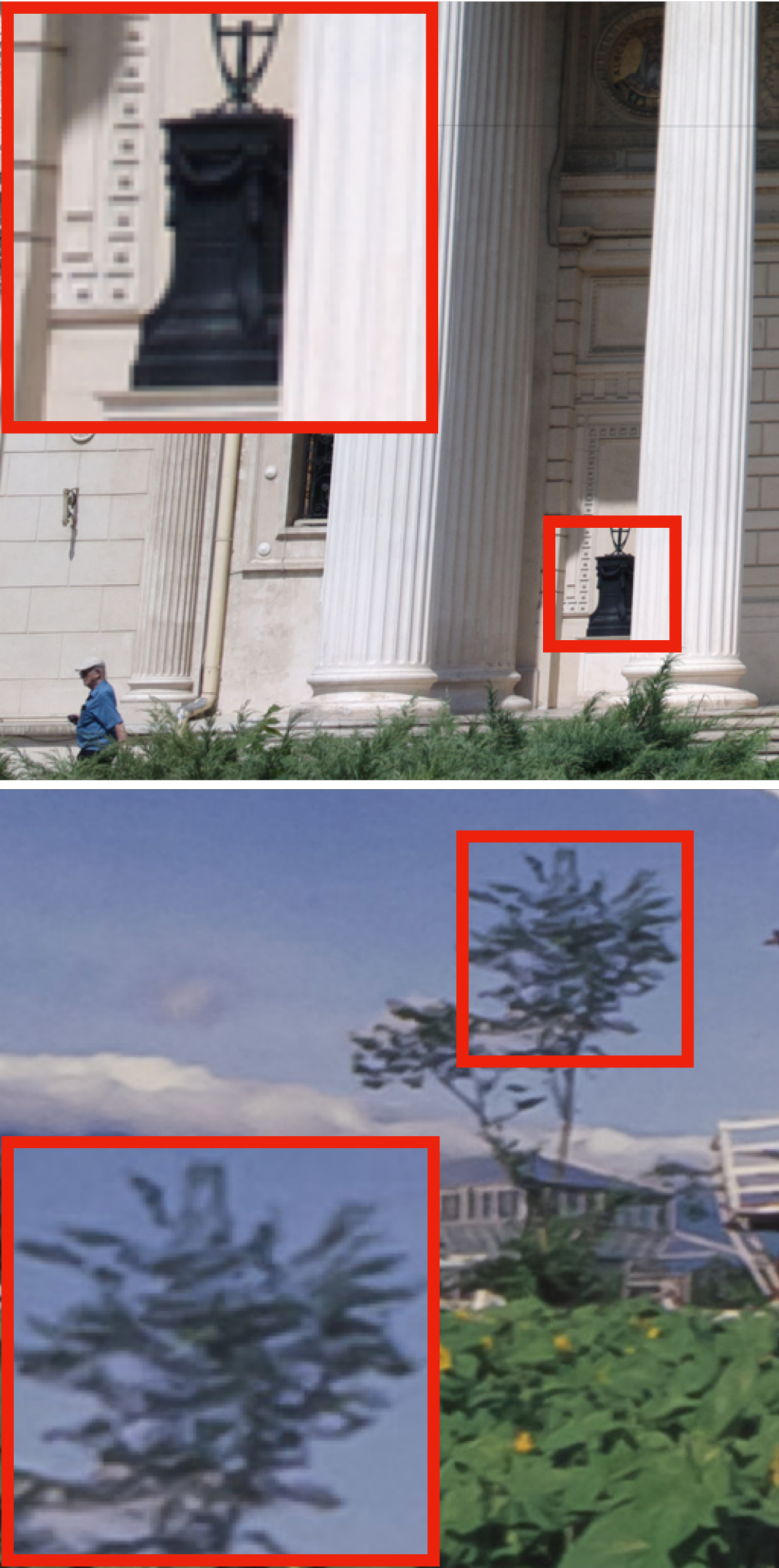}
        \caption*{\centering \scriptsize Ground Truth}}
    \end{subfigure}
    \vspace{-2pt}
    \caption{Illustrative examples presenting the effect of employing the proposed Dense Semantic Guidance (DSG). With the guidance from DSG, \ourmethod{} generates semantically correct content.}
    \label{fig:ablation_dsg}
    \vspace{-3pt}
\end{figure}

With complete configuration, the proposed DSG effectively conveys both forms of guidance, achieving favorable perceptual quality. In addition, Fig.~\ref{fig:ablation_dsg} presents a qualitative example demonstrating the impact of incorporating the DSG signals during generation. With rich semantics provided by the DSG, \ourmethod{} prevents hallucinations and accurately recovers the objects, validating the effectiveness of the proposed DSG for guiding T2I diffusion models.

\vspace{-6pt}
\paragraph{Influence of Semantic Priors on Texture Generation.}
\label{sec::ablation_prior}
\begin{figure}[t]
    \centering
    
    \vspace{1mm}
    \begin{subfigure}[b]{0.24\linewidth}
        \vtop{\centering
        \includegraphics[width=\linewidth]{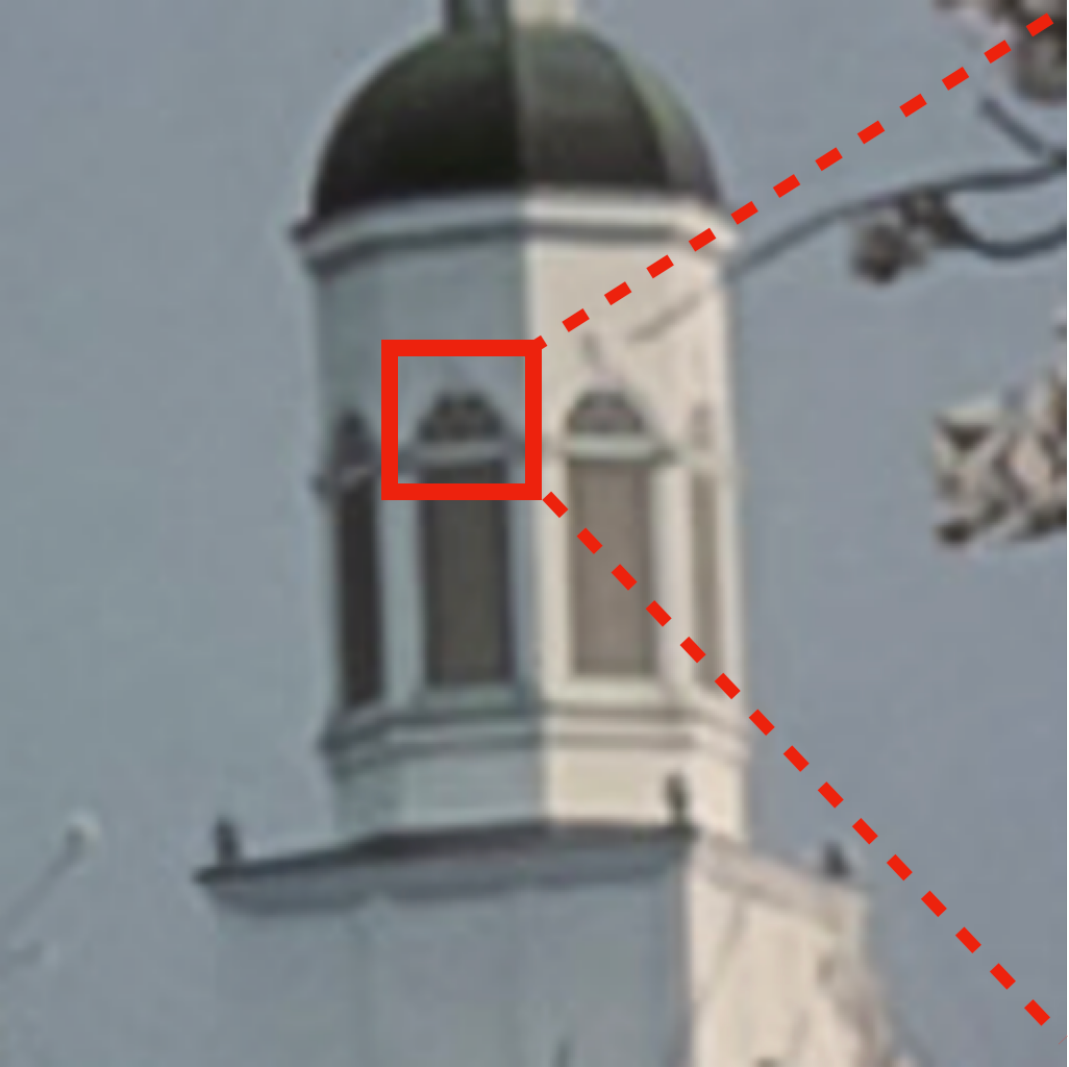}
        \caption*{\centering \scriptsize LR Image}}
    \end{subfigure}
    \begin{subfigure}[b]{0.24\linewidth}
        \vtop{\centering
        \includegraphics[width=\linewidth]{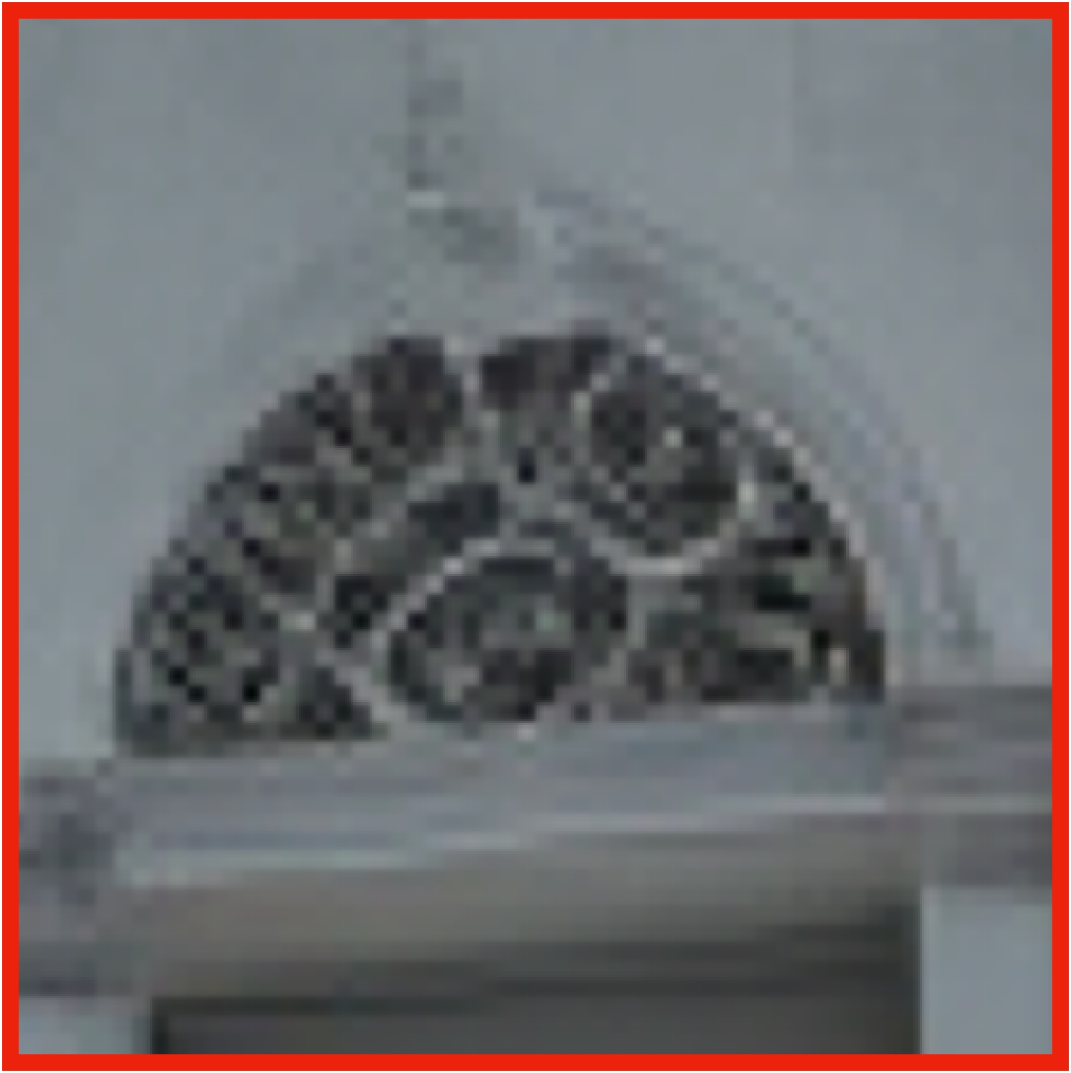}
        \caption*{\centering \scriptsize w/o Prior}}
    \end{subfigure}
    \begin{subfigure}[b]{0.24\linewidth}
        \vtop{\centering
        \includegraphics[width=\linewidth]{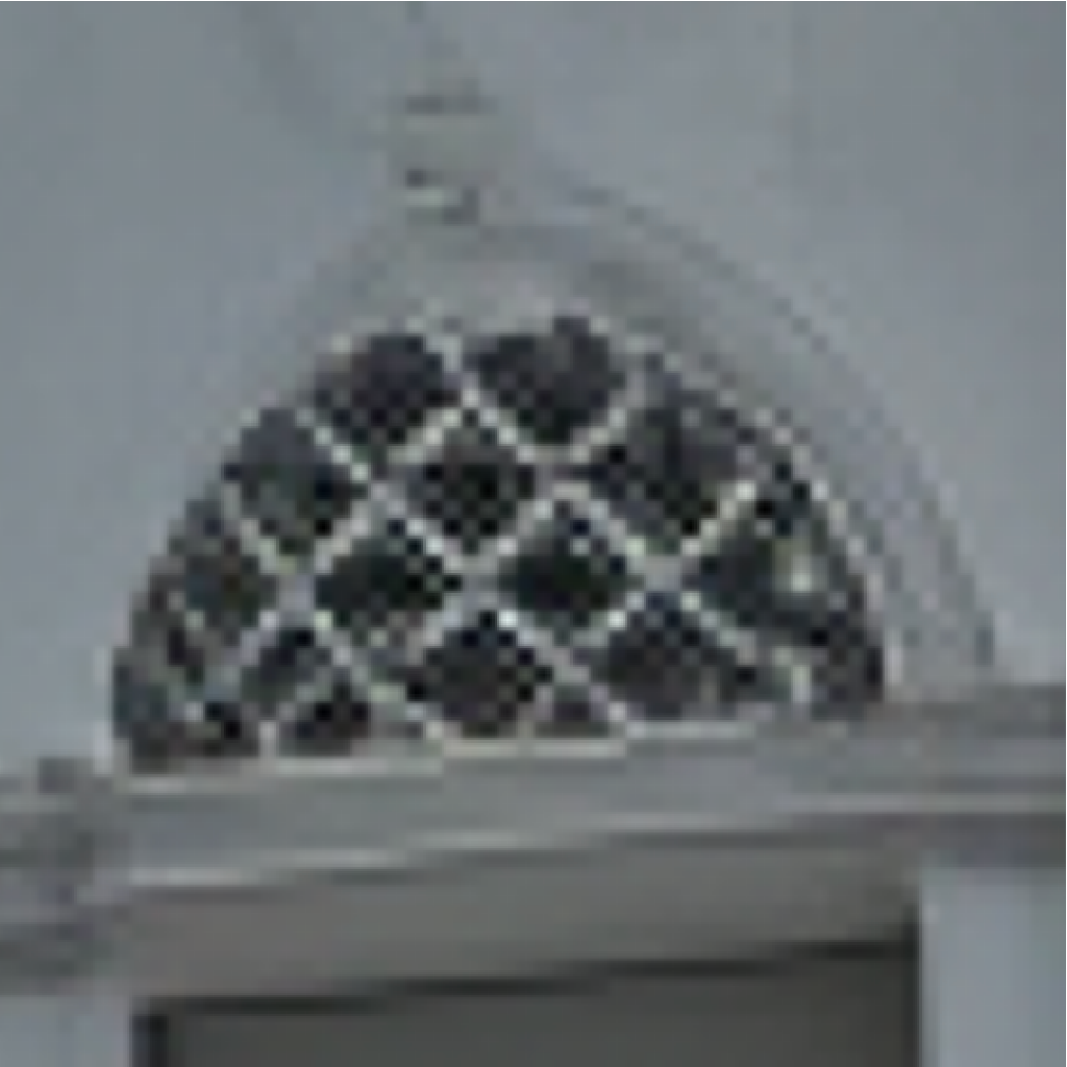}
        \caption*{\centering \scriptsize \textcolor{red}{w/ ``Building'' Prior}}}
    \end{subfigure}
    \begin{subfigure}[b]{0.24\linewidth}
        \vtop{\centering
        \includegraphics[width=\linewidth]{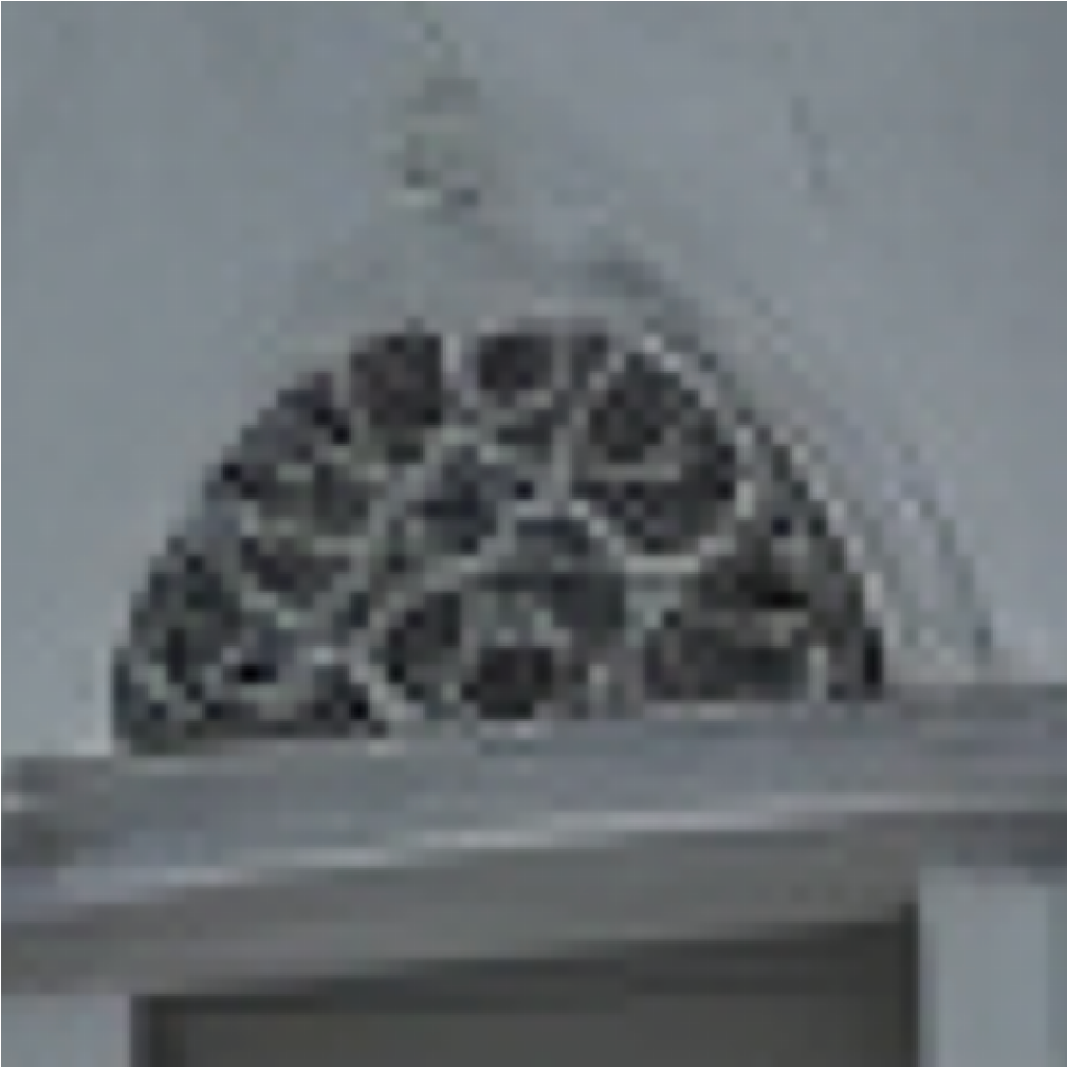}
        \caption*{\centering \scriptsize w/ ``Soil'' Prior}}
    \end{subfigure}

    \vspace{1mm}

    \begin{subfigure}[b]{0.24\linewidth}
        \vtop{\centering
        \includegraphics[width=\linewidth]{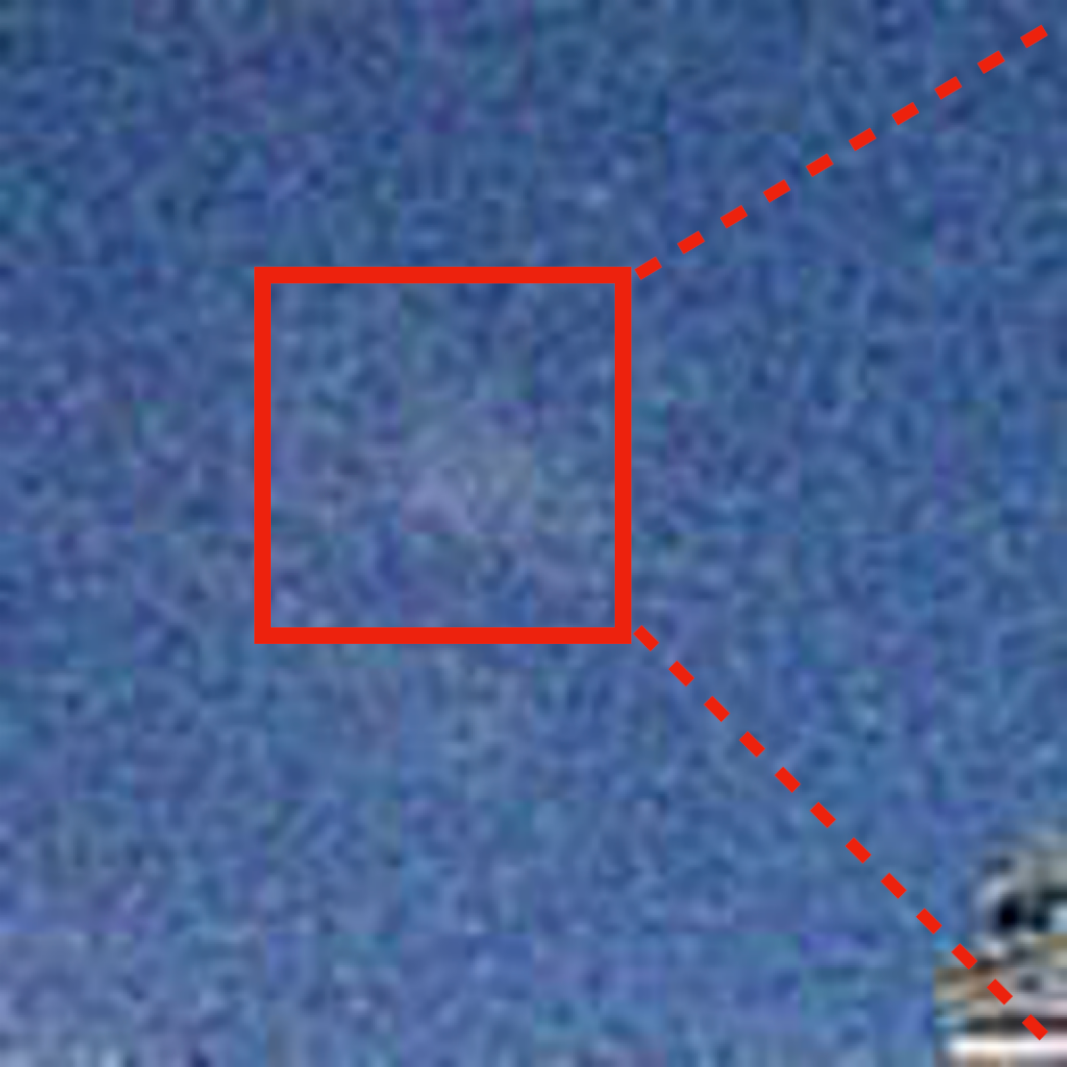}
        \caption*{\centering \scriptsize LR Image}}
    \end{subfigure}
    \begin{subfigure}[b]{0.24\linewidth}
        \vtop{\centering
        \includegraphics[width=\linewidth]{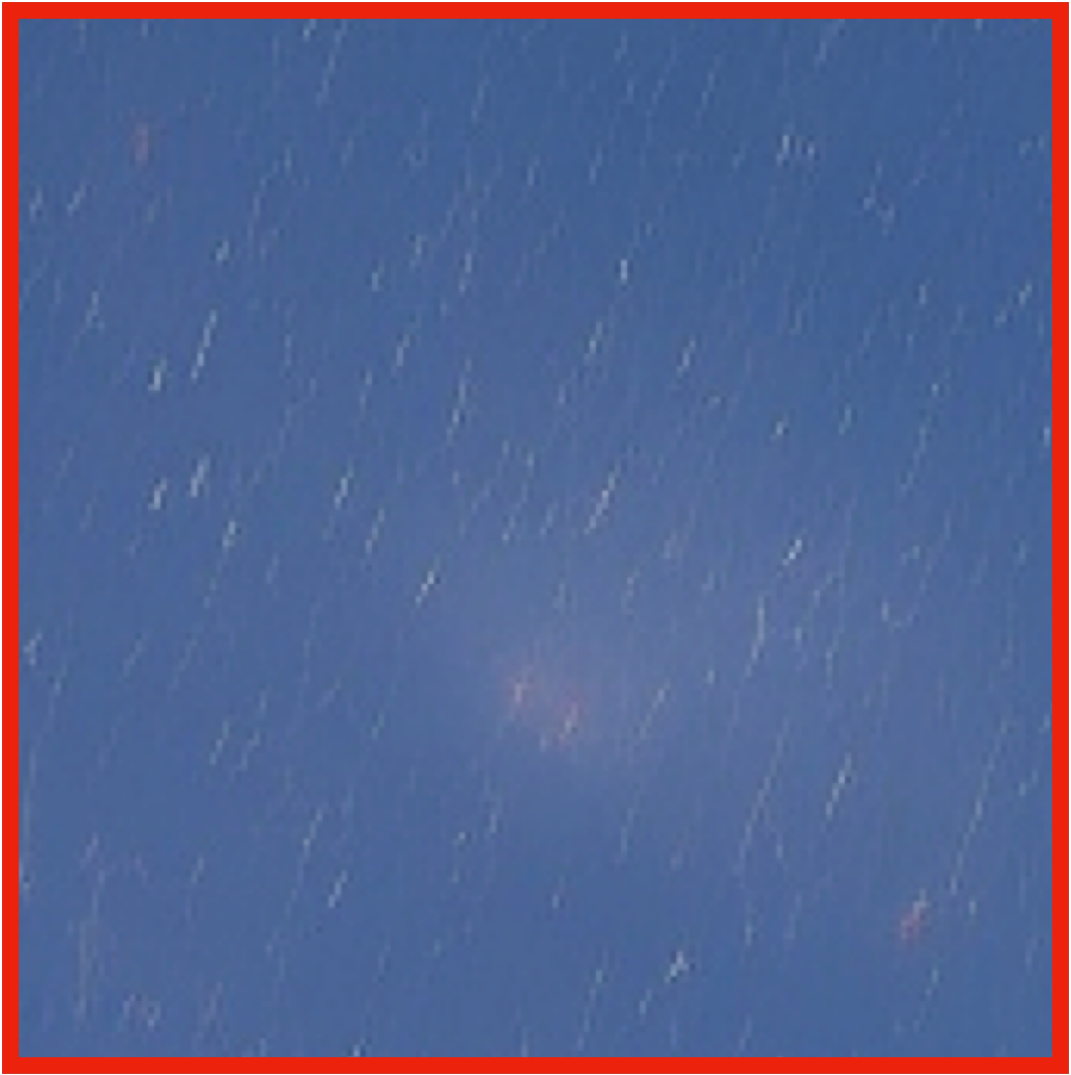}
        \caption*{\centering \scriptsize w/o Prior}}
    \end{subfigure}
    \begin{subfigure}[b]{0.24\linewidth}
        \vtop{\centering
        \includegraphics[width=\linewidth]{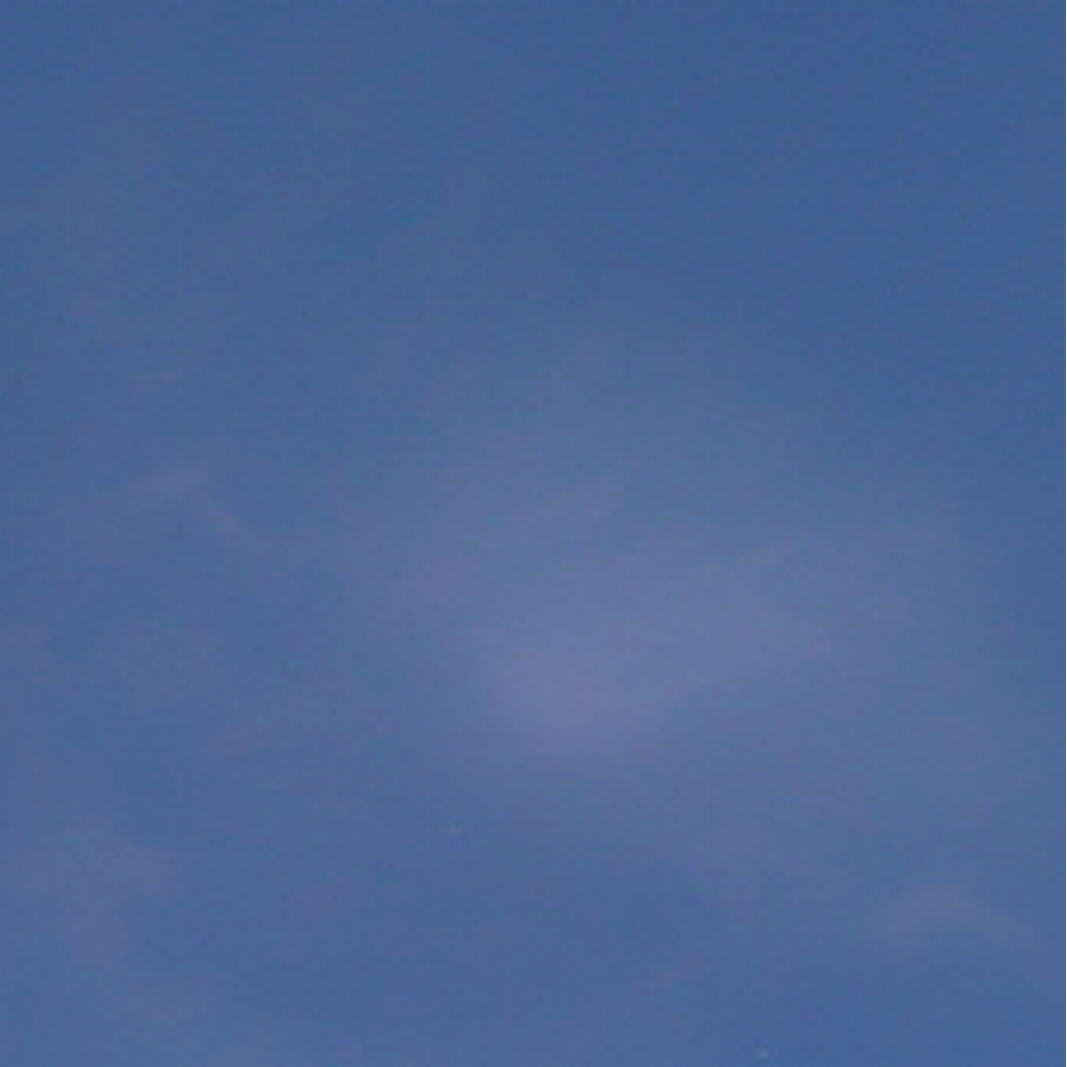}
        \caption*{\centering \scriptsize \textcolor{red}{w/ ``Sky'' Prior}}}
    \end{subfigure}
    \begin{subfigure}[b]{0.24\linewidth}
        \vtop{\centering
        \includegraphics[width=\linewidth]{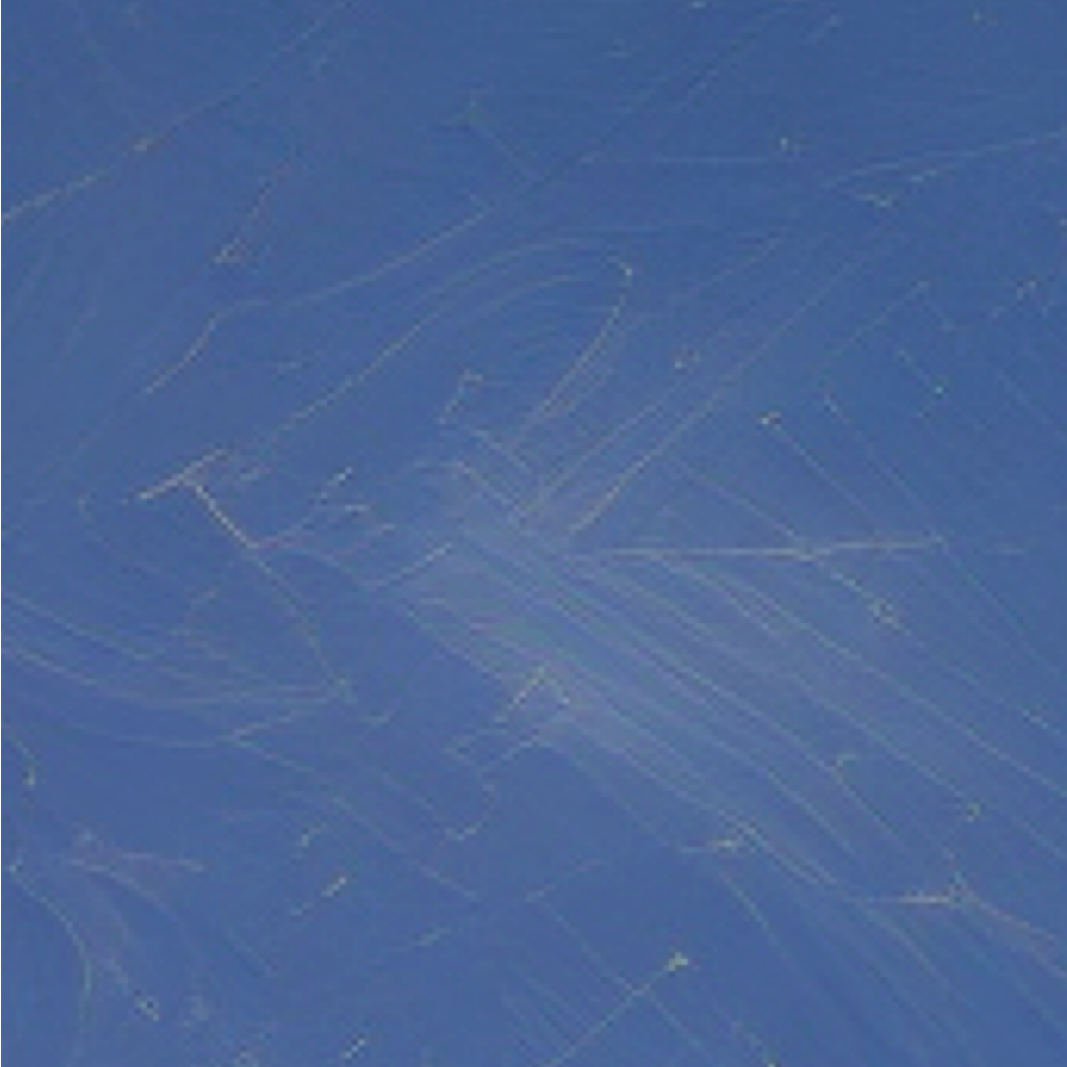}
        \caption*{\centering \scriptsize w/ ``Painting'' Prior}}
    \end{subfigure}
    \vspace{-2pt}
    \caption{Illustrative examples showing the impact of semantic priors on synthesized texture. With an appropriate prior, our framework can generate visually plausible results.}
    \label{fig:ablation_prior}
    \vspace{-3pt}
\end{figure}

This study examines how different semantic priors influence the synthesis of textures and the alignment with the actual scene context. Fig.~\ref{fig:ablation_prior} demonstrates the outcomes across three different scenarios: using a favorable semantic prior, an inappropriate prior, and no prior. As shown in the top row of Fig.~\ref{fig:ablation_prior}, when the proper ``Building'' \kelvin{You need to use `` and '' in latex (see the syntax)}%
\yuan{11/14 fixed}%
prior is used, {\ourmethod} generates fine structural details that accurately match the architectural content. In contrast, using an inappropriate prior such as ``Soil'' results in a less coherent structure. 
Similarly, in the bottom row of Fig.~\ref{fig:ablation_prior}, the ``Sky'' prior leads to visually pleasing and realistic sky textures. However, applying the ``Painting'' prior leads to the creation of streak-like textures with an artistic style, mismatching the open sky setting. 
In both examples, the absence of any prior leads to unwanted artifacts. These results highlight the crucial role of semantic priors in generating visually plausible image content and emphasize the necessity of the proposed guidance fusion strategy, demonstrating how semantic priors effectively guide the generation of fine details.
\kelvin{Examples?}
\yuan{I can try to make a graph showing the effect of using T2I and fusion strategy in supp. I should remove the text here.}

\vspace{-6pt}
\paragraph{Impact of Semantic Segmentation Quality.}
\label{sec::ablation_seg_quality}
\begin{table}[t]
\renewcommand{\arraystretch}{1.25}
\newcommand{\mytoprule}{\toprule[1.5pt]}
\newcommand{\mybottomrule}{\bottomrule[1.5pt]}
\newcommand{\first}[1]{\textcolor{red}{\underline{#1}}}
\newcommand{\second}[1]{\textcolor{blue}{#1}}
\centering
\caption{Quantitative evaluation utilizing different segmentation models in {\ourmethod}, where ``-T'', ``-B'', and ``-L'' denote using Mask2Former with Swin-T, Swin-B, and Swin-L backbones, respectively. The mIoU values are evaluated on the ADE20K~\cite{ade20k} validation set. The best results are in \first{red}, with the second-best in \second{blue}. Better segmentation quality enhances \ourmethod{}'s output, while \ourmethod{} remains robust to varying segmentation quality.} 
\resizebox{1\columnwidth}{!}{%
    \begin{tabular}{lccccc}
    \toprule
    Method & PSNR$\uparrow$ & LPIPS$\downarrow$ & 
    MANIQA$\uparrow$ & CLIPIQA$\uparrow$ & mIoU$\uparrow$ \\
    \midrule 
    {\ourmethod}-T & 23.1408 & 0.3195 & \second{0.5642} & \second{0.7046} & 47.7 \\
    {\ourmethod}-B & \second{23.4003} & \second{0.3179} & 0.5603 & 0.7021 & 52.4 \\
    \midrule
    \textbf{{\ourmethod}-L} & \first{23.5303} & \first{0.3165} & \first{0.5741} & \first{0.7109} & 56.1 \\
    \bottomrule
    \end{tabular}
}
\label{tab:ablation_seg}
\end{table}
In this ablation study, we employ Mask2Former~\cite{mask2former} with three different backbones: Swin-T~\cite{swint}, Swin-B~\cite{swint}, and Swin-L~\cite{swint}, to assess the impact of segmentation quality on the performance of {\ourmethod}. As shown in Table~\ref{tab:ablation_seg}, the Mask2Former model with Swin-T backbone achieves the highest mIoU on the ADE20K semantic segmentation benchmark, making it the most reliable source of segmentation information for our framework. Consequently, the best results across all the adopted metrics are achieved when using the segmentation results generated by Mask2Former with Swin-T backbone, showing that higher segmentation quality directly benefits the SR process. Additionally, {\ourmethod} maintains preferable visual quality even when employing the less capable Swin-B and Swin-L backbones in Mask2Former, presenting only minor variations in perceptual-oriented metrics such as LPIPS~\cite{lpips}, MANIQA~\cite{maniqa}, and CLIPIQA~\cite{clipiqa}. These results demonstrate that the proposed {\ourmethod} is robust to different levels of segmentation quality, effectively leveraging semantic priors throughout the generation process.
\ds{How about a different semantic segmentation method, even for inference only?}

\ds{Discussions: will our method apply to similar tasks, such as denoising/deblurring/restoration?}
\yuan{I think it works. For example, in DiffBIR~\cite{diffbir}, they first uses a pre-trained restoration module to derive a clean LR image, then apply SD model to synthesize high-quality images. Our framework can be apply to the latter part to enhance performance. The above example is a relatively brute-force pipeline, but if we can apply some trick to remove the degradation first (maybe also utilize segmentation information), then it's just a conditional generation task for the SD model.}

\section{Conclusion}
\label{sec::conclusion}

We have presented {\ourmethod}, a novel framework for Real-world Image Super-Resolution (Real-ISR) that leverages semantic segmentation to guide pre-trained T2I diffusion models. {\ourmethod} combines concise semantic label-based prompts with dense spatial guidance, effectively addressing the limitations of existing methods that rely solely on noisy, coarse-grained text prompts. Extensive experiments in diverse real-world scenarios show that {\ourmethod} produces high-quality results with enhanced semantic consistency compared to existing Real-ISR methods.

{
    \small
    \bibliographystyle{ieeenat_fullname}
    \bibliography{main}
}


\end{document}


\maketitle

\section{Overview}
\label{sec::supp_overview}
This supplementary material provides additional details and analyses to further support and complement the findings presented in the main text. Specifically, we first provides an anonymous GitHub link to our codebase in Section~\ref{sec::supp_github} for reproducibility assessment. Additionally, we include extensive analyses in Section~\ref{sec::supp_analysis}, along with additional implementation details in Section~\ref{sec::supp_implementation} to deliver further understanding of the proposed \ourmethod{}. 
Furthermore, we discuss the effects of employing open-world segmentation as alternative semantic guidance in Section~\ref{sec::supp_design}, demonstrating how this design contrast with the overall success of our approach. Finally, we include qualitative comparisons with state-of-the-art (SoTA) methods in Section~\ref{sec::supp_qualitative}.

\section{Anonymous GitHub Link}
\label{sec::supp_github}
For evaluating reproducibility, we provide the anonymous GitHub link of our project: \url{https://anonymous.4open.science/r/CVPR2025-1822-HoliSDiP}

\section{More Analyses}
\label{sec::supp_analysis}
In this section, we present analyses on model size, semantic-adaptive feature transformation, and the perception-fidelity trade-off to offer in-depth understanding of our proposed framework.

\paragraph{Model Size.}
Table~\ref{tab:supp_model_size} presents a quantitative comparison between \ourmethod{} and its baseline model. The baseline model is a re-implementation of SeeSR~\cite{seesr}, with training configurations aligned to our framework, such as batch size and total training iterations. This comparison highlights both the performance improvements and parameter efficiency achieved by our proposed method.

First, \ourmethod{}-T achieves remarkable enhancements in perceptual-oriented metrics, including a gain of 1.6428 in MUSIQ~\cite{musiq},  MANIQA~\cite{maniqa}, and CLIPIQA~\cite{clipiqa}, compared to the baseline model, with only a marginal 2.3\% increase in parameter count. This demonstrates the efficiency of our framework in leveraging semantic guidance for enhanced visual quality. Furthermore, \ourmethod{}-B builds upon these gains, delivering favorable performance over the baseline in both visual quality and fidelity metrics. Notably, this is achieved with less than a 5\% increase in parameters compared to the baseline, showcasing the effectiveness of our framework in capturing precise and holistic semantics to improve overall image quality. Finally, \ourmethod{}-L achieves the highest scores across all evaluated metrics with an enhanced segmentation model. These results validate the critical role of segmentation information in guiding the text-to-image diffusion process and highlight the ability of our framework to effectively leverage segmentation cues to optimize generative performance.
\begin{table}[t]
\renewcommand{\arraystretch}{1.3}
\newcommand{\first}[1]{\textcolor{red}{\textbf{#1}}}
\newcommand{\second}[1]{\textcolor{blue}{\underline{#1}}}
\centering
\caption{Quantitative comparison on the RealSR~\cite{realsr} dataset. ``\#Param'' denotes the total number of parameters in the T2I SR model (in billion). The best and second-best results are highlighted in \first{red}, and \second{blue}, respectively.} 
\resizebox{1\columnwidth}{!}{%
    \begin{tabular}{lccccc}
    \toprule
    Method & PSNR$\uparrow$ & MUSIQ$\uparrow$ & 
    MANIQA$\uparrow$ & CLIPIQA$\uparrow$ & \#Param (B) \\
    \midrule
    Baseline & 23.3547 & 67.5997 & 0.5315 & 0.6632 & 2.511 \\
    \ourmethod{}-T & 23.1408 & \second{69.2425} & \second{0.5642} & \second{0.7046} & 2.569 (+2.3\%)\\
    \ourmethod{}-B & \second{23.4003} & 69.2185 & 0.5603 & 0.7021 & 2.629 (+4.7\%)\\
    \ourmethod{}-L & \first{23.5303} & \first{69.7217} & \first{0.5741} & \first{0.7109} & 2.737 (+9.0\%) \\
    \bottomrule
    \end{tabular}
}
\label{tab:supp_model_size}
\end{table}

\paragraph{Semantic-Adaptive Feature Transformation.}
\begin{figure}[t]
    \centering
    \begin{subfigure}[b]{0.24\linewidth}
        \vtop{\centering
        \includegraphics[width=\linewidth]{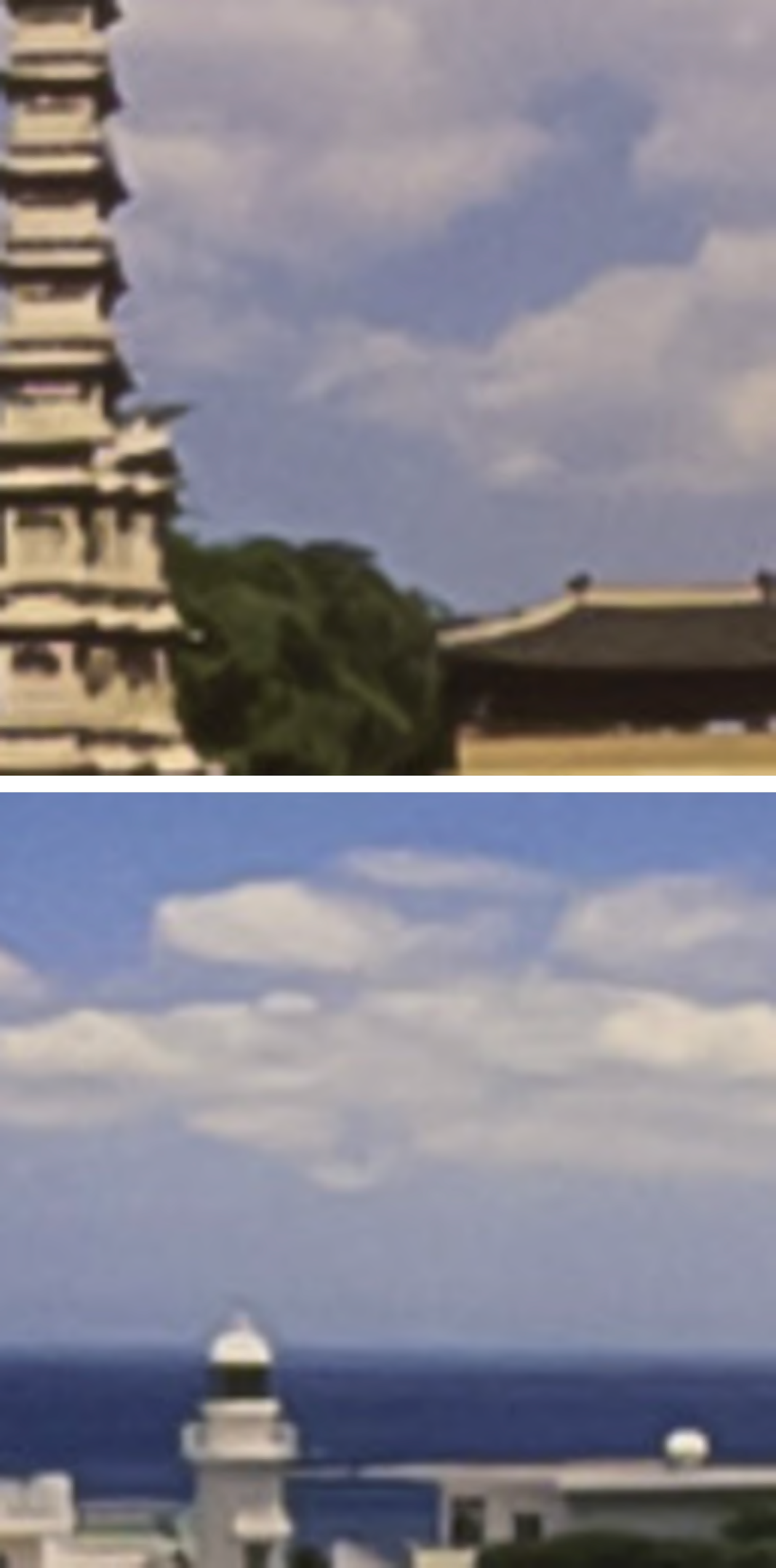}
        \caption*{\centering \scriptsize LR Image}}
    \end{subfigure}
    \begin{subfigure}[b]{0.24\linewidth}
        \vtop{\centering
        \includegraphics[width=\linewidth]{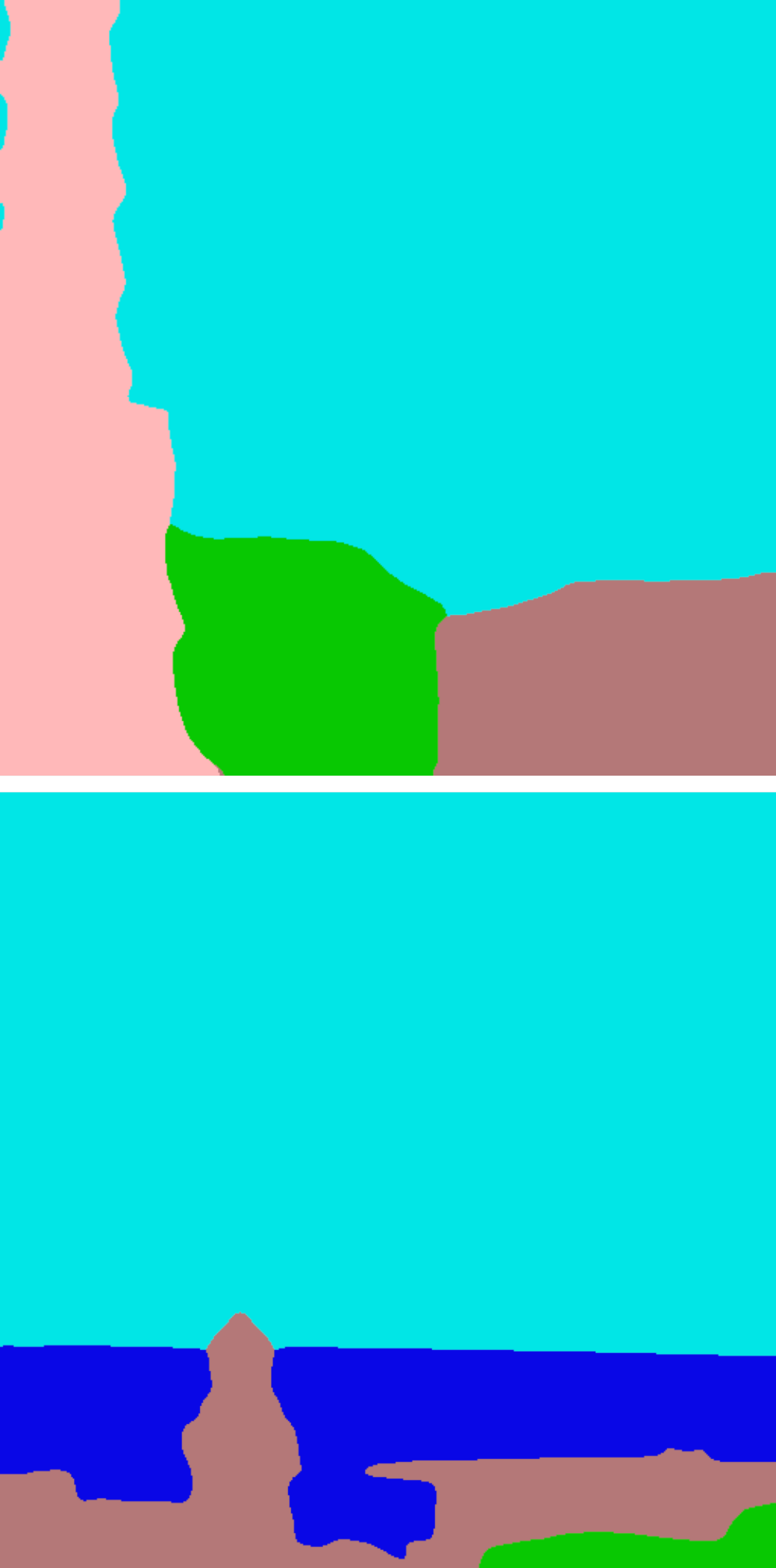}
        \caption*{\centering \scriptsize Segmentation Mask}}
    \end{subfigure}
    \begin{subfigure}[b]{0.24\linewidth}
        \vtop{\centering
        \includegraphics[width=\linewidth]{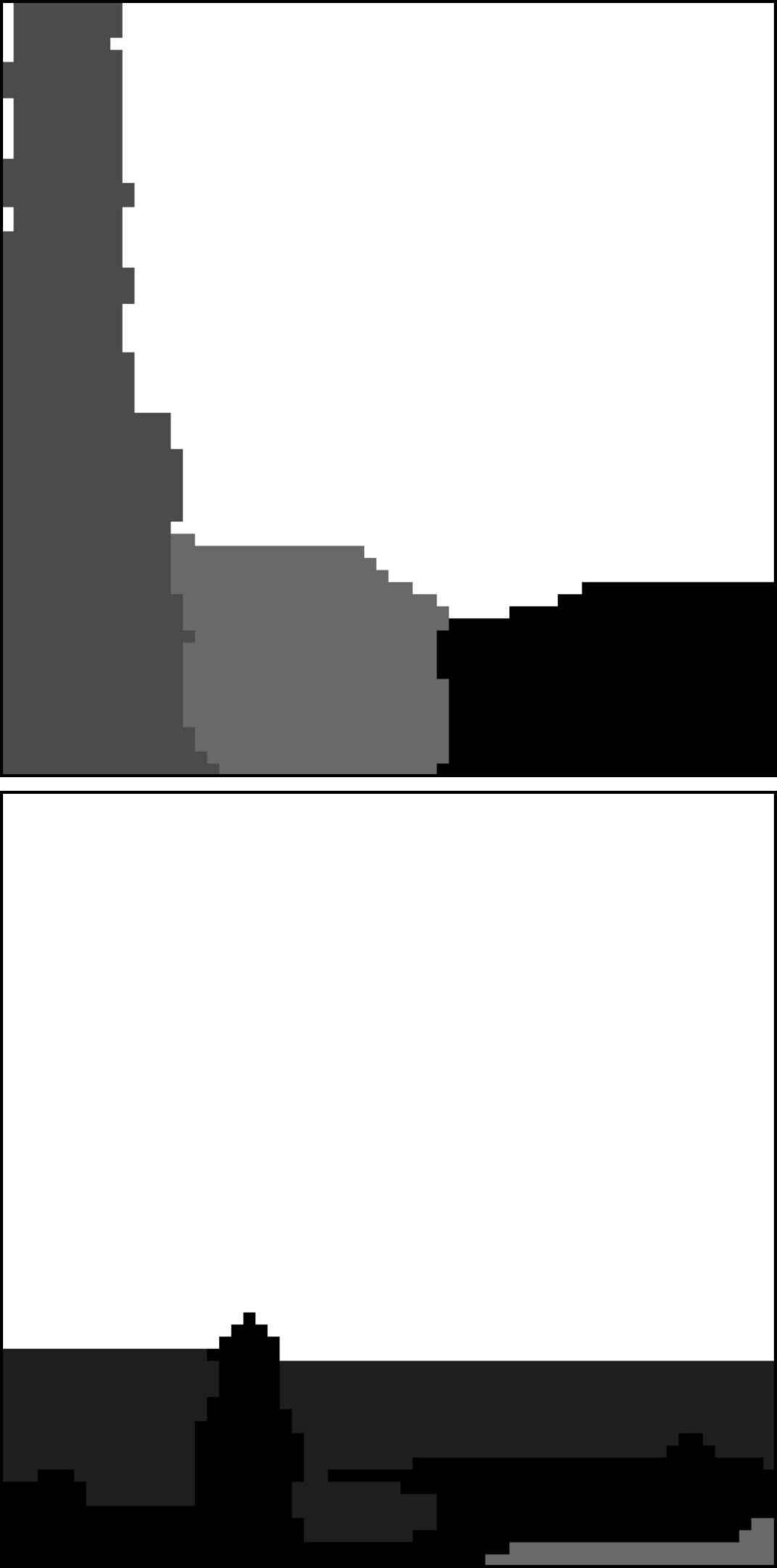}
        \caption*{\centering \scriptsize $\gamma_0$}}
    \end{subfigure}
    \begin{subfigure}[b]{0.24\linewidth}
        \vtop{\centering
        \includegraphics[width=\linewidth]{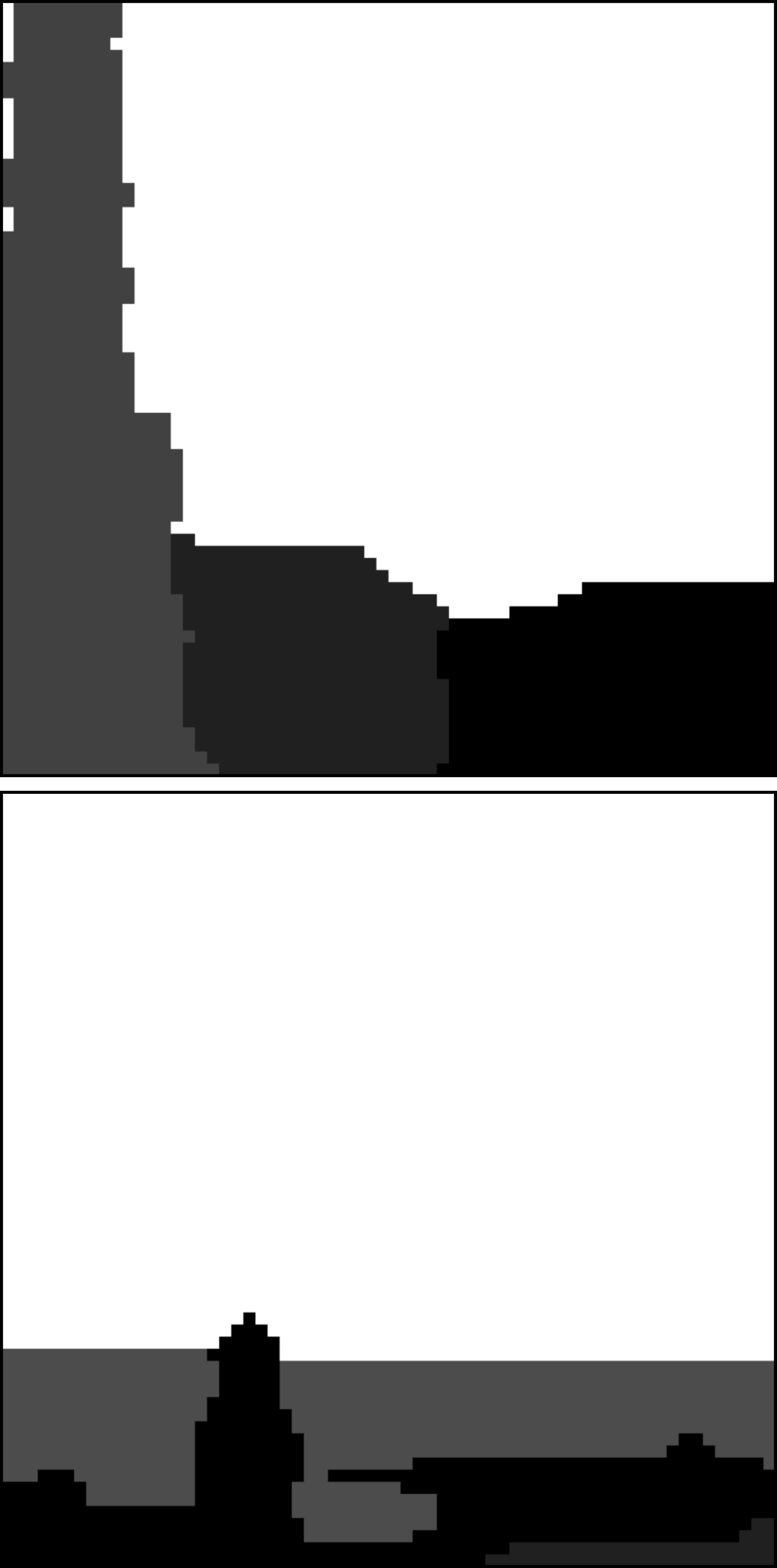}
        \caption*{\centering \scriptsize $\beta_0$}}
    \end{subfigure}
    \caption{Illustrative examples of $\gamma$ and $\beta$ parameter maps in the semantic-adaptive feature transformation (SAFT) block, where $\gamma_i$ and $\beta_i$ denote the $i$-th channels of the $\gamma$ and $\beta$ maps, respectively. These parameter maps align with the spatial structure of the segmentation mask, ensuring object-specific transformations.}
    \label{fig:supp_saft}
\end{figure}

In Section 3.4, we present the semantic-adaptive feature transformation (SAFT) block, which employs a sequence of 1$\times$1 convolutions to encode information from either the segmentation mask or the Segmentation-CLIP Map to generate feature refinement parameters. This design enables distinct semantic regions to apply tailored transformations. 

Fig.~\ref{fig:supp_saft} depicts the segmentation results of the LR image along with the corresponding $\gamma$ and $\beta$ transformation parameter maps. Specifically, we visualize the $0$-th channel of the encoded parameters $\gamma$ and $\beta$ derived from the segmentation mask as an example, where other channels demonstrate consistent behavior. These results indicate the spatial structures of $\gamma$ and $\beta$ maps closely align with the segmentation mask, validating that the SAFT blocks effectively leverage class-specific priors to refine individual objects, which is essential for enhancing local detail generation.

\paragraph{Perception-Distortion Trade-off.}
%
\begin{figure}[t]
    \centering
    \begin{tabular}{@{}c@{\hskip 5pt}c@{\hskip 5pt}c@{}}
        \begin{minipage}{0.19\textwidth}
            \includegraphics[width=\textwidth]{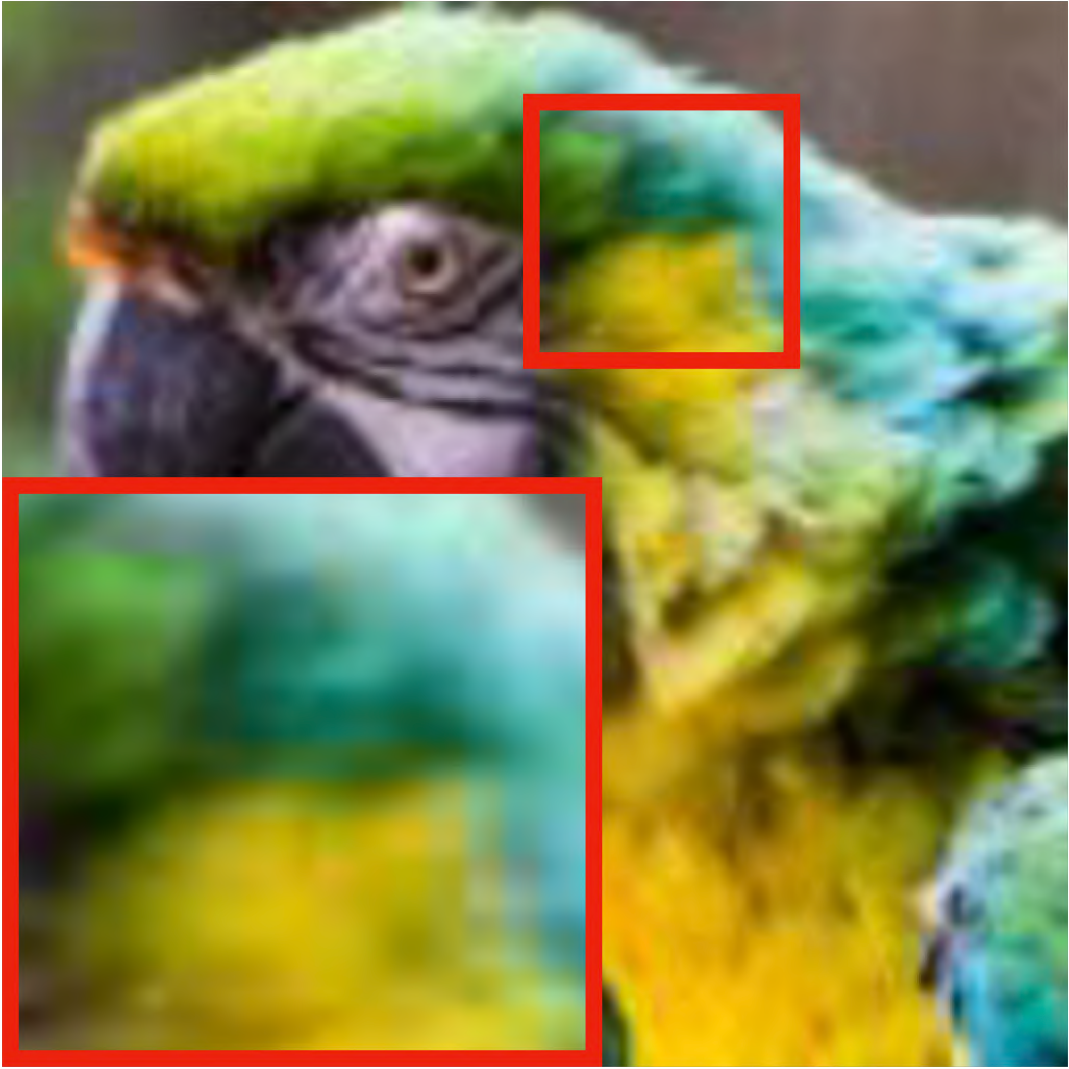} \\ [2pt]
            \centering \footnotesize {LR Image}
        \end{minipage}
        &
        \begin{minipage}{0.19\textwidth}
            \includegraphics[width=\textwidth]{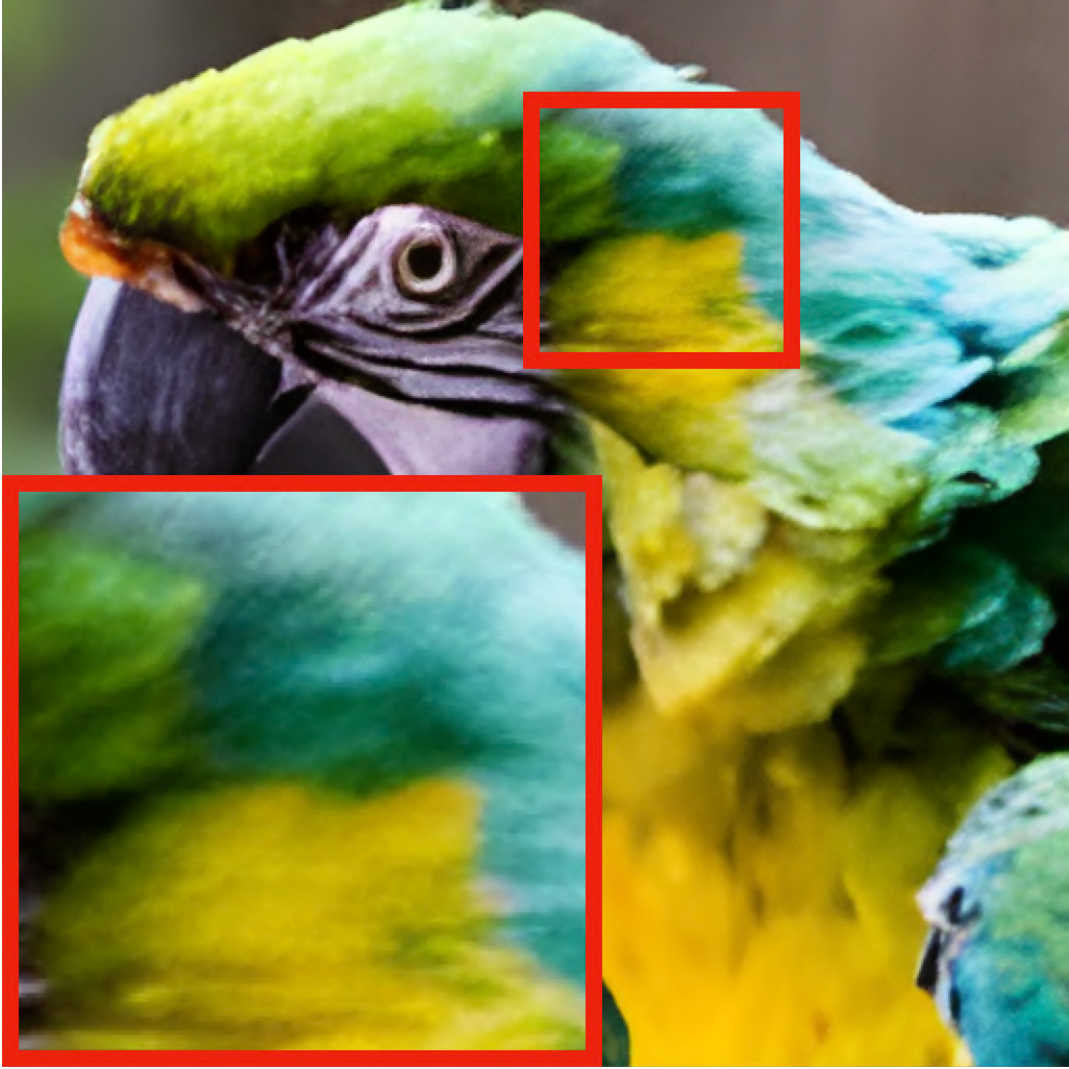} \\ [2pt]
            \centering \footnotesize {Real-ESRGAN}
        \end{minipage}
        &
        \begin{minipage}{0.05\textwidth}
            \raggedright \scriptsize
            {PSNR:} \textcolor{red}{23.6065} \\
            {SSIM:} \textcolor{red}{0.7209} \\
            {LPIPS:} \textcolor{red}{0.2374} \\
            {MUSIQ:} 69.4836 \\
            {MANIQA:} 0.3114 \\
            {CLIPIQA:} 0.4926
        \end{minipage}
        \\ \\
        \begin{minipage}{0.19\textwidth}
            \includegraphics[width=\textwidth]{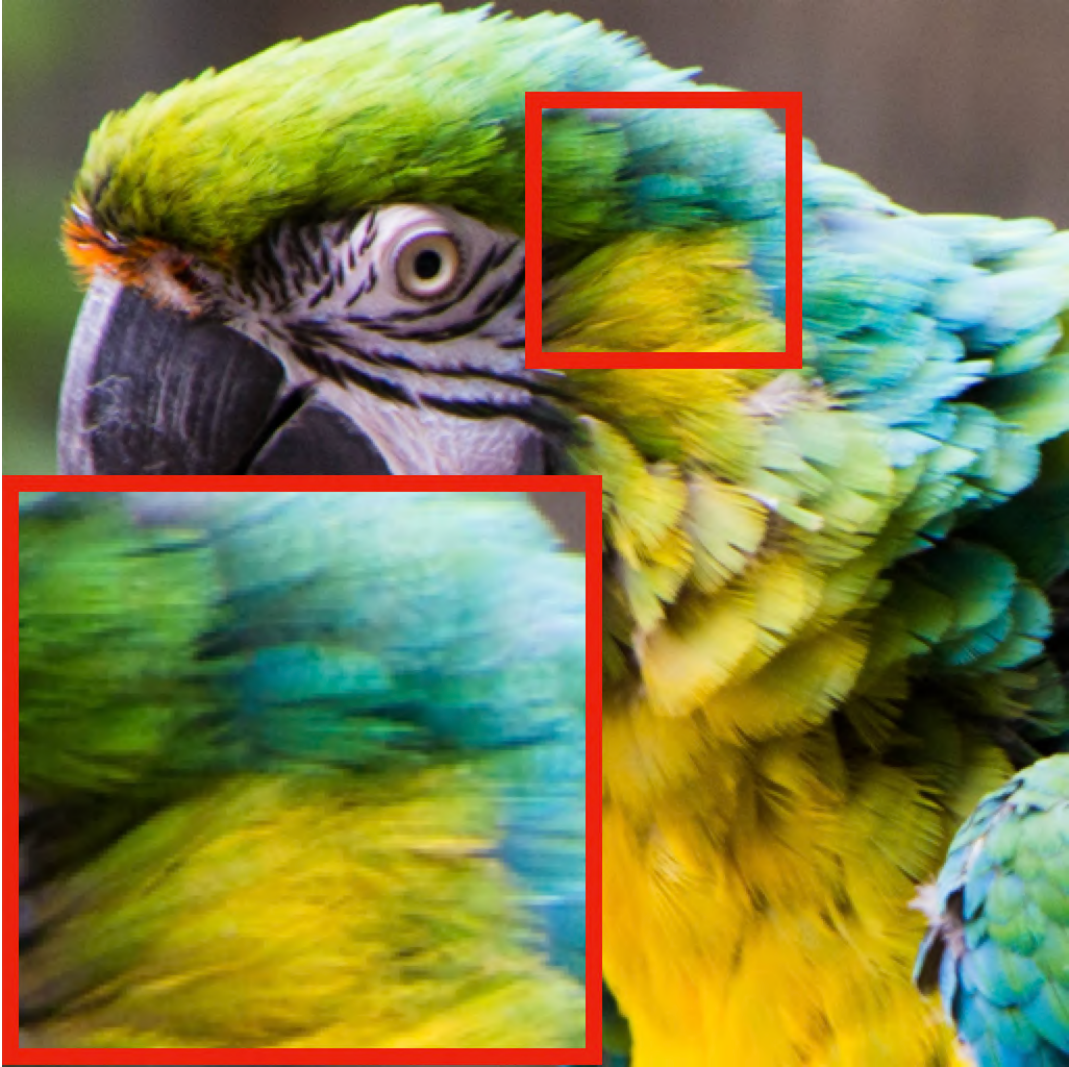} \\ [2pt]
            \centering \footnotesize {Ground Truth}
        \end{minipage}
        &
        \begin{minipage}{0.19\textwidth}
            \includegraphics[width=\textwidth]{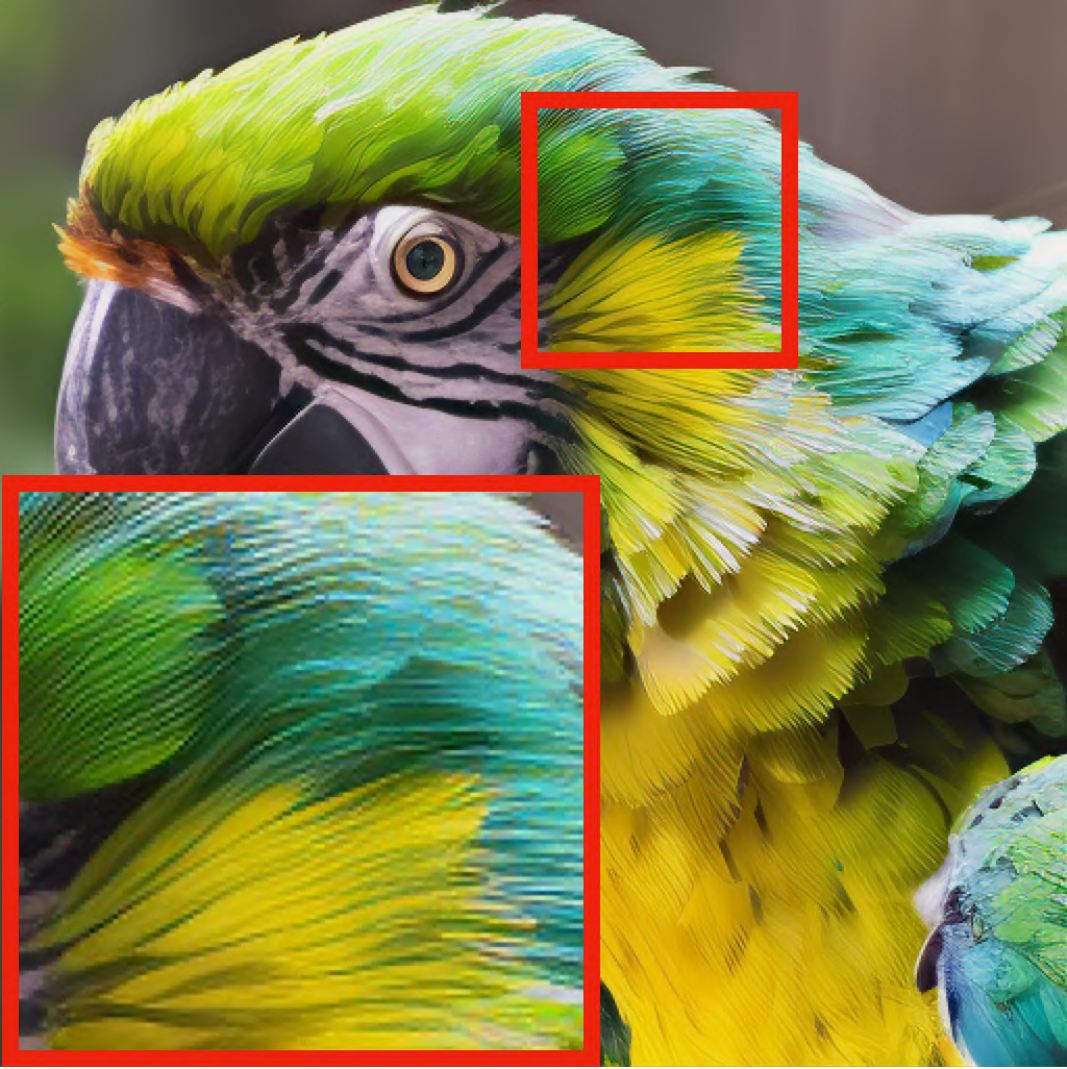} \\ [2pt]
            \centering \footnotesize {\ourmethod{} (Ours)}
        \end{minipage}
        &
        \begin{minipage}{0.05\textwidth}
            \raggedright \scriptsize
            {PSNR:} 21.1837 \\
            {SSIM:} 0.5734 \\
            {LPIPS:} 0.3100 \\
            {MUSIQ:} \textcolor{red}{77.7279} \\
            {MANIQA:} \textcolor{red}{0.6927} \\
            {CLIPIQA:} \textcolor{red}{0.8931}
        \end{minipage}
    \end{tabular}
    \caption{Quantitative and qualitative comparisons demonstrating the perception-distortion trade-off. Evaluation scores are computed over the entire image. Our \ourmethod{} produces sharper fur details, while Real-ESRGAN generates smoother textures, leading to better non-reference and reference-based scores, respectively.}
    \label{fig:supp_trade-off}
\end{figure}
In Section 4.2 of our manuscript, we discuss the perception-fidelity trade-off~\cite{tradeoff}, pointing out that visually sharper images do not necessarily achieve better performance on reference-based metrics, such as PSNR, SSIM~\cite{ssim}, and LPIPS~\cite{lpips}. As a result, while diffusion prior-based methods are capable of generating high-quality images, they often struggle to achieve consistent advantages across all evaluation metrics. Fig.~\ref{fig:supp_trade-off} illustrates this phenomenon: the output of the proposed \ourmethod{} exhibits sharper fur textures, leading to favorable performance on non-reference metrics, including MUSIQ, MANIQA, and CLIPIQA. In contrast, Real-ESRGAN~\cite{realesrgan} produces smoother textures, achieving higher scores on reference-based metrics.

These results highlight that the two categories of metrics prioritize different texture characteristics, making it challenging to optimize for both simultaneously. Nevertheless, as demonstrated in Table 1, \ourmethod{} performs well against other diffusion prior-based methods by consistently delivering better perceptual quality across diverse scenarios while maintaining comparable fidelity levels. This validates the effectiveness of our approach in improving the overall quality of super-resolved outputs.

\section{Implementation Details}
\label{sec::supp_implementation}
In this section, we provide the implementation details of our proposed Semantic Label-Based Prompting and Segmentation-CLIP Map, facilitating better reproducibility.

\paragraph{Semantic Label-Based Prompting.}
Our proposed Semantic Label-Based Prompting (SLBP) leverages semantic labels as text prompts for the text-to-image (T2I) diffusion model, where these labels are derived from the 150 categories defined in the ADE20K~\cite{ade20k} dataset. 
Most categories are represented by a single word; for instance, class 0 is labeled as ``wall'', and class 1 as ``building''. When an LR image contains semantic contexts such as ``wall'' and ``building'', these labels are concatenated into a text prompt of ``wall, building'' for the T2I model.
For categories represented by multiple words, such as class 34 labeled as ``rock, stone'', both terms are included in the text prompt.

\paragraph{Segmentation-CLIP Map.}
In our Segmentation-CLIP Map (SCMap) implementation, each pixel’s label in the segmentation mask is mapped to its corresponding CLIP~\cite{clip} text embedding. To eliminate redundant computation of these embeddings during training, we pre-compute the CLIP embeddings for all the 150 class labels defined in the ADE20K dataset and store them in a list. Consequently, when converting a segmentation mask into the SCMap, we simply retrieve the corresponding embeddings from the pre-computed list, greatly improving computational efficiency.

\section{Open-World Segmentation}
\label{sec::supp_design}
\begin{table}[t]
\renewcommand{\arraystretch}{1.3}
\newcommand{\first}[1]{\textcolor{red}{\textbf{#1}}}
\newcommand{\second}[1]{\textcolor{blue}{\underline{#1}}}
\centering
\caption{Quantitative comparison on the RealSR~\cite{realsr} dataset. ``+o'' denotes using open-world segmentation mask as semantic guidance. The best results are highlighted in \first{red}. Open-world segmentation guidance can not effectively boost image quality due to the lack of class priors.} 
\resizebox{1\columnwidth}{!}{%
    \begin{tabular}{lcccc}
    \toprule
    Method & PSNR$\uparrow$ & MUSIQ$\uparrow$ & 
    MANIQA$\uparrow$ & CLIPIQA$\uparrow$ \\
    \midrule
    Baseline & 23.3547 & 67.5997 & 0.5315 & 0.6632 \\
    Baseline (+o) & 23.1601 & 67.2741 & 0.5232 & 0.6553 \\
    \ourmethod{} & \first{23.5303} & \first{69.7217} & \first{0.5741} & \first{0.7109} \\
    \bottomrule
    \end{tabular}
}
\label{tab:supp_sam}
\end{table}
\begin{figure}[t]
    \centering
    \begin{tabular}{@{}c@{\hskip 5pt}c}
        \begin{minipage}{0.2\textwidth}
            \includegraphics[width=\textwidth]{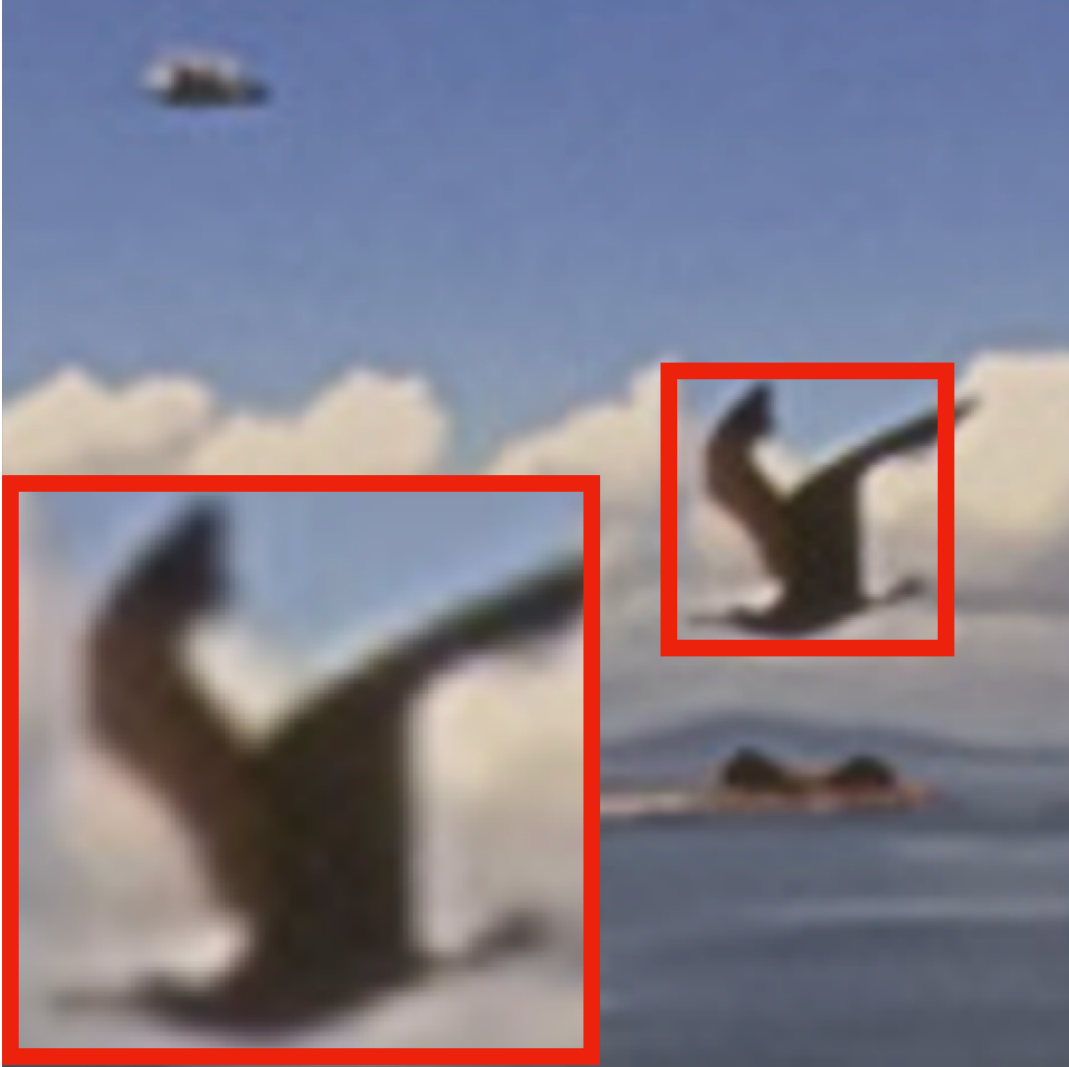} \\ [2pt]
            \centering \footnotesize {LR Image}
        \end{minipage}
        &
        \begin{minipage}{0.2\textwidth}
            \includegraphics[width=\textwidth]{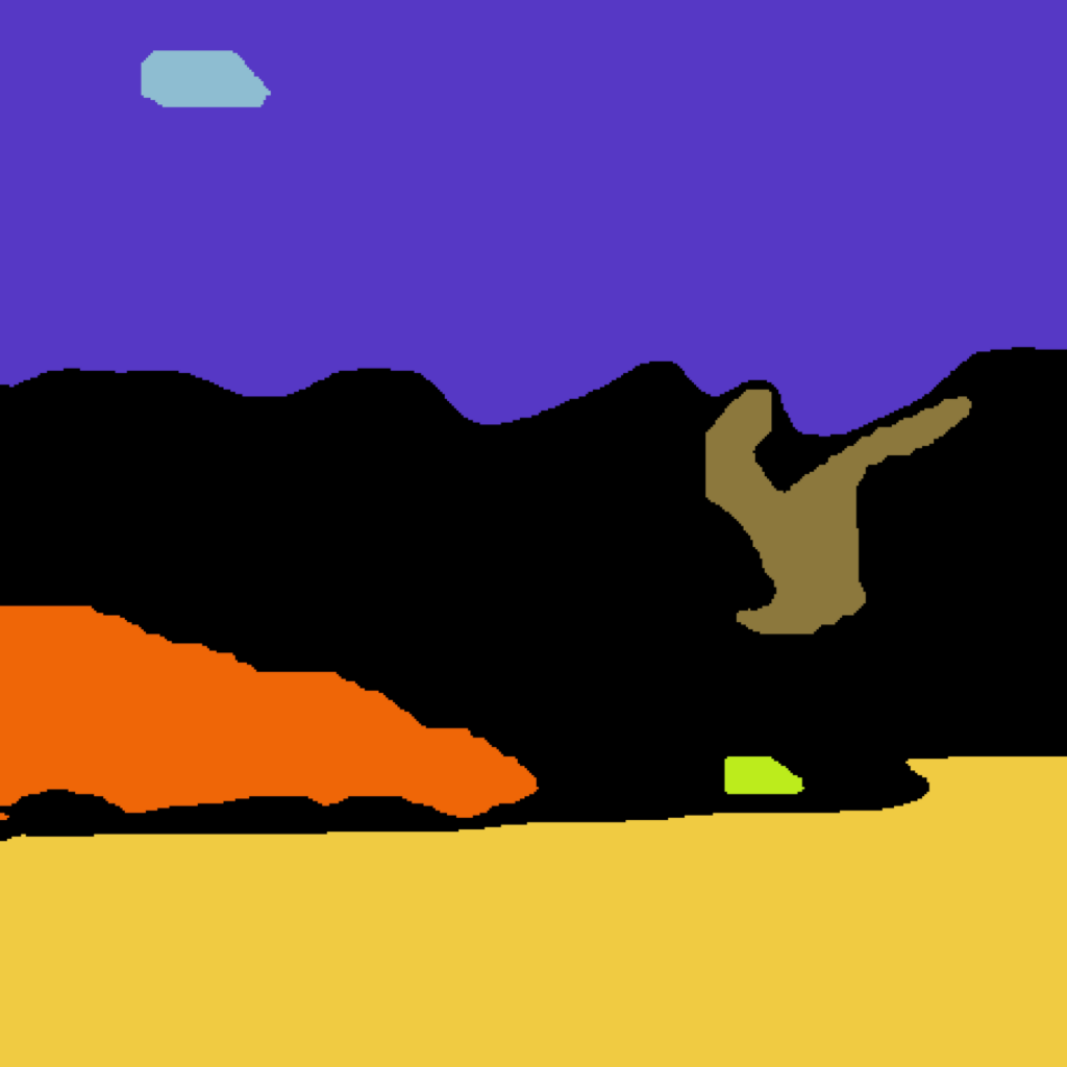} \\ [2pt]
            \centering \footnotesize {Segmentation Mask}
        \end{minipage}
        \\ \\
        \begin{minipage}{0.2\textwidth}
            \includegraphics[width=\textwidth]{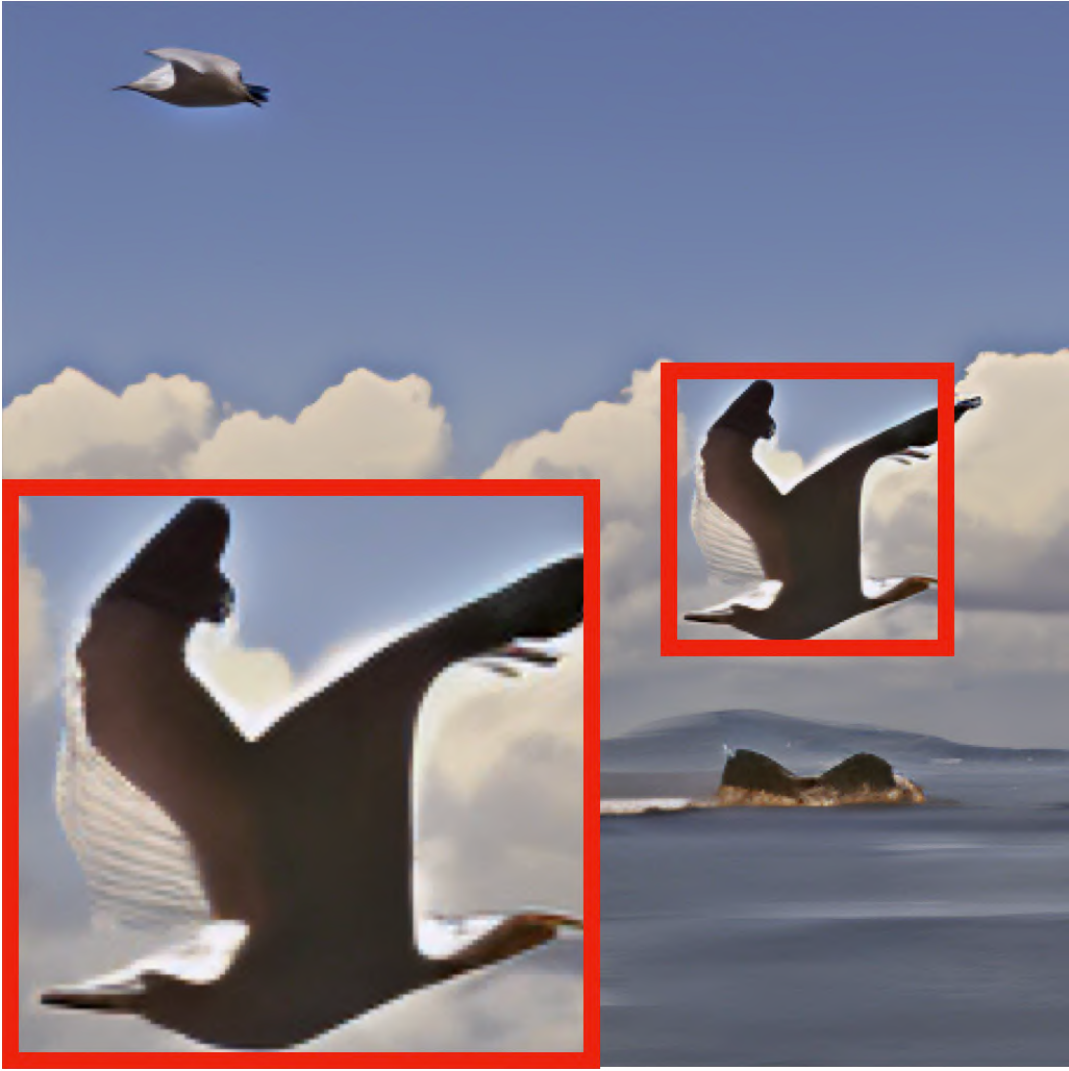} \\ [2pt]
            \centering \footnotesize {w/o mask}
        \end{minipage}
        &
        \begin{minipage}{0.2\textwidth}
            \includegraphics[width=\textwidth]{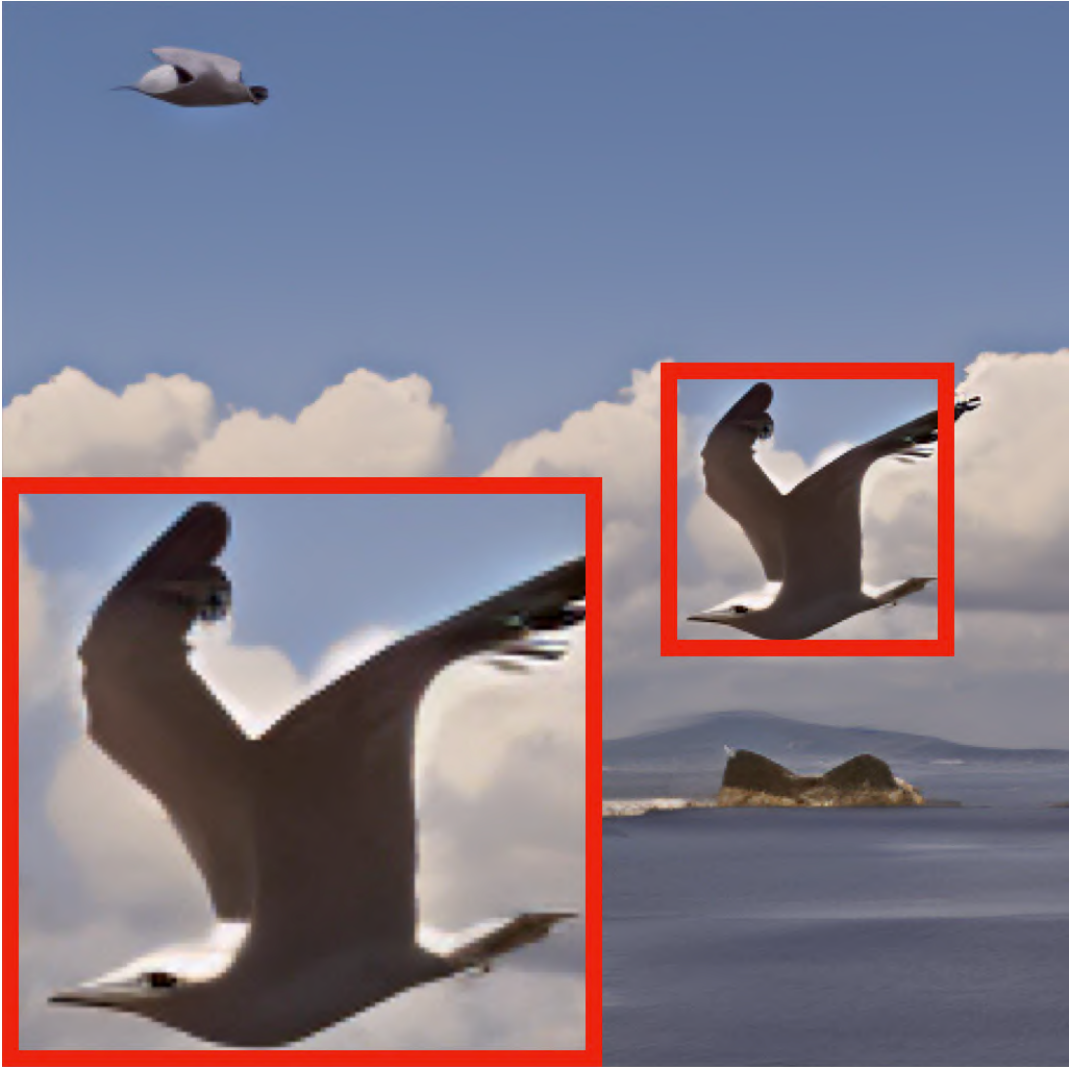} \\ [2pt]
            \centering \footnotesize {w/ mask}
        \end{minipage}
    \end{tabular}
    \caption{Open-world segmentation masks offer fine-grained object boundaries, which can sometimes prevent the model from hallucinating content beyond the object’s shape. However, due to the absence of categorical priors, these masks contribute minimally to improving overall image quality.}
    \label{fig:supp_sam}
\end{figure}
In this section, we analyze the impact of using open-world segmentation masks as alternative semantic guidance, utilizing masks generated by FastSAM~\cite{fastsam}. Unlike semantic segmentation, open-world segmentation lacks class labels, preventing the application of our proposed Semantic Label-Based Prompting. Additionally, the absence of labels hinders the transformation of segmentation masks into the Segmentation-CLIP Map (SCMap). In practice, we fuse the open-world segmentation mask with the LR image to create a conditioning signal for ControlNet, enabling it to provide semantic guidance to the T2I diffusion model.
As shown in Table~\ref{tab:supp_sam}, using open-world segmentation masks for guidance fails to improve overall image quality. The primary limitation lies in the absence of class priors, as these masks only capture object boundaries without offering meaningful semantic context. This constraint restricts the model's ability to perform semantic-aware refinement. In contrast, \ourmethod{} achieves notable improvements by leveraging semantic segmentation to provide dense semantic priors and precise text prompts simultaneously. 

While open-world segmentation masks generally struggles to improve overall image quality due to the lack of categorical priors, they can still be beneficial in specific cases. As shown in Fig.~\ref{fig:supp_sam}, these masks provide fine-grained boundaries, preventing T2I diffusion models from hallucinating details beyond objects. This indicates the potential of open-world segmentation in guiding generative models, particularly in ensuring spatial accuracy.
However, effectively leveraging open-world segmentation to provide meaningful semantics remains a challenge, highlighting the need for future research to address this limitation.

\section{Qualitative Comparison with SoTA}
\label{sec::supp_qualitative}
\begin{figure*}[t]
    \centering
    \vspace{1mm}
    \begin{subfigure}[b]{0.18\linewidth}
        \vtop{\centering
        \includegraphics[width=\linewidth]{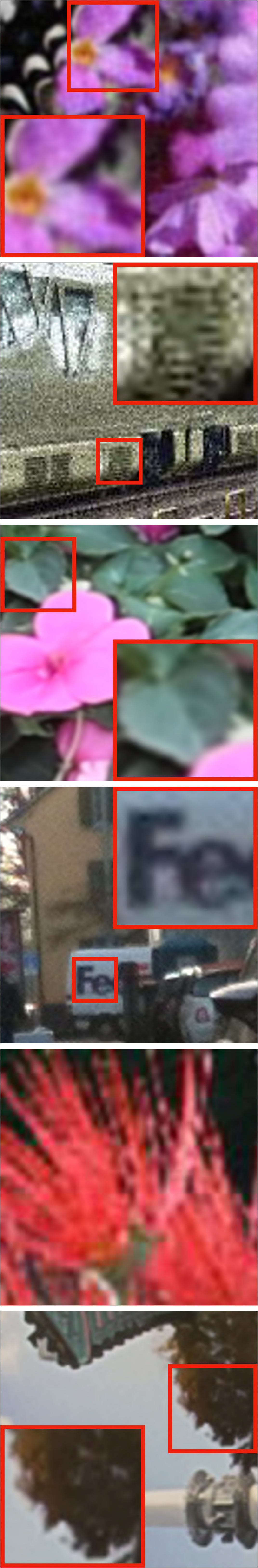}
        \caption*{\centering \scriptsize LR Image}}
    \end{subfigure}
    \begin{subfigure}[b]{0.18\linewidth}
        \vtop{\centering
        \includegraphics[width=\linewidth]{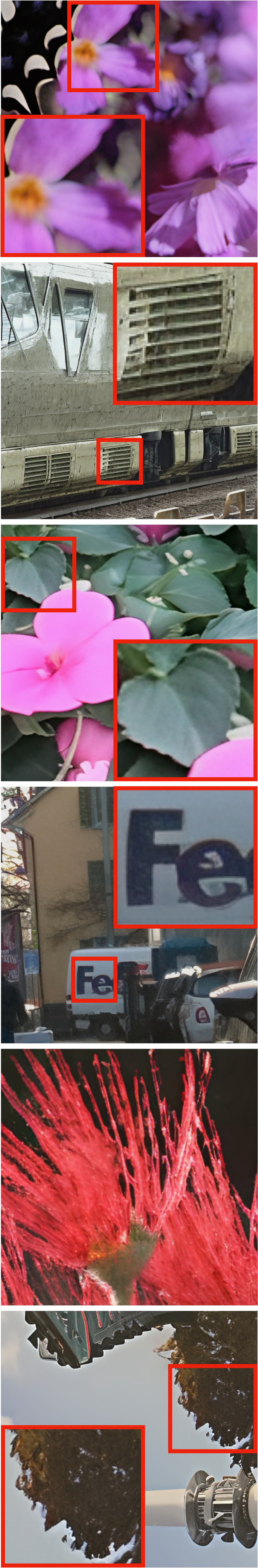}
        \caption*{\centering \scriptsize StableSR}}
    \end{subfigure}
    \begin{subfigure}[b]{0.18\linewidth}
        \vtop{\centering
        \includegraphics[width=\linewidth]{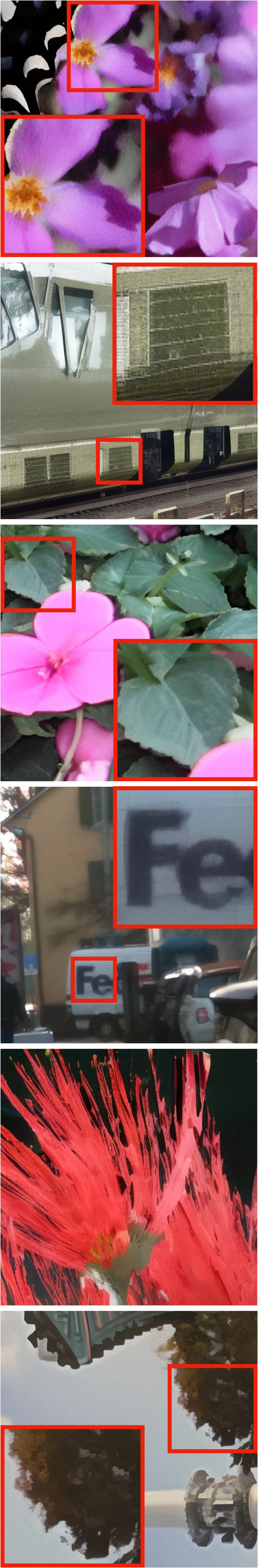}
        \caption*{\centering \scriptsize DiffBIR}}
    \end{subfigure}
    \begin{subfigure}[b]{0.18\linewidth}
        \vtop{\centering
        \includegraphics[width=\linewidth]{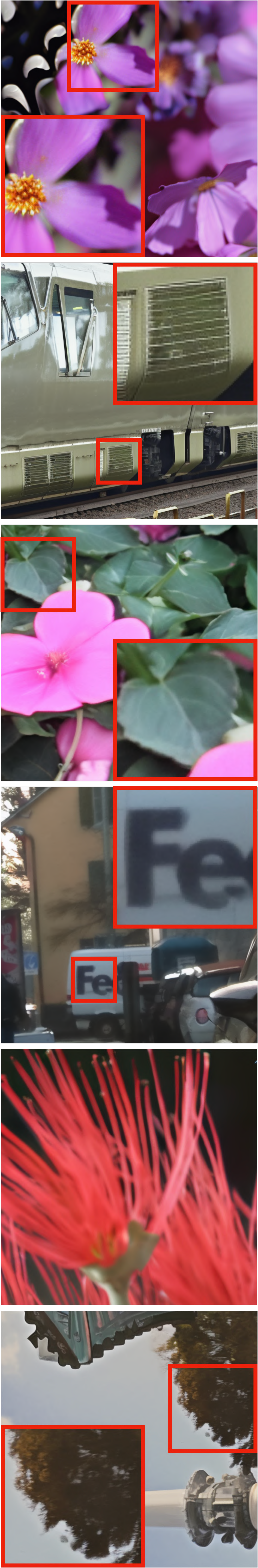}
        \caption*{\centering \scriptsize SeeSR}}
    \end{subfigure}
    \begin{subfigure}[b]{0.18\linewidth}
        \vtop{\centering
        \includegraphics[width=\linewidth]{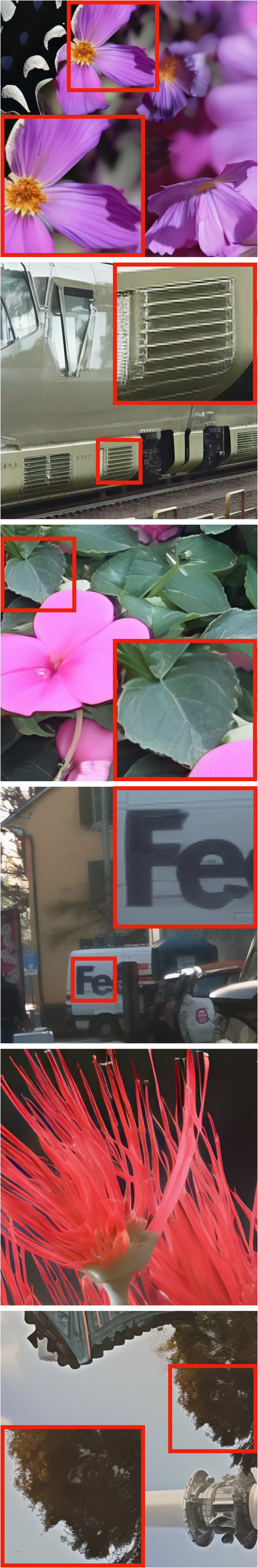}
        \caption*{\centering \scriptsize \textbf{{\ourmethod} (Ours)}}}
    \end{subfigure}
    
    \caption{Qualitative comparison between the proposed {\ourmethod} and contemporary diffusion prior-based Real-ISR methods. \ourmethod{} presents visually plausible details across various Real-ISR scenarios.}
    \label{fig:supp-qualitative-1}
\end{figure*}
Fig.~\ref{fig:supp-qualitative-1} presents additional qualitative comparisons between diffusion prior-based real-world image super-resolution (Real-ISR) methods for assessing the effectiveness of our framework. The proposed \ourmethod{} consistently synthesizes realistic texture and sharper details.

{
    \small
    \bibliographystyle{ieeenat_fullname}
    \bibliography{main}
}
